\def\year{2021}\relax
\documentclass[letterpaper]{article} 
\usepackage{aaai21}  
\usepackage{times}  
\usepackage{helvet} 
\usepackage{courier}  
\usepackage[hyphens]{url}  
\usepackage{graphicx} 
\usepackage{natbib}  
\usepackage{caption} 
\usepackage{mkolar_definitions}
\usepackage{booktabs}       

\title{Multinomial Logit Contextual Bandits:\\
Provable Optimality and Practicality}
\author{

    Min-hwan Oh\textsuperscript{\rm 1}
    and Garud Iyengar\textsuperscript{\rm 2}
    \\

}
\affiliations{

    \textsuperscript{\rm 1}Seoul National University, Seoul, South Korea\\
    \textsuperscript{\rm 2}Columbia University, New York, USA
    \\\textsuperscript{\rm 1}minoh@snu.ac.kr,  \textsuperscript{\rm 2}garud@ieor.columbia.edu

}

\begin{document}

\maketitle

\begin{abstract}
  We consider a sequential assortment selection problem where the user choice is given by a multinomial logit (MNL) choice model whose parameters are unknown. In each period, the learning agent observes a $d$-dimensional contextual information about the user and the $N$ available items, and  offers an assortment of size $K$ to the user, and observes the bandit feedback of the item chosen from the assortment. We propose upper confidence bound based algorithms for this MNL contextual bandit. The first algorithm is a simple and practical method which achieves an $\Tilde{\mathcal{O}}(d\sqrt{T})$ regret over $T$ rounds.
  Next, we propose a second algorithm which achieves a $\Tilde{\mathcal{O}}(\sqrt{dT})$ regret. This matches the lower bound for the MNL bandit problem, up to logarithmic terms, and improves on the best known result by a $\sqrt{d}$ factor. To establish this sharper regret bound, we present a non-asymptotic confidence bound for the maximum likelihood estimator of the MNL model that may be of independent interest as its own theoretical contribution. We then revisit the simpler, significantly more practical, first algorithm and show that a simple variant of the algorithm achieves the optimal regret for a broad class of important applications.
\end{abstract}

\section{Introduction}

In many of the human-algorithm interactions today, a  learning agent
(algorithm) makes sequential decisions and receives user (human) feedback
\emph{only} for the chosen decisions.
The multi-armed bandit \citep{LS19bandit-book} is a model for this
sequential decision making with partial feedback. It is a classic
reinforcement learning problem that exemplifies the dilemma of
exploration vs. exploitation. 
This multi-armed bandit model has found diverse applications, e.g. 
learning click-through rates in
search engines, product recommendations in online retailing, movie
suggestions on streaming services, news feeds, etc. Note that in
  several of the applications, the goal is to maximize an appropriate
  ``clickthrough'' 
rate.
Often information about the features of the agent's actions  
and contextual information
about the user are available. 
The contextual bandit extends the multi-armed
bandit by making the decision conditional on this context and feature
information.  
In many real-world problems including the aforementioned examples, the
agent offers  
a menu of options to the user, rather than a single option as
in traditional bandit action selection. The user chooses at most one of the offered options, 
and the agent receives a reward
associated with the user 
choice. 

In this paper, we consider a sequential assortment selection problem which
is a combinatorial variant of the bandit problem. The goal is to offer a
sequence of assortments of at most $K$ items from a set of $N$ possible
items. The sequence can be chosen as a function of the contextual
information of items, and possibly users, in order to minimize the expected
regret, which is defined as the gap between the expected revenue generated
by the algorithm and the offline optimal expected revenue when the true
parameter is known. 
The $d$-dimensional contextual information, or a set of feature vectors,
is revealed at each round $t$, allowing the feature information of items
to change over time. 
The feedback here is the particular item chosen by the user from the
offered assortment. We assume that the item   choice follows a multinomial
logistic (MNL) distribution \cite{mcfadden1978modeling}.  
This is one of the most widely used model in dynamic assortment
optimization literature \citep{caro2007dynamic,
  rusmevichientong2010dynamic, saure2013optimal, agrawal2019mnl,
  agrawal2017thompson, aouad2018greedy}.  


\begin{table*}[t]
\vskip 0.15in
\begin{center}
\begin{sc}
\begin{tabular}{lllll}
\toprule
& Method & Context &  Regret \\
\midrule
\citet{agrawal2019mnl}$\qquad \qquad $   & UCB  & No &  $\tilde{\mathcal{O}}(\sqrt{NT})$, $\Omega(\sqrt{NT/K})$ \\
\citet{agrawal2017thompson}  & TS  & No &  $\tilde{\mathcal{O}}(\sqrt{NT})$ \\
\citet{cheung2017thompson}$\qquad \qquad$ & TS  & Yes &  $\tilde{\mathcal{O}}(d\sqrt{T})^*$ \\
\citet{chen2017note} & N/A & N/A &  $\Omega(\sqrt{NT})$ ($\equiv \Omega(\sqrt{dT})$)\\
\citet{ou2018multinomial}  & UCB  & Yes &  $\tilde{\mathcal{O}}(Kd\sqrt{T})$\\
\citet{chen2018dynamic}  & UCB  & Yes &  $\tilde{\mathcal{O}}(d\sqrt{T})$, $\Omega(d\sqrt{T}/K)$\\
\citet{oh2019thompson}  & TS  & Yes &  $\tilde{\mathcal{O}}(d\sqrt{T})^*$, $\tilde{\mathcal{O}}(d^{3/2}\sqrt{T})$\\
\textbf{This Work} (Algorithm~\ref{algo:UCB-MNL})  & UCB  & Yes &  $\tilde{\mathcal{O}}(d\sqrt{T})$\\
\textbf{This Work} (Algorithms~\ref{algo:DBL-MNL}) $\qquad$  & UCB  & Yes &  $\tilde{\mathcal{O}}(\sqrt{dT})$\\
\bottomrule
\end{tabular}
  \caption{
    Comparison of 
  regret bounds in 
  related works on MNL bandits. $T$ is the number of total rounds, $K$ is the assortment size, 
$N$ is the total number of items, and $d$ is the feature dimension. UCB
denotes upper-confidence bound and TS denotes Thompson sampling, and
starred ($^*$) regrets denote Bayesian regrets. $\tilde{\mathcal{O}}$ is a big-$\mathcal{O}$ notation up to logarithmic factors.} 
\label{sample-table}
\end{sc}
\end{center}
\vskip -0.1in
\end{table*}

For sequential decision-making with contextual information, (generalized)
linear bandits \citep{abe1999associative, auer2002using, filippi2010parametric, rusmevichientong2010linearly, abbasi2011improved,chu2011contextual, li2017provably}
and their variants have been widely studied. 
However, these methods are only limited to a single item
selection which is increasingly rarer in practice as compared to multiple item offering that we consider in this work.  
There are a line of works in combinatorial variants of contextual bandit 
problems \citep{qin2014contextual, wen2015efficient, kveton2015cascading,
  zong2016cascading} mostly with semi-bandit feedback or cascading
feedback. However, these methods do not take the user choice into
account. Hence, substitution effect is not considered. 
In contrast to these contextual bandit problems and their combinatorial
variants, in the multinomial logit (MNL) contextual bandit, the item
choice (feedback) is a function of all items in the offered
assortment. The key challenges are how to design an algorithm that offers
assortments to simultaneously learn the unknown parameter and maximize the
expected revenue through sequential interactions with users and how to
guarantee its performance. 
There has been an emerging body of literature on MNL bandits in both non-contextual and contextual settings \cite{agrawal2017thompson,agrawal2019mnl,cheung2017thompson,ou2018multinomial,chen2018dynamic,oh2019thompson}.
However, an open question in the MNL contextual bandit problem is whether one can close the gap between lower and upper bounds of regret. Often, meeting such a criterion comes at the cost of practicality. Hence, designing a practical algorithm that achieves the provable optimality becomes a greater challenge.
Our contributions are as follows:
\begin{itemize}
\item \texttt{UCB-MNL} (Algorithm~\ref{algo:UCB-MNL}) is an upper confidence
  bound based algorithm 
  for MNL contextual bandits that, to our
  knowledge, is the first polynomial time algorithm that achieves an  $N$
  independent  
  $\Tilde{\mathcal{O}}(d\sqrt{T})$ regret. This result matches the
  previous best upper bound (up to logarithmic factors). 

\item We show that
  $\Tilde{\mathcal{O}}(\sqrt{dT})$ 
  regret is achievable in the MNL contextual bandits (Theorem~\ref{thm:expected_regret_supCb-mnl}). This improves on the best previous result by $\sqrt{d}$ factor, and
  matches the lower bound for the MNL bandit problem to within logarithmic
  factor. However, the resulting algorithm is not practical
as with other provably optimal bandit algorithms that rely on
 a framework proposed in~\citet{auer2002using}.

\item \texttt{DBL-MNL} (Algorithms~\ref{algo:DBL-MNL}), a simple variant 
  of \texttt{UCB-MNL}, 
  achieves $\Tilde{\mathcal{O}}(\sqrt{dT})$
  regret when  revenue is uniform for all items --- a setting that arises in a wide range
  of applications. \texttt{DBL-MNL}  does \emph{not} rely on the framework
  in~\citet{auer2002using}, and has
  state-of-the-art
  computational  efficiency. 
  Thus, this work is the first one to provide a practical
  algorithm with provable
  $\sqrt{d}$ dependence on the dimension of the context.
\item To establish a sharper regret bound, we prove a non-asymptotic
  confidence bound for the maximum likelihood estimator of the MNL model,
  which may be of independent interest. 
\end{itemize}

\section{Problem Formulation}

\subsection{Notations}
For a vector $x \in \mathbb{R}^d$, we use $\| x \|$ to denote its $\ell_2$-norm.
The weighted $\ell_2$-norm associated with a positive-definite matrix $V$ is defined by $\| x \|_V := \sqrt{x^\top V x}$. The minimum and maximum eigenvalues of a symmetric matrix $V$ are written as $\lambda_{\min}(V)$ and $\lambda_{\max}(V)$ respectively. The trace of a matrix $V$ is $\text{trace}(V)$. 
For two symmetric matrices $V$ and $W$ of the same dimensions, $V \succeq
W$ means that $V - W$ is positive semi-definite. For a positive integer $n$, we define $[n] = \{1, 2, ... , n\}$.
Finally, we define $\mathcal{S}$ to be the set
of candidate assortments with size constraint at most $K$, i.e.
$\mathcal{S} = \{ S \subset [N]: |S| \leq K \}$. Although we treat
$\mathcal{S}$ as stationary for ease of exposition, we can allow
$\mathcal{S}$ (as well as the item set $[N]$) to change over time. 

\subsection{MNL Contextual Bandits}
The MNL contextual bandits problem is defined as follows.
The agent has a set of  $N$ distinct items.
At each round $t$, the agent observes feature vectors $x_{ti} \in \mathbb{R}^d$ for every
item $i \in [N]$.
Given this contextual information, at every round $t$, the agent
offers an assortment $S_t  = \{i_1, \ldots, i_\ell\} \in
\mathcal{S}$, $\ell \leq K$, and
observes the user purchase 
decision $c_t \in S_t \cup \{0\}$, where $\{0\}$ denotes ``outside
option'' which means the user did not choose any item offered in
$S_t$. This selection is given by a  multinomial logit (MNL) choice model
\cite{mcfadden1978modeling} 
under which
the choice probability for item $i_k \in S_t$ (and the outside option)
is defined as 
\begin{align*}
    p_{t}(i_k | S_t, \theta^*) &=
        \frac{\exp\{x_{ti_k}^\top \theta^*\}}{1 + \sum_{j \in S_t}
      \exp\{x_{tj}^\top \theta^* \} },\\
p_{t}(0 | S_t, \theta^*) &=
\frac{1}{1 + \sum_{j \in S_t}
      \exp\{x_{tj}^\top \theta^* \} }
\end{align*}
where $\theta^* \in \mathbb{R}^d$ is a time-invariant parameter unknown to
the agent.   
The choice response for each item $i_k \in S_t$ 
is defined as $y_{ti_k} 
:= \mathbb{1}(c_t = i_k) \in \{0,1\}$ and $y_{t0} := \mathbb{1}(c_t = 0)$ for the outside option. Hence the choice response variable
$y_t = (y_{t0}, y_{ti_1},  ..., y_{ti_\ell})$  
is a sample from this multinomial distribution: 
\begin{equation*} 
    y_{t} \sim \text{multinomial}\left\{1, \left(p_{t}(0|S_t, \theta^*),
      ..., p_{t}(i_{\ell}|S_t, \theta^*) \right)\right\} 
\end{equation*}
where the parameter $1$ indicates that $y_t$ is a single-trial
sample,  i.e. $y_{t0} + \sum_{k=1}^{\ell} y_{ti_k} = 1$.
For each $i \in S_t \cup \{0\}$ and $t$, we define the noise $\epsilon_{ti} := y_{ti} -
p_{t}(i|S_t, \theta^*)$. Since each 
$\epsilon_{ti}$ is a bounded random variable in $[0,1]$, $\epsilon_{ti}$ is
$\sigma^2$-sub-Gaussian with $\sigma^2 = 1/4$; however, 
$\epsilon_{ti}$ is \emph{not} independent across $i
\in S_t$ due to the substitution effect in the MNL model. 
The revenue parameter $r_{ti}$ for each item is also given at round
$t$. $r_{ti}$ is the revenue from the sale if item $i$ is sold in round
$t$. Without loss of generality, assume $| r_{ti} | \leq 1$ for all $i$
and $t$. 
Then, the expected revenue of the assortment $S_t$ is given by
\begin{equation}\label{eq:expected_revenue}
    R_t(S_t, \theta^*) = \sum_{i \in S_t} r_{ti} p_{t}( i|S_t, \theta^*)
\end{equation}
Note that for a very broad class of MNL applications, including search
ranking and media recommendation, the goal is to maximize the
click-through rate; therefore,  the item revenue is uniform.

We define $S^*_t$ to be the offline optimal assortment at time $t$ 
when $\theta^*$ is known apriori, i.e.
when the true MNL probabilities $p_{t}(i | S, \theta^*)$ are known a priori: 
\begin{equation}\label{eq:S_star}
  S^*_t = \argmax_{S \subset \mathcal{S}}{R_t(S, \theta^*)}.
\end{equation}
The learning agent does not know the value of $\theta^*$,  and therefore, can only
choose the assortment $S_t$ in period $t$ based on the choices $S_{\tau}$
for periods $\tau < t$, and the observed responses. 
We measure the performance of the agent 
by the regret $\mathcal{R}_T$ for the time horizon of $T$ periods, which is the gap between the expected revenue
generated by the assortment chosen by the agent and that of the
offline optimal assortment, i.e., 
\begin{equation*}
    \mathcal{R}_T = \mathbb{E} \left[ \sum_{t=1}^T \Big( R_t(S^*_t, \theta^*) - R_t(S_t, \theta^*) \Big) \right]
\end{equation*}
where $R_t(S^*_t, \theta^*)$ is the expected revenue corresponding to the
offline optimal assortment in period $t$, i.e., the highest revenue which
can be obtained with the knowledge of $\theta^*$. 
Hence, maximizing the cumulative expected revenue is equivalent to
minimizing the cumulative expected regret.

\subsection{MLE for Multinomial Logistic Regression}\label{sec:MLE_MNL}
We briefly discuss the  maximum likelihood estimation of the unknown parameter $\theta^*$ for the MNL model. First, recall that $y_t \in
\{0,1\}^{|S_t|+1}$ is the user choice response variable where $y_{ti}$ is the $i$-th
component of $y_t$. 
Then, the negative log-likelihood function under parameter $\theta$ is then given by $\ell_n(\theta) := 
    -\sum_{t=1}^n \sum_{i \in S_t \cup \{0\}} y_{ti} \log p_{t}(i|S_t, \theta)$
which is also known as the cross-entropy error function for the multi-class
classification problem. Taking the gradient of this negative
log-likelihood with respect to $\theta$, we obtain 
\begin{equation*}
    \nabla_{\theta} \ell(\theta) = \sum_{t=1}^n \sum_{i \in S_t} (p_{t}(i|S_t, \theta)  - y_{ti}) x_{ti}
\end{equation*}
As the sample size $n$ goes to infinity, 
the MLE $\Hat{\theta}_n$ is asymptotically according to the classical likelihood theory \citep{lehmann2006theory}, with $\Hat{\theta}_n - \theta^* \rightarrow \mathcal{N}(0, \mathcal{I}^{-1}_{\theta^*})$ where $\mathcal{I}_{\theta^*}$ is the Fisher information matrix. We show in the proof of Theorem \ref{thm:normalityMLE} that $\mathcal{I}_{\theta^*}$ is lower bounded by $\sum_t \sum_{i\in S_t} p_{t}(i|\theta^*)p_{t}(0|\theta^*) x_{ti}x_{ti}^\top$. Hence, if 
$p_{t}(i|\theta^*)p_{t}(0|\theta^*) \geq \kappa > 0$, then we can ensure that $\mathcal{I}_{\theta^*}$ is invertible and prevent asymptotic variance of $x^\top \Hat{\theta}$ from going to infinity for any $x$. 

\section{Algorithms and Main Results}
In this section, we present algorithms for the MNL contextual bandit problem and their regret bounds.

\subsection{Algorithm: \texttt{UCB-MNL}}\label{sec:ucb-mnl}
The basic idea of our first algorithm is to maintain a confidence set for the
parameter $\theta^\ast$.
The techniques of upper confidence bounds (UCB) have been widely
known to be effective in balancing the exploration and exploitation
trade-off in many bandit problems, including $K$-arm bandits
\citep{auer2002finite, LS19bandit-book}, linear bandits
\citep{auer2002using, dani2008stochastic, abbasi2011improved,
  chu2011contextual} and generalized linear bandits
\citep{filippi2010parametric, li2017provably}. 

For each round $t$, the confidence set $\mathcal{C}_t$ for  $\theta^*$ is constructed from
the feature vectors $\{x_{t' i}, i\in S_{t'}\}_{t' \leq t}$ and the
observed feedback of selected items $y_1, ..., y_{t-1}$ from all previous
rounds. Let $\hat{\theta}_t$ 
  denote the estimate 
  of the unknown
parameter $\theta^*$ after $t$ periods, and suppose we are guaranteed that
$\theta^*$ lies within the 
confidence set $\mathcal{C}_t$ centered at MLE $\hat{\theta}_t$ with
radius $\alpha_t > 0$  
with a high probability.
The radius $\alpha_t$ has to be chosen carefully:
larger 
$\alpha_t$ 
induces 
more
exploration; 
however, too large $\alpha_t$ can cause regret to increase. 
In the MNL setting,  exploitation is to offer 
$\argmax_{S \in \mathcal{S}} R_t(S, \hat{\theta}_t)$, whereas
exploration is to  choose a set $S$
that has the potential for 
high 
expected revenue $R_t(S, \theta)$ as $\theta$ varies over
$\mathcal{C}_t$. Thus, a direct way to introduce optimism, and induce
exploration, is to define an optimistic revenue for each ${N \choose K}$
assortments. This is the approach taken in~\citet{chen2018dynamic};
however, this enumeration has exponential complexity when $N$ is large and $K$ is
relatively small. We show that one can induce sufficient exploration by
defining an optimistic 
utility $z_{ti}$ for each item, and defining the optimistic revenue for any
assortment $S$ using the optimistic utility. 
\begin{equation}\label{eq:z_definition}
    z_{ti} :=  x_{ti}^\top \Hat{\theta}_{t-1} + \alpha_t \| x_{ti} \|_{V^{-1}_{t-1}}
\end{equation}
where $V_t = \sum_{t'=1}^t \sum_{i \in S_t} x_{t' i} x_{t' i}^\top \in
\mathbb{R}^{d \times d}$ is a symmetric positive definite matrix. 
The optimistic utility $z_{ti}$
consists of two components:
mean utility  estimate $ x_{ti}^\top \Hat{\theta}_{t-1}$ and standard
deviation $\alpha_t \| x_{ti} \|_{V^{-1}_{t-1}}$. 
In the proof of the regret bound of the algorithm, we show that $z_{ti}$
is, indeed, an upper bound of $x_{ti}^\top \theta^*$ if $\theta^*$ lies
within in the confidence ellipsoid centered at $\Hat{\theta}_{t-1}$.
Based on $z_{ti}$, we
construct the following optimistic estimate of the expected revenue 
\begin{equation}\label{eq:ucb_revenue}
    \Tilde{R}_t(S) := \frac{\sum_{i \in S} r_{ti}  \exp \left(z_{ti}
      \right)}{1 + \sum_{j \in S} \exp \left(z_{tj}\right)} \,. 
\end{equation} 
We assume an access to an assortment optimization method which returns the
assortment at time $t$ for a given parameter estimate, $S_t = \arg\max_{S \subset \mathcal{S}}
\Tilde{R}_t(S)$.  
There are efficient polynomial-time algorithms
available to solve this optimization problem 
\citep{rusmevichientong2010dynamic, davis2014assortment}. 
We now have all the ingredients for our algorithm,
\texttt{UCB-MNL} (see Algorithm \ref{algo:UCB-MNL}).



\begin{algorithm}[t]
\caption{\texttt{UCB-MNL}}
\begin{algorithmic}[1]
    \STATE \textbf{Input}: initialization $T_0$, confidence radius $\alpha_t$
    \STATE \textbf{Initialization}: \textbf{for} $t \in [T_0]$
    \STATE \quad Randomly choose $S_t$ with $|S_t|=K$
    \STATE \quad $V_t \leftarrow V_{t-1} + \sum_{i \in S_t} x_{ti} x_{ti}^\top$
    \FOR{all $t = T_0 + 1$ to $T$}
        \STATE Compute $z_{ti} =  x_{ti}^\top \Hat{\theta}_{t-1} + \alpha_t \| x_{ti} \|_{V^{-1}_{t-1}}$ for all $i$
        \STATE Offer $S_t = \argmax_{S \subset \mathcal{S}} \Tilde{R}_t(S)$ and observe $y_t$ 
        \STATE Update $V_t \leftarrow V_{t-1} + \sum_{i \in S_t} x_{ti} x_{ti}^\top$
        \STATE Compute MLE $\Hat{\theta}_t$ by solving\\
        $\sum_{t'= 1}^{t} \sum_{i \in S_{t'}} \big(  p_{t'}(i|S_{t'}, \Hat{\theta}_{t}) - y_{t' i} \big) x_{t' i} = \vec{0}$
    \ENDFOR
\end{algorithmic}
\label{algo:UCB-MNL}
\end{algorithm}

In Algorithm \ref{algo:UCB-MNL}, during the initialization phase, we first
randomly choose an assortment $S_t$ with exactly $K$ items 
(after initialization, $S_t$ can be smaller than $K$) to ensure a unique
MLE solution.  
The initialization $T_0$, specified in Theorem
\ref{thm:expected_regret_ucb-mnl}, is chosen to ensure that
$\lambda_{\min}(V_{T_0})$ is large enough. 

\subsection{Regret Bound for \texttt{UCB-MNL} Algorithm}\label{sec:regret_ucb-mnl}
We present the regret upper-bound of \texttt{UCB-MNL}  under the following assumptions on the
context process and the MNL model, both standard in the literature.

\begin{assumption}\label{assum:x_bound}
Each feature vector $x_{ti}$ is drawn i.i.d. from an unknown distribution $p_x$, with $\| x_{ti} \| \leq 1$ all $t, i$ and there exists a constant $\sigma_0 > 0$ such that $\mathbb{E}[x_{ti} x_{ti}^\top] \geq \sigma_0$.
\end{assumption}

The boundedness is used to make the regret bounds scale-free. The i.i.d. assumption is also made in generalized linear bandit \cite{li2017provably} and MNL contextual bandit  \cite{chen2018dynamic,oh2019thompson} literature.


\begin{assumption}\label{assum:prob_bound}
There exists $\kappa > 0$ such that for every item $i \in S$ and any $S \in \mathcal{S}$ and all round $t$, $\min_{\|\theta - \theta^*\| \leq 1} p_t(i|S, \theta)p_t(0|S, \theta) \geq \kappa$.
\end{assumption}

The asymptotic normality of MLE implies the necessity of this assumption. This is a standard assumption in MNL contextual bandits \cite{cheung2017thompson,chen2018dynamic,oh2019thompson}, which is also equivalent to the standard assumption for the link function in generalized linear contextual bandits \citep{filippi2010parametric, li2017provably} to ensure the Fisher information matrix is invertible.

\begin{theorem}[Regret of \texttt{UCB-MNL}]\label{thm:expected_regret_ucb-mnl}
Suppose Assumptions~\ref{assum:x_bound} and \ref{assum:prob_bound} hold and we run \texttt{UCB-MNL} with confidence width $\alpha_t = \frac{1}{2\kappa}\sqrt{2d \log\left( 1 + \frac{t}{d} \right) + 2\log t }$ and $T_0 = \Ocal(\max\{ \kappa^{-2} \left( d \log(T/d) + 4\log T \right), K/\sigma^2  \})$. Then the expected regret of \texttt{UCB-MNL} is upper-bounded by
\begin{align*}
      \mathcal{R}_T 
      &= \mathcal{O}\left(  d\sqrt{ T  \log \left(  1 + T/d \right) \log(T/d)}  \right).
\end{align*}
\end{theorem}

\textbf{Discussion of Theorem \ref{thm:expected_regret_ucb-mnl}.} In terms
of key problem primitives, Theorem \ref{thm:expected_regret_ucb-mnl}
demonstrates  $\Tilde{\mathcal{O}}(d\sqrt{T})$ regret bound for
\texttt{UCB-MNL} which is independent of $N$; hence, it is applicable to
the case with a very large number of candidate items. 
\citet{chen2018dynamic} established the lower bound result
$\Omega(d\sqrt{T}/K)$ for MNL bandits. When $K$ is small, which is
typically true in many applications, the regret upper-bound in Theorem
\ref{thm:expected_regret_ucb-mnl} demonstrates that \texttt{UCB-MNL} is
almost optimal. The established regret of \texttt{UCB-MNL} improves the
previous worst-case regret bound of \citet{oh2019thompson} by $\sqrt{d}$
factor and that of \citet{chen2018dynamic} in both logarithmic and additive
factors. Moreover, although having the same rate of
$\Tilde{\mathcal{O}}(d\sqrt{T})$  regret up to logarithmic factors, the
UCB method in \citet{chen2018dynamic} has exponential computational cost,
since it needs to enumerate  all of the possible ($N$ choose $K$)
assortments. Therefore, 
\texttt{UCB-MNL} is the first polynomial-time algorithm that achieves
$\Tilde{\mathcal{O}}(d\sqrt{T})$ worst-case regret. 

\textbf{Extension to online parameter update.} 
\texttt{UCB-MNL} is simple to implement and works very well in practice. 
We further improve both the time and space complexities of the algorithm
by using an online parameter update version (Algorithm~3 in the appendix). Exploiting the fact
that the loss for the MNL model is strongly convex over bounded domain, we
apply a variant of the online Newton step inspired by
\citet{hazan2014logistic,zhang2016online} to find an approximate solution
rather than computing the exact MLE. We show that the modified algorithm
still enjoys the same order of the statistical efficiency with
$\Tilde{\mathcal{O}}(d\sqrt{T})$ regret even with the online update. 
\begin{corollary}\label{cor:online_update_regret}
\texttt{UCB-MNL} with online parameter update still has $\Tilde{\mathcal{O}}(d\sqrt{T})$ regret.
\end{corollary}

\subsection{Non-asymptotic Normality of the MLE for\\the MNL Model}

We have shown that \texttt{UCB-MNL} is both statistically and
computationally efficient. The algorithm also shows state-of-the-art  practical performances 
as we report later in the numerical experiments.
However, the regret bound in Theorem~\ref{thm:expected_regret_ucb-mnl} 
has a linear 
dependence on feature dimension $d$ and, therefore, is not very attractive
when the feature vectors are high dimensional.  
We next investigate whether a sublinear dependence on $d$ is possible.
In the regret analysis for \texttt{UCB-MNL}, we upper-bound the prediction error
 $x^\top (\theta^*-\hat{\theta}_t)$ 
 using H\"older's inequality, $|x^\top
\hat{\theta}_t - x^\top \theta^*| \leq \| x \|_{V^{-1}_t} \|
\hat{\theta}_t - \theta^*\|_{V_t}$, 
where we show each of the terms on the right hand side is bounded by
$\tilde{\mathcal{O}}(\sqrt{d})$, hence resulting in a linear dependence on $d$ when
combined. A potential solution to circumvent this challenge is to control the
prediction error directly without bounding two terms separately.

In Theorem~\ref{thm:normalityMLE}, we propose a  non-asymptotic normality
bound for the 
MLE for the MNL model
in order to establish a sharper concentration result for $|x^\top (\hat{\theta}_t -\theta^*)|$. This is a generalization of Theorem~1 in
\citet{li2017provably} to the MNL model. 
To the best of  
our knowledge, there was no existing finite-sample normality results
for the prediction error of the utility for the MNL model. This concentration result can
be of independent interest beyond the bandit problem we address in this work.

\begin{theorem}[Non-asymptotic normality of MLE]\label{thm:normalityMLE}

Suppose we have independent responses $y_1, ..., y_n$
conditioned on feature vectors $\{x_{t i}\}_{t=1,i=1}^{n,K}$.
Define $V_n = \sum_{t=1}^n \sum_{i \in S_{t}} x_{t i} x_{t i}^\top$, and let $\delta > 0$ be
given. Furthermore, assume that 
$\lambda_{\min}(V_n) \geq \max\!\left\{\frac{9 \Dcal^4}{\kappa^4 \log(1/\delta)}, \frac{144  \Dcal^2}{\kappa^4} \right\}$ where
$\Dcal := \min\!\left\{4\sqrt{2d + \log \frac{1}{\delta}},  \sqrt{ d \log\!\left(n/d \right)   + 2\log  \frac{1}{\delta}} \right\}$.
    Then, for any $x \in \mathbb{R}^d$, the maximum likelihood estimator $\hat{\theta}_n$ of the MNL model satisfies with probability at least $1 - 3\delta$ that
\begin{equation*}
    | x^\top \Hat{\theta}_n - x^\top \theta^* | \leq \frac{5}{\kappa} \sqrt{\log \frac{1}{\delta} } \|x\|_{V^{-1}_n} \,.
\end{equation*}
\end{theorem}
Hence, 
the prediction error can be bounded by $\tilde{\mathcal{O}}(\sqrt{d})$
with high probability as long as the conditions on independence of samples and the
minimum eigenvalue are satisfied. 
Note that although the statement of Theorem~\ref{thm:normalityMLE} is
similar to that of the generalized linear model version in
\citet{li2017provably}, the extension to the MNL model is non-trivial because
choice probability for any given item $i \in S_t$ is function of the all the items in
the assortment 
$S_t$, and hence the analysis is much more involved.
Theorem~\ref{thm:normalityMLE} implies that we can control the behavior of
the MLE in every direction allowing us to handle the prediction error in a
tighter fashion.

\subsection{Provably Optimal but
    Impractical}\label{sec:optimality_v_practicality} 
Unfortunately, we cannot directly apply the tight bound for the MLE shown
in Theorem~\ref{thm:normalityMLE} to \texttt{UCB-MNL} since
Theorem~\ref{thm:normalityMLE} requires independent samples (as well as
the minimum eigenvalue being large enough, but this condition can be
satisfied by initial exploration). 
\texttt{UCB-MNL} is not guaranteed to produce independent
samples 
since the algorithm chooses assortments based on previous
observations, causing dependence between collected samples. 
This issue can be 
handled 
by generating independent samples
using a
framework in \citet{auer2002using}, which we denote as ``Auer-framework.'' This Auer-framework 
has been previously used in 
several 
variants of (generalized) linear bandits \cite{chu2011contextual,li2017provably, zhou2019learning}. 
We show that the adaptation of the Auer-framework to the MNL contextual bandit problem is possible\footnote{We defer the details of the algorithm to the appendix since this is not the focus of the paper.} and 
establish the following regret bound.

\begin{theorem}[Provably optimal regret]\label{thm:expected_regret_supCb-mnl}
Suppose Assumptions~\ref{assum:x_bound} and \ref{assum:prob_bound} hold. There exists an algorithm which establishes $\tilde{\Ocal}(\sqrt{dT})$ regret for the MNL contextual bandits.
\end{theorem}

$\Omega(\sqrt{NT})$ lower bound was shown in \citet{chen2017note}  for the
non-contextual MNL bandits. This lower bound can be translated to
$\Omega(\sqrt{dT})$ if each item is represented as one-hot encoding. Hence
the regret bound in Theorem \ref{thm:expected_regret_supCb-mnl} matches the lower
bound for the MNL bandit problem with finite items.  To our knowledge,
this is the first result that achieves the rate of
$\Tilde{\mathcal{O}}(\sqrt{dT})$ regret and establishes the provable
optimality in the MNL contextual bandit problem. However, this comes at a
cost. 
The algorithm based on the Auer-framework, although
provably optimal, is not practical (see the numerical
experiments)! In fact, this is true for
\emph{all} optimal methods \cite{chu2011contextual, li2017provably, zhou2019learning}
that rely on the Auer-framework~\cite{auer2002using} 
because the framework  
wastes too many samples with random exploration.\footnote{These previous
methods~\cite{chu2011contextual, li2017provably, zhou2019learning} that
use techniques in \cite{auer2002using}  do not provide numerical evaluations.}
Next, we investigate whether 
$\tilde{\Ocal}(\sqrt{dT})$ regret can be achieved in a practical manner 
for the class of applications where the revenue for each item is uniform. As discussed
earlier that this class includes web search and media recommendations.

\begin{algorithm}[t]
\caption{\texttt{DBL-MNL}}
\begin{algorithmic}[1]
    \STATE \textbf{Input}: sampling parameter $q_k$, confidence radius $\beta_k$
    \STATE Set $\tau_1 \leftarrow d$, $t \leftarrow 1$, $V_0 \leftarrow \vec{0}_{d \times d}$
    \STATE \textbf{Initialization}: \textbf{for} $t \in [d]$
    \STATE \quad Randomly choose $S_t  \in \mathcal{S}$ with $|S_t|=K$
    \STATE \quad $V_t \leftarrow V_{t-1} + \sum_{i \in S_t} x_{ti} x_{ti}^\top$
    \FOR{each episode $k = 2, 3,...$}
    \STATE Set the last round of $k$-th episode: $\tau_k \leftarrow 2^{k-1}$
    \STATE Compute MLE $\Hat{\theta}_k$ by solving\\ $\sum_{t = \tau_{k-2}+1}^{\tau_{k-1}} \sum_{i \in S_t} \big(  p_{t}(i| S_t, \Hat{\theta}_k) - y_{t i} \big) x_{t i} = \vec{0}$
    \STATE Update $W_{k-1} \leftarrow V_{\tau_{k-1}+1}$; Reset $V_{\tau_{k-1}+1} \leftarrow 0_{d \times d}$
    \FOR{each round $t = \tau_{k-1}+1, ..., \tau_k$}
    \IF{$\tau_k - t \leq q_k$ and $\lambda_{\min}(V_t) \leq \frac{K q_k \sigma_0}{2}$}
        \STATE Randomly choose $S_t \in \mathcal{S}$ with $|S_t|=K$
    \ELSE
        \STATE Offer $S_t = \argmax_{S \in \mathcal{S}} \tilde{R}_t(S)$ 
    \ENDIF
    \STATE Update $V_{t+1} \leftarrow V_t + \sum_{i \in S_t} x_{ti} x_{ti}^\top$
    \ENDFOR
    \ENDFOR
\end{algorithmic}
\label{algo:DBL-MNL}
\end{algorithm}


\subsection{Algorithm: \texttt{DBL-MNL}}
We propose a new algorithm, \texttt{DBL-MNL}
(Algorithm~\ref{algo:DBL-MNL}) that 
is \emph{both} provably optimal and practical.
\texttt{DBL-MNL} operates in an episodic manner. At the beginning of each
episode, the MLE is computed using the samples from a previous episode. 
Within an episode, the parameter is not updated, but the algorithm takes
an UCB action based on the parameter computed at the beginning of the
episode. In particular, for round $t$ in the $k$-th episode, the
upper-bound of an utility estimate is computed as 
\begin{align*}
    &\tilde{z}_{ti} =  x_{ti}^\top \Hat{\theta}_k + \alpha_k \| x_{ti} \|_{W_{k-1}^{-1}} \notag\\ &\text{where } W_{k-1} = \sum_{t'= \tau_{k-1}+1}^{\tau_{k-1}} \sum_{i \in S_{t'}} x_{t' i} x_{t' i}^\top 
\end{align*}
and $\tau_{k-1}$ is the last round of the $k-1$-th episode. Note that the
Gram matrix resets every episode. 
Under this action selection, samples within each
episode are independent of each other. Episode lengths are doubled over
time such that the length of the $k$-th episode is twice as large as the
$k-1$-th episode. This doubling technique is inspired by
\citet{jaksch2010near, javanmard2019dynamic}.
Towards the end
of each episode, the algorithm checks whether $\lambda_{\min}(V_t)$ is
suitably large. If not, it performs random exploration. 
Since episode lengths are growing exponentially and the threshold for
$\lambda_{\min}(V_t)$ is only logarithmic in $t$, even in the worst case, the
algorithm draws $\Ocal(\log T)$ random samples. 
Note that the algorithm may not even take these
exploratory actions since $\lambda_{\min}(V_t)$ may already surpass the
threshold for large enough episodes (this is clearly observed in numerical
evaluations). This makes \texttt{DBL-MNL} much more
practical since it would perform minimal random exploration. Furthermore,
the algorithm is computationally efficient with only logarithmic number of
parameter updates instead of updating in every period.

\subsection{Regret Bound of \texttt{DBL-MNL}}

We analyze the regret of \texttt{DBL-MNL} for which we aim to establish $\tilde{\Ocal}(\sqrt{dT})$ regret. For our analysis, we add the following mild assumption which encompasses many canonical distributions.

\begin{assumption}[Relaxed symmetry]\label{assum:rho}
For a joint distribution $p_X$, there exists $\rho_0 < \infty$ such that $\frac{p_X(-x)}{p_X(x)} \leq \rho_0$ for all $x$.
\end{assumption}

This assumption is also used in the analysis of sparse bandits \citet{oh2020sparsity}.
Assumption~\ref{assum:rho} states that the joint distribution $p_\mathcal{X}$ can be skewed but this skewness is bounded. For symmetrical distributions, $\rho_0 = 1$. One can see that a large class of continuous and discrete distributions satisfy Assumption~\ref{assum:rho}, e.g., Gaussian, truncated Gaussian, uniform distribution, and Rademacher distribution, and many more. Under this suitable regularity, we establish the following regret bound for \texttt{DBL-MNL}.

\begin{theorem}[Regret bound of \texttt{DBL-MNL}]\label{thm:expected_regret_DBL-MNL}
Suppose Assumptions~\ref{assum:x_bound}-\ref{assum:rho} hold and the
revenue $r_i \equiv r$ is uniform. Then the expected regret of \texttt{DBL-MNL}
over horizon  $T$  is  
$\Rcal_T = \Ocal\big( \sqrt{d T \log \left( T/d\right) \log(T N) \log(T) } \big)$.
\end{theorem}

\textbf{Discussion of Theorem~\ref{thm:expected_regret_DBL-MNL}.}
\texttt{DBL-MNL} achieves $\tilde{\Ocal}(\sqrt{dT})$ regret when the
revenue for each item is uniform. This encompasses all applications where the
goal is to maximize an appropriate ``click-through rate'' from offering
the assortment.
Theorem~\ref{thm:expected_regret_DBL-MNL} provides insights beyond the
MNL contextual bandits: it shows that under the suitable regularity
condition, it is possible for a practical algorithm to attain
$\tilde{\Ocal}(\sqrt{dT})$ regret. We expect this technique to yield
practical provably optimal algorithms for other variants of contextual bandit problems. 
The regret bound of \texttt{UCB-MNL} is $N$ independent; in
contrast,  \texttt{DBL-MNL} 
has a logarithmic dependence on
$N$ 
(as is common for $\tilde{\Ocal}(\sqrt{dT})$
regret algorithms). 
In fact, the numerical experiments suggest
that 
performance does have at least logarithmic dependence on $N$ for all methods 
(as indicated by 
Theorem~\ref{thm:expected_regret_DBL-MNL} for \texttt{DBL-MNL}).

\subsection{Proof Outline of Theorem~\ref{thm:expected_regret_DBL-MNL}}\label{sec:proof-outline-dbl-mnl}

Since the length of an episode grows exponentially, the number of episodes up to round $T$  is logarithmic in $T$. In particular, the $T$-th round belongs to the $L$-th episode with $L = \lfloor \log_2 T \rfloor + 1$. 
Let $\Tcal_k := \{\tau_{k-1}+1, ..., \tau_k\}$ denote an index set of rounds that belong to the $k$-th episode. 
Note that the length of the $k$-th episode is $|\Tcal_k| = \tau_k/2$. 
Then, we let $\texttt{Reg}(k\text{-th episode})$ denote the cumulative regret of the $k$-th episode, i.e., 
\begin{align*}
    \texttt{Reg}(k\text{-th episode}) := \mathbb{E} \left[ \sum_{t \in \Tcal_k} \Big( R_t(S^*_t, \theta^*) - R_t(S_t, \theta^*) \Big) \right]
\end{align*}
so that the cumulative expected regret over $T$ rounds is $\mathcal{R}(T) = \sum_{k=1}^L \texttt{Reg}(k\text{-th episode})$. 
Therefore, it suffices to bound each $\texttt{Reg}(k\text{-th episode})$.
Now, for each episode $k \in [L]$, we consider the following two cases.
\begin{enumerate}[(i)]
    \item $|\Tcal_k| \leq q_k$: In this case, the length of an episode is not large enough to have the concentration of the prediction error due to the failure of ensuring the lower bound on $\lambda_{\min}(V_t)$. Therefore, we cannot control the regret in this case. However, the total number of such rounds is only logarithmic in $T$, hence the regret corresponding to this case contributes minimally to the total regret.
    \item $|\Tcal_k| > q_k$: We can apply the fast convergence result in Theorem~\ref{thm:normalityMLE} as long as the lower bound on $\lambda_{\min}(V_t)$ is guaranteed --- note that the independence condition is already satisfied since samples in each episode are independent of each other. We show that $\lambda_{\min}(V_t)$ grows linearly as $t$ increases in each episode with high probability.
    In case of $\lambda_{\min}(V_t)$ not growing as fast as the rate we require, we perform random sampling to satisfy this criterion towards the end of each episode. Therefore, with high probability, the lower bound on $\lambda_{\min}(V_t)$ is satisfied.
\end{enumerate}

\begin{figure*}[t]
\centering
\begin{subfigure}[b]{0.31\textwidth}
    \includegraphics[width=\textwidth]{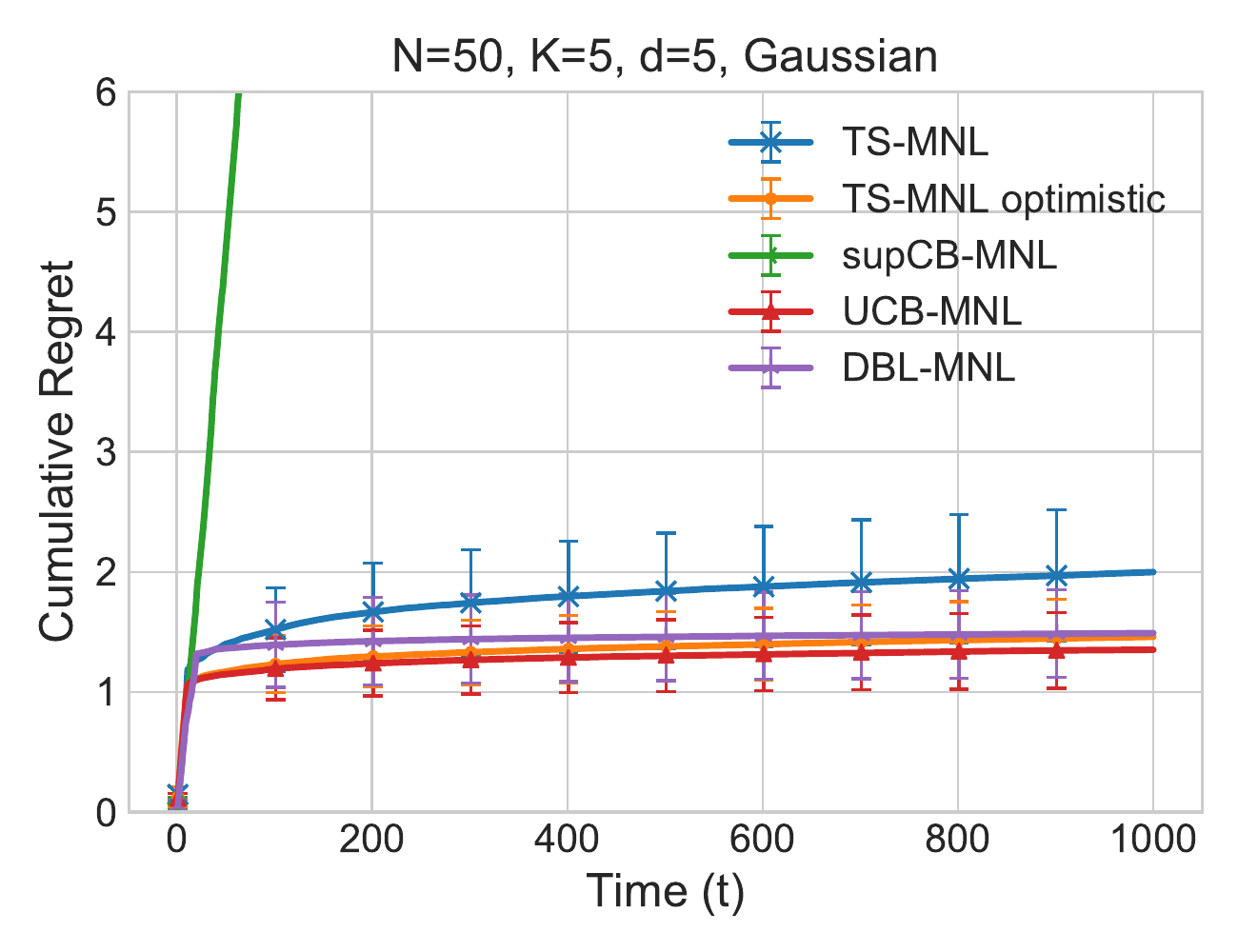}
\end{subfigure}
\begin{subfigure}[b]{0.31\textwidth}
    \includegraphics[width=\textwidth]{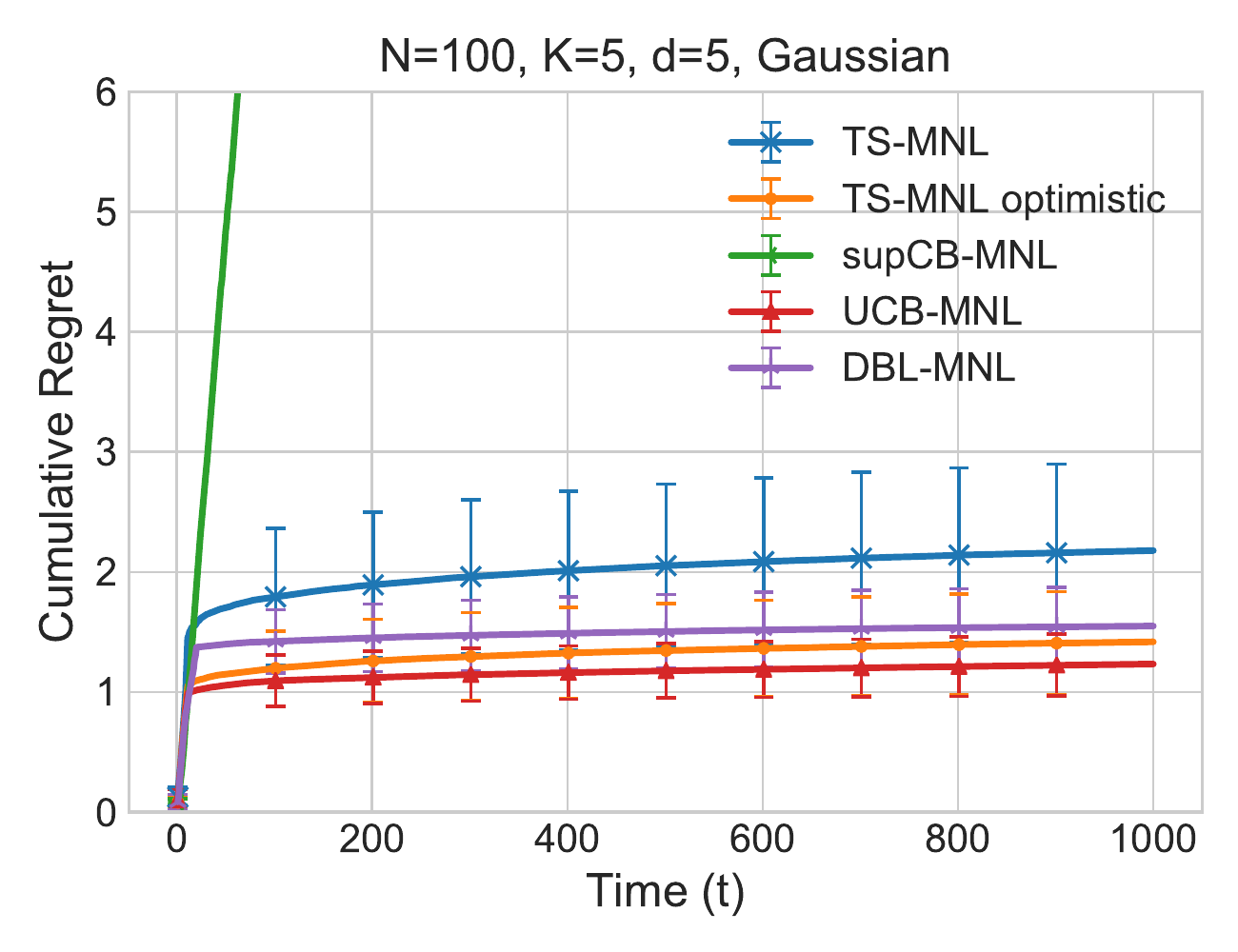}
\end{subfigure}
\begin{subfigure}[b]{0.31\textwidth}
    \includegraphics[width=\textwidth]{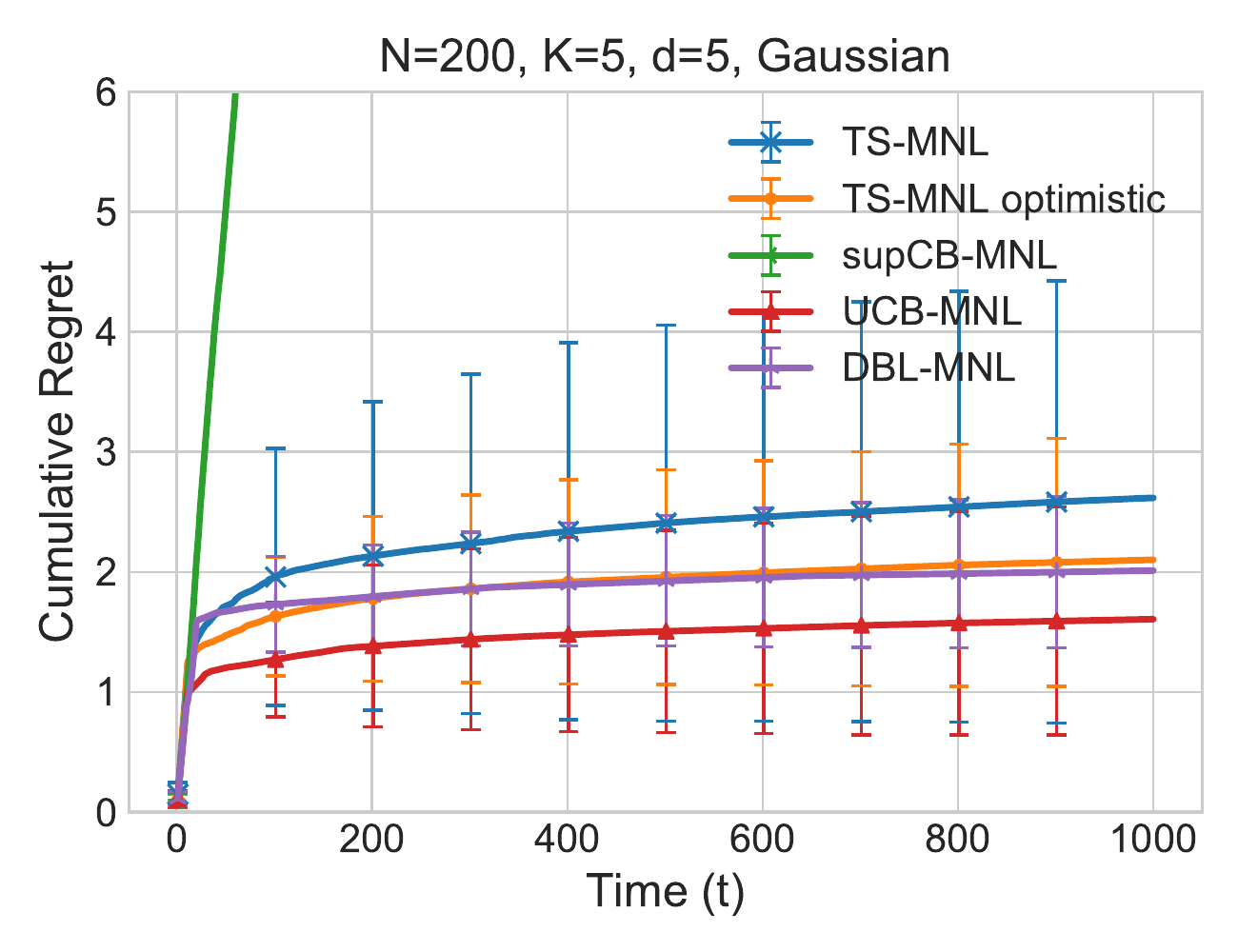}
\end{subfigure}\\
\begin{subfigure}[b]{0.31\textwidth}
    \includegraphics[width=\textwidth]{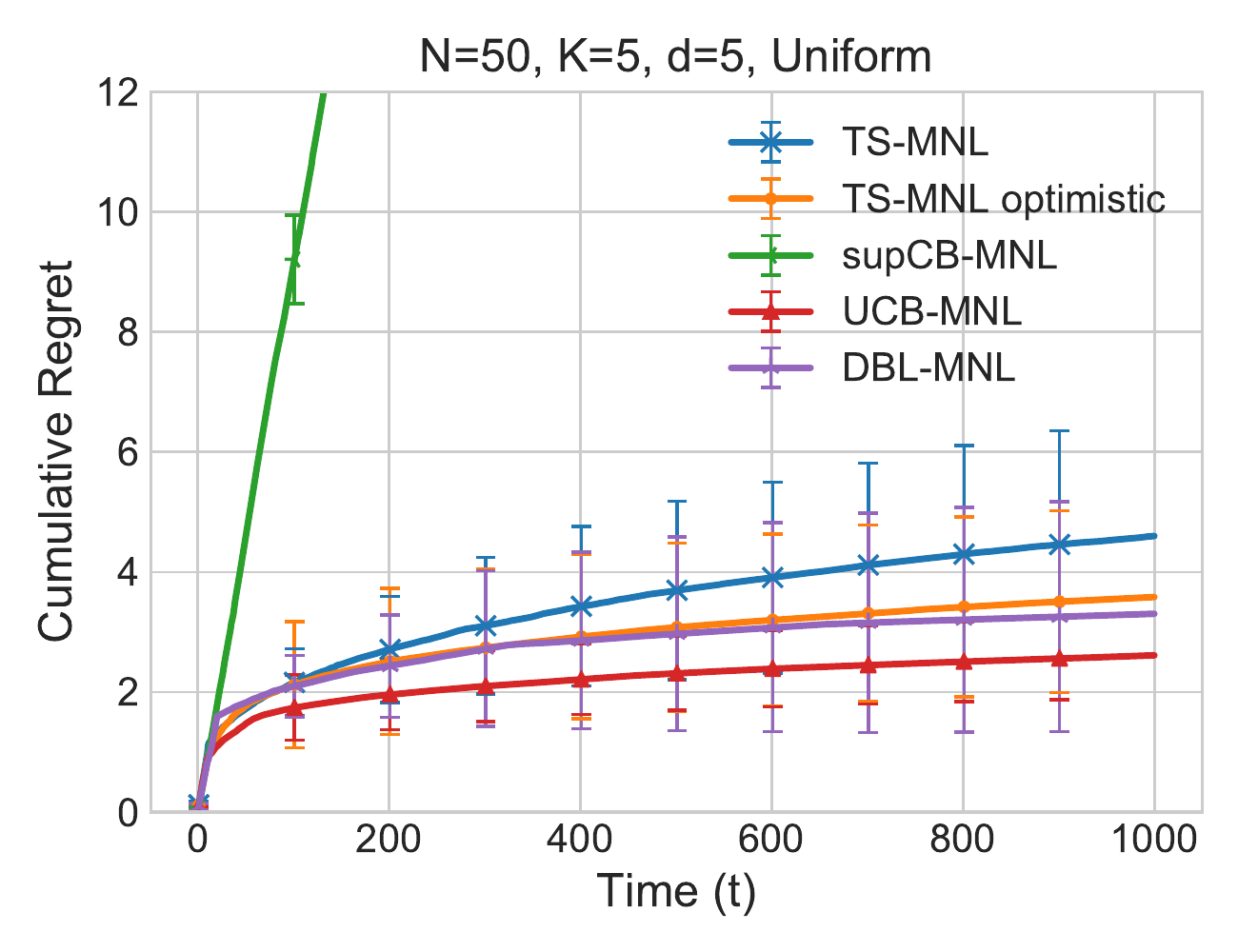}
\end{subfigure}
\begin{subfigure}[b]{0.31\textwidth}
    \includegraphics[width=\textwidth]{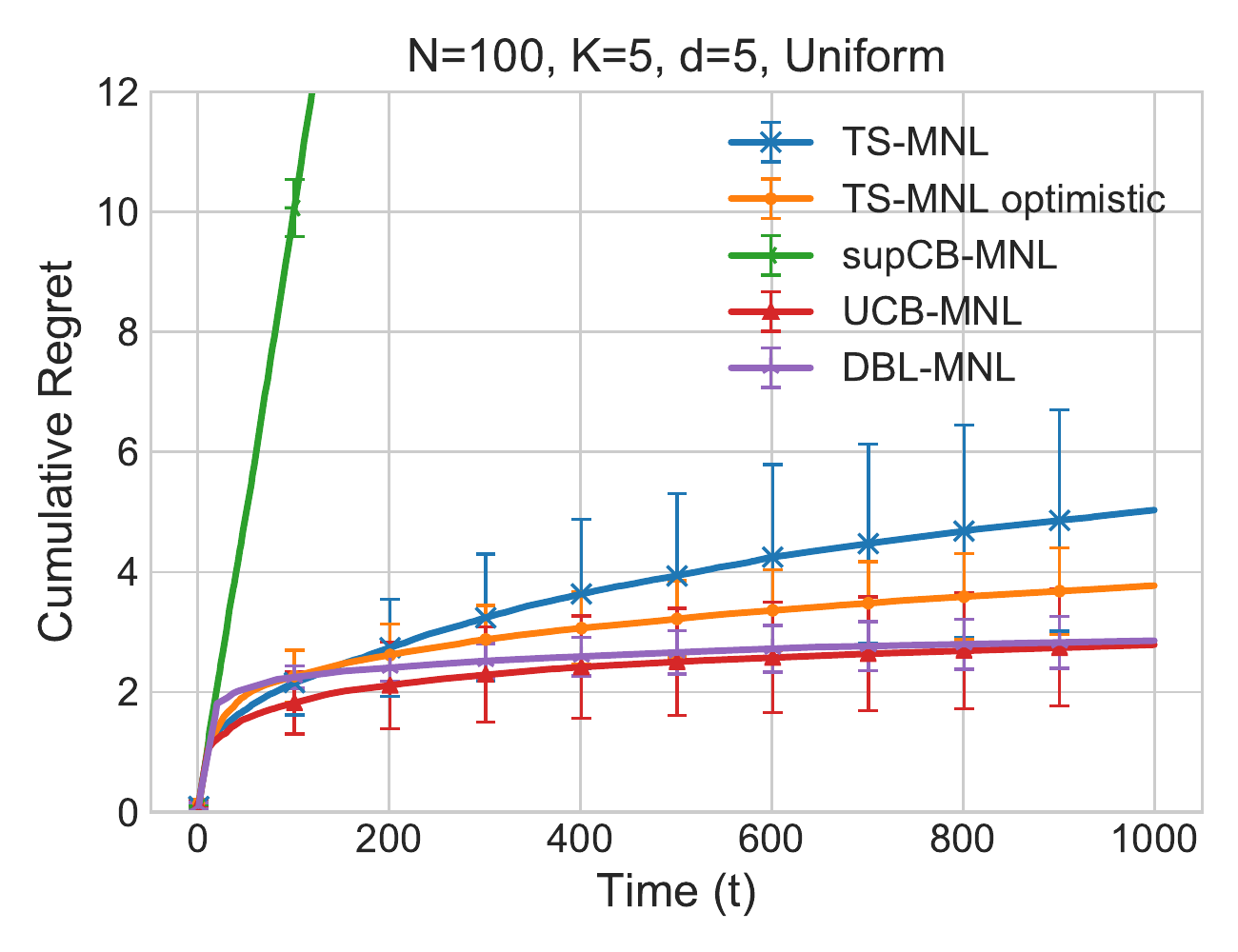}
\end{subfigure}
\begin{subfigure}[b]{0.31\textwidth}
    \includegraphics[width=\textwidth]{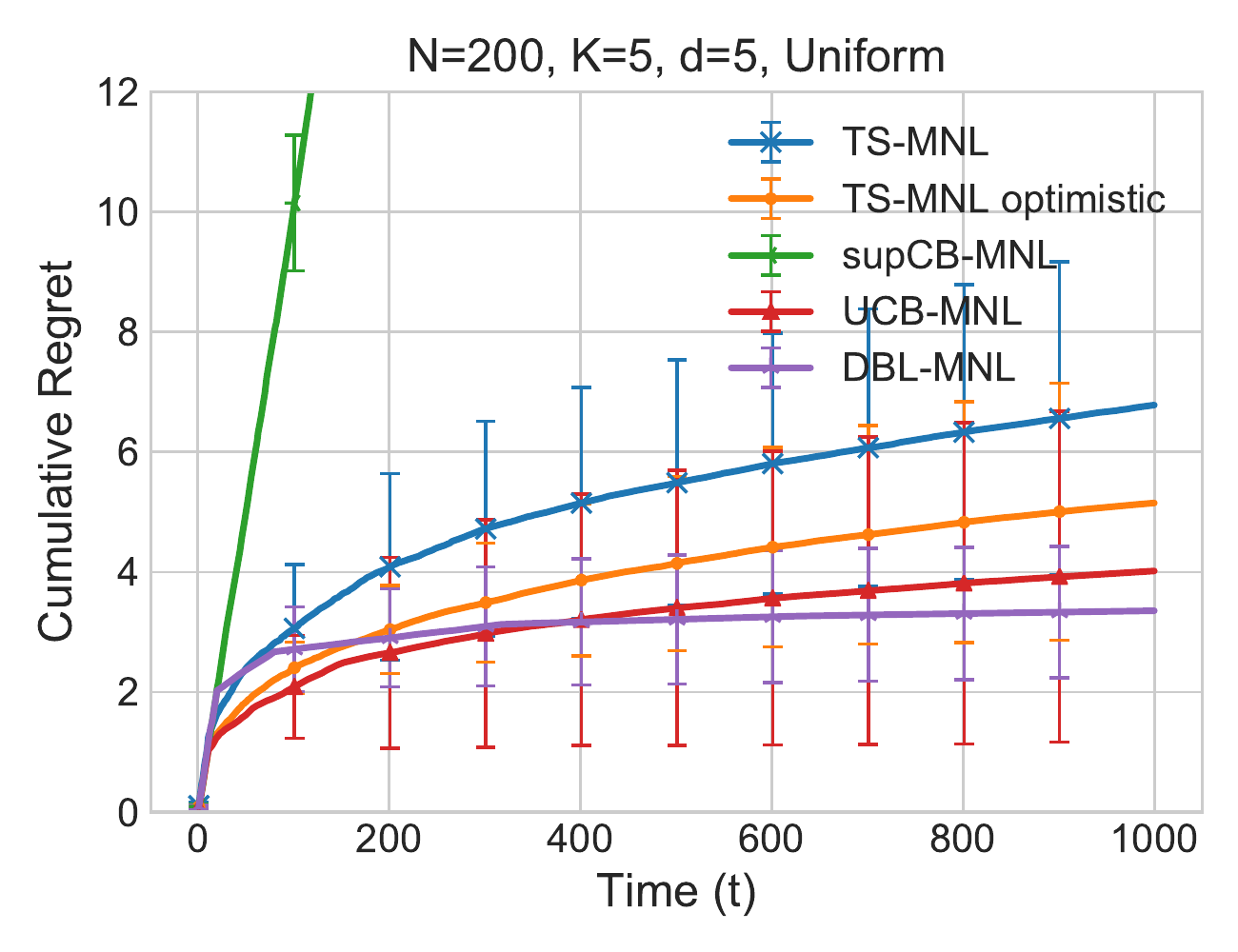}
\end{subfigure}
\vspace{-0.3cm}
\caption{The regret plots show that \texttt{UCB-MNL} and \texttt{DBL-MNL} perform at start-of-the-art levels across different problem instances. Evaluations are for features drawn from a multivariate Gaussian (first row) and uniform (second row) distributions.}
\label{fig:mnl_regret}
\vspace{-0.3cm}
\end{figure*}


For case (i), clearly $q_k \leq q_L$ for any $k \in \{1, ..., L\}$. 
$|\Tcal_k|$ eventually grows to be larger than $q_L$ for some $k$ since $q_L$ is logarithmic in $T$. Let $k'$ be the first episode such that $|\Tcal_{k'}| \geq q_L$. Hence, $|\Tcal_{k'}| \leq 2 q_L$. Thus, the cumulative regret prior to the $k'$-th episode is $\Ocal\!\left( \log d + d^2 + \log^2(TN) \right)$.
Then, letting $k''$ be the first episode such that $|\Tcal_{k''}| \geq q_{k''}$ and noting that $k'' \leq k'$ gives
\begin{align*}
    \sum_{k=1}^{k''-1} \texttt{Reg}(k\text{-th episode}) \leq \sum_{k=1}^{k'-1} \texttt{Reg}(k\text{-th episode}) \,.
\end{align*}
Hence, the cumulative regret corresponding to case (i) is at most poly-logarithic in $T$.

For case (ii), it suffices to show random sampling ensures the growth of $\lambda_{\min}(V_t)$. 
We show that random sampling with duration $q_k$ specified in Theorem~\ref{thm:expected_regret_DBL-MNL} ensures the minimum eigenvalue condition for the Gram matrix, i.e., $\lambda_{\min}(V_{\tau_k}) \geq \max\!\left\{\frac{9 \Dcal_k^4}{\kappa^4 \log(\tau_k N/2)}, \frac{144  \Dcal_k^2}{\kappa^4} \right\} $ with high probability 
for each episode $k \in [L]$.
We then apply the confidence bound in Theorem~\ref{thm:normalityMLE} to the $k$-th episode which requires samples in the $(k-1)$-th episode are independent and $\lambda_{\min}(V_{\tau_{k-1}})$ at the end of the $(k-1)$-th episode is large enough.
That is, with a lower bound guarantee on $\lambda_{\min}(V_{\tau_{k-1}})$  and the fact that samples are independent of each other in each episode, we have with high probability
\begin{align*}
    | x_{ti}^\top(\hat{\theta}_k - \theta^*)| \leq \beta_k \|x_{ti}\|_{W^{-1}_{k-1}}, \enskip \forall i \in [N], \forall t \in \Tcal_k 
\end{align*}
with suitable confidence width $\beta_k$ specified in Theorem~\ref{thm:expected_regret_DBL-MNL}.
Therefore, the expected regret in the $k$-th episode can be bounded by $\tilde{\Ocal}(\sqrt{d \tau_k})$. Then we combine the results over all episodes to establish $\tilde{\Ocal}(\sqrt{d T})$ regret.

\section{Numerical Experiments}\label{sec:numerics}

In this section, we evaluate the performances of our proposed algorithms: \texttt{UCB-MNL} (Algorithm~\ref{algo:UCB-MNL}) and \texttt{DBL-MNL} (Algorithm~\ref{algo:DBL-MNL}) in numerical experiments.
In our evaluations, we report the cumulative regret for each round $t \in \{1, ..., T\}$. 
For each experimental configuration, we evaluate the algorithms on 20 independent instances and report average performances. 
In each instance, the underlying parameter $\theta^*$ is sampled from the $d$-dimensional uniform distribution, with each element of $\theta^*$  uniformly distributed in $[0,1]$. The underlying parameters are fixed during each problem instance but not known to the algorithms. 
For efficient evaluations, we consider uniform revenues, i.e., $r_{ti} = 1$ for all $i$ and $t$. Therefore, the combinatorial optimization step to solve for the optimal assortment reduces to sorting items according to their utility estimate. Also, recall that the regret bound for \texttt{DBL-MNL} (Theorem~\ref{thm:expected_regret_DBL-MNL}) is derived under the uniform revenue assumption, therefore, the uniform revenue setting provides a suitable test bed for all methods considered in this section.

\begin{table}[t]
\centering
\begin{tabular}{crr}\\\toprule  
 &  \multicolumn{2}{c}{Horizon ($T$)}\\
Method & $1000$ & $5000$ \\\midrule
\texttt{TS-MNL} \cite{oh2019thompson} & 6.65 & 73.99\\  \midrule
\texttt{TS-MNL Opt.} \cite{oh2019thompson} & 6.81 & 77.18\\  \midrule
\texttt{UCB-MNL} (Algorithm~\ref{algo:UCB-MNL}) & 6.62 & 74.28\\  \midrule
\texttt{DBL-MNL} (Algorithm~\ref{algo:DBL-MNL}) &\textbf{1.20} & \textbf{5.92}\\  \bottomrule
\end{tabular}
\caption{Runtime evaluation (sec),
 $N=100, K=5, d=5$}\label{tab:runtime}
\end{table} 

We compare the performances of the proposed algorithms with those of the
state-of-the-art Thompson sampling based algorithms, \texttt{TS-MNL} and
``optimistic'' \texttt{TS-MNL}, proposed in \citet{oh2019thompson}. Additionally, we evaluate the performance of the provably optimal but impractical algorithm, \texttt{supCB-MNL} (see Algorithm~5 in the appendix), that is based on the Auer-framework. 
Figure~\ref{fig:mnl_regret} shows that
the performances of \texttt{UCB-MNL} and \texttt{DBL-MNL} 
are superior to or comparable to the state-of-the-art Thompson sampling
methods. 
Moreover, the runtime evaluation shows that   \texttt{DBL-MNL} is significantly faster than the other methods due to its logarithmic number of parameter updates.



\section{Ethical Statement}
We conform that our work meets the standards listed in the ethics and malpractice statement of the AAAI.

\bibliography{refs.bib}

\begin{thebibliography}{44}
\providecommand{\natexlab}[1]{#1}
\providecommand{\url}[1]{\texttt{#1}}
\providecommand{\urlprefix}{URL }
\expandafter\ifx\csname urlstyle\endcsname\relax
  \providecommand{\doi}[1]{doi:\discretionary{}{}{}#1}\else
  \providecommand{\doi}{doi:\discretionary{}{}{}\begingroup
  \urlstyle{rm}\Url}\fi

\bibitem[{Abbasi-Yadkori, P{\'a}l, and
  Szepesv{\'a}ri(2011)}]{abbasi2011improved}
Abbasi-Yadkori, Y.; P{\'a}l, D.; and Szepesv{\'a}ri, C. 2011.
\newblock Improved algorithms for linear stochastic bandits.
\newblock In \emph{Advances in Neural Information Processing Systems},
  2312--2320.

\bibitem[{Abe and Long(1999)}]{abe1999associative}
Abe, N.; and Long, P.~M. 1999.
\newblock Associative reinforcement learning using linear probabilistic
  concepts.
\newblock In \emph{International Conference on Machine Learning}, 3--11.

\bibitem[{Agrawal et~al.(2016)Agrawal, Avadhanula, Goyal, and
  Zeevi}]{agrawal2016near}
Agrawal, S.; Avadhanula, V.; Goyal, V.; and Zeevi, A. 2016.
\newblock A near-optimal exploration-exploitation approach for assortment
  selection.
\newblock In \emph{Proceedings of the 2016 ACM Conference on Economics and
  Computation}, 599--600.

\bibitem[{Agrawal et~al.(2017)Agrawal, Avadhanula, Goyal, and
  Zeevi}]{agrawal2017thompson}
Agrawal, S.; Avadhanula, V.; Goyal, V.; and Zeevi, A. 2017.
\newblock Thompson Sampling for the MNL-Bandit.
\newblock In \emph{Conference on Learning Theory}, 76--78.

\bibitem[{Agrawal et~al.(2019)Agrawal, Avadhanula, Goyal, and
  Zeevi}]{agrawal2019mnl}
Agrawal, S.; Avadhanula, V.; Goyal, V.; and Zeevi, A. 2019.
\newblock MNL-bandit: A dynamic learning approach to assortment selection.
\newblock \emph{Operations Research} 67(5): 1453--1485.

\bibitem[{Aouad, Levi, and Segev(2018)}]{aouad2018greedy}
Aouad, A.; Levi, R.; and Segev, D. 2018.
\newblock Greedy-like algorithms for dynamic assortment planning under
  multinomial logit preferences.
\newblock \emph{Operations Research} 66(5): 1321--1345.

\bibitem[{Auer(2002)}]{auer2002using}
Auer, P. 2002.
\newblock Using confidence bounds for exploitation-exploration trade-offs.
\newblock \emph{Journal of Machine Learning Research} 3(Nov): 397--422.

\bibitem[{Auer, Cesa-Bianchi, and Fischer(2002)}]{auer2002finite}
Auer, P.; Cesa-Bianchi, N.; and Fischer, P. 2002.
\newblock Finite-time analysis of the multiarmed bandit problem.
\newblock \emph{Machine learning} 47(2-3): 235--256.

\bibitem[{Bartlett et~al.(2005)Bartlett, Bousquet, Mendelson
  et~al.}]{bartlett2005local}
Bartlett, P.~L.; Bousquet, O.; Mendelson, S.; et~al. 2005.
\newblock Local rademacher complexities.
\newblock \emph{The Annals of Statistics} 33(4): 1497--1537.

\bibitem[{Cao et~al.(2015)Cao, Li, Tao, and Li}]{cao2015top}
Cao, W.; Li, J.; Tao, Y.; and Li, Z. 2015.
\newblock On top-k selection in multi-armed bandits and hidden bipartite
  graphs.
\newblock In \emph{Advances in Neural Information Processing Systems},
  1036--1044.

\bibitem[{Caro and Gallien(2007)}]{caro2007dynamic}
Caro, F.; and Gallien, J. 2007.
\newblock Dynamic assortment with demand learning for seasonal consumer goods.
\newblock \emph{Management Science} 53(2): 276--292.

\bibitem[{Chen and Wang(2017)}]{chen2017note}
Chen, X.; and Wang, Y. 2017.
\newblock A Note on Tight Lower Bound for MNL-Bandit Assortment Selection
  Models.
\newblock \emph{arXiv preprint arXiv:1709.06109} .

\bibitem[{Chen, Wang, and Zhou(2018)}]{chen2018dynamic}
Chen, X.; Wang, Y.; and Zhou, Y. 2018.
\newblock Dynamic Assortment Optimization with Changing Contextual Information.
\newblock \emph{arXiv preprint arXiv:1810.13069} .

\bibitem[{Cheung and Simchi-Levi(2017)}]{cheung2017thompson}
Cheung, W.~C.; and Simchi-Levi, D. 2017.
\newblock Thompson sampling for online personalized assortment optimization
  problems with multinomial logit choice models.
\newblock \emph{Available at SSRN 3075658} .

\bibitem[{Chu et~al.(2011)Chu, Li, Reyzin, and Schapire}]{chu2011contextual}
Chu, W.; Li, L.; Reyzin, L.; and Schapire, R. 2011.
\newblock Contextual bandits with linear payoff functions.
\newblock In \emph{Proceedings of the Fourteenth International Conference on
  Artificial Intelligence and Statistics}, 208--214.

\bibitem[{Dani, Hayes, and Kakade(2008)}]{dani2008stochastic}
Dani, V.; Hayes, T.~P.; and Kakade, S.~M. 2008.
\newblock Stochastic linear optimization under bandit feedback.
\newblock In \emph{Proceedings of the 21st Annual Conference on Learning
  Theory}, 355–366.

\bibitem[{Davis, Gallego, and Topaloglu(2013)}]{davis2013assortment}
Davis, J.; Gallego, G.; and Topaloglu, H. 2013.
\newblock Assortment planning under the multinomial logit model with totally
  unimodular constraint structures .

\bibitem[{Davis, Gallego, and Topaloglu(2014)}]{davis2014assortment}
Davis, J.~M.; Gallego, G.; and Topaloglu, H. 2014.
\newblock Assortment optimization under variants of the nested logit model.
\newblock \emph{Operations Research} 62(2): 250--273.

\bibitem[{Filippi et~al.(2010)Filippi, Cappe, Garivier, and
  Szepesv{\'a}ri}]{filippi2010parametric}
Filippi, S.; Cappe, O.; Garivier, A.; and Szepesv{\'a}ri, C. 2010.
\newblock Parametric bandits: The generalized linear case.
\newblock In \emph{Advances in Neural Information Processing Systems},
  586--594.

\bibitem[{Ghose, Ipeirotis, and Li(2014)}]{ghose2014examining}
Ghose, A.; Ipeirotis, P.~G.; and Li, B. 2014.
\newblock Examining the impact of ranking on consumer behavior and search
  engine revenue.
\newblock \emph{Management Science} 60(7): 1632--1654.

\bibitem[{Hazan, Agarwal, and Kale(2007)}]{hazan2007logarithmic}
Hazan, E.; Agarwal, A.; and Kale, S. 2007.
\newblock Logarithmic regret algorithms for online convex optimization.
\newblock \emph{Machine Learning} 69(2-3): 169--192.

\bibitem[{Hazan, Koren, and Levy(2014)}]{hazan2014logistic}
Hazan, E.; Koren, T.; and Levy, K.~Y. 2014.
\newblock Logistic regression: Tight bounds for stochastic and online
  optimization.
\newblock In \emph{Conference on Learning Theory}, 197--209.

\bibitem[{Jaksch, Ortner, and Auer(2010)}]{jaksch2010near}
Jaksch, T.; Ortner, R.; and Auer, P. 2010.
\newblock Near-optimal regret bounds for reinforcement learning.
\newblock \emph{Journal of Machine Learning Research} 11(Apr): 1563--1600.

\bibitem[{Javanmard and Nazerzadeh(2019)}]{javanmard2019dynamic}
Javanmard, A.; and Nazerzadeh, H. 2019.
\newblock Dynamic pricing in high-dimensions.
\newblock \emph{The Journal of Machine Learning Research} 20(1): 315--363.

\bibitem[{Kveton et~al.(2015)Kveton, Szepesvari, Wen, and
  Ashkan}]{kveton2015cascading}
Kveton, B.; Szepesvari, C.; Wen, Z.; and Ashkan, A. 2015.
\newblock Cascading bandits: Learning to rank in the cascade model.
\newblock In \emph{International Conference on Machine Learning}, 767--776.

\bibitem[{Kveton et~al.(2019)Kveton, Zaheer, Szepesvari, Li, Ghavamzadeh, and
  Boutilier}]{kveton2019randomized}
Kveton, B.; Zaheer, M.; Szepesvari, C.; Li, L.; Ghavamzadeh, M.; and Boutilier,
  C. 2019.
\newblock Randomized Exploration in Generalized Linear Bandits.
\newblock \emph{arXiv preprint arXiv:1906.08947} .

\bibitem[{Lattimore and Szepesv\'{a}ri(2019)}]{LS19bandit-book}
Lattimore, T.; and Szepesv\'{a}ri, C. 2019.
\newblock \emph{Bandit Algorithms}.
\newblock Cambridge University Press (preprint).

\bibitem[{Lehmann and Casella(2006)}]{lehmann2006theory}
Lehmann, E.~L.; and Casella, G. 2006.
\newblock \emph{Theory of point estimation}.
\newblock Springer Science \& Business Media.

\bibitem[{Li, Lu, and Zhou(2017)}]{li2017provably}
Li, L.; Lu, Y.; and Zhou, D. 2017.
\newblock Provably Optimal Algorithms for Generalized Linear Contextual
  Bandits.
\newblock In \emph{International Conference on Machine Learning}, 2071--2080.

\bibitem[{McFadden(1978)}]{mcfadden1978modeling}
McFadden, D. 1978.
\newblock Modeling the choice of residential location.
\newblock \emph{Transportation Research Record} (673).

\bibitem[{Oh and Iyengar(2019)}]{oh2019thompson}
Oh, M.-h.; and Iyengar, G. 2019.
\newblock Thompson Sampling for Multinomial Logit Contextual Bandits.
\newblock In \emph{Advances in Neural Information Processing Systems},
  3145--3155.

\bibitem[{Oh, Iyengar, and Zeevi(2020)}]{oh2020sparsity}
Oh, M.-h.; Iyengar, G.; and Zeevi, A. 2020.
\newblock Sparsity-agnostic lasso bandit.
\newblock \emph{arXiv preprint arXiv:2007.08477} .

\bibitem[{Ou et~al.(2018)Ou, Li, Zhu, and Jin}]{ou2018multinomial}
Ou, M.; Li, N.; Zhu, S.; and Jin, R. 2018.
\newblock Multinomial Logit Bandit with Linear Utility Functions.
\newblock In \emph{Proceedings of the 27th International Joint Conference on
  Artificial Intelligence}, IJCAI’18, 2602–2608. AAAI Press.

\bibitem[{Pollard(1990)}]{pollard1990empirical}
Pollard, D. 1990.
\newblock Empirical processes: theory and applications.
\newblock In \emph{NSF-CBMS regional conference series in probability and
  statistics}, i--86. JSTOR.

\bibitem[{Qin, Chen, and Zhu(2014)}]{qin2014contextual}
Qin, L.; Chen, S.; and Zhu, X. 2014.
\newblock Contextual combinatorial bandit and its application on diversified
  online recommendation.
\newblock In \emph{Proceedings of the 2014 SIAM International Conference on
  Data Mining}, 461--469. SIAM.

\bibitem[{Rusmevichientong, Shen, and
  Shmoys(2010)}]{rusmevichientong2010dynamic}
Rusmevichientong, P.; Shen, Z.-J.~M.; and Shmoys, D.~B. 2010.
\newblock Dynamic assortment optimization with a multinomial logit choice model
  and capacity constraint.
\newblock \emph{Operations research} 58(6): 1666--1680.

\bibitem[{Rusmevichientong and Tsitsiklis(2010)}]{rusmevichientong2010linearly}
Rusmevichientong, P.; and Tsitsiklis, J.~N. 2010.
\newblock Linearly parameterized bandits.
\newblock \emph{Mathematics of Operations Research} 35(2): 395--411.

\bibitem[{Saur{\'e} and Zeevi(2013)}]{saure2013optimal}
Saur{\'e}, D.; and Zeevi, A. 2013.
\newblock Optimal dynamic assortment planning with demand learning.
\newblock \emph{Manufacturing \& Service Operations Management} 15(3):
  387--404.

\bibitem[{Thompson(1933)}]{thompson1933likelihood}
Thompson, W.~R. 1933.
\newblock On the likelihood that one unknown probability exceeds another in
  view of the evidence of two samples.
\newblock \emph{Biometrika} 25(3/4): 285--294.

\bibitem[{Tropp(2012)}]{tropp2012user}
Tropp, J.~A. 2012.
\newblock User-friendly tail bounds for sums of random matrices.
\newblock \emph{Foundations of computational mathematics} 12(4): 389--434.

\bibitem[{Wen, Kveton, and Ashkan(2015)}]{wen2015efficient}
Wen, Z.; Kveton, B.; and Ashkan, A. 2015.
\newblock Efficient learning in large-scale combinatorial semi-bandits.
\newblock In \emph{International Conference on Machine Learning}, 1113--1122.

\bibitem[{Zhang et~al.(2016)Zhang, Yang, Jin, Xiao, and Zhou}]{zhang2016online}
Zhang, L.; Yang, T.; Jin, R.; Xiao, Y.; and Zhou, Z.-H. 2016.
\newblock Online stochastic linear optimization under one-bit feedback.
\newblock In \emph{International Conference on Machine Learning}, 392--401.

\bibitem[{Zhou, Xu, and Blanchet(2019)}]{zhou2019learning}
Zhou, Z.; Xu, R.; and Blanchet, J. 2019.
\newblock Learning in generalized linear contextual bandits with stochastic
  delays.
\newblock In \emph{Advances in Neural Information Processing Systems},
  5198--5209.

\bibitem[{Zong et~al.(2016)Zong, Ni, Sung, Ke, Wen, and
  Kveton}]{zong2016cascading}
Zong, S.; Ni, H.; Sung, K.; Ke, N.~R.; Wen, Z.; and Kveton, B. 2016.
\newblock Cascading Bandits for Large-scale Recommendation Problems.
\newblock In \emph{Proceedings of the Thirty-Second Conference on Uncertainty
  in Artificial Intelligence}, UAI'16, 835--844.

\end{thebibliography}

\clearpage
\appendix
\onecolumn

\section{Related Work}\label{sec:related_work}

Besides the contextual bandit literature and their combinatorial variants mentioned in Introduction, our work also falls into the category of dynamic assortment optimization. \citet{rusmevichientong2010dynamic} and \citet{saure2013optimal} consider the problem of minimizing regret under the MNL choice model where \citet{rusmevichientong2010dynamic} showed $\mathcal{O}(N^2 \log^2 T)$ regret bound. \citet{saure2013optimal} improved the bound to $\mathcal{O}(N \log T)$. However, these methods require a priori knowledge of ``separability'' between the true optimal assortment and the other sub-optimal alternatives.
\citet{agrawal2017thompson, agrawal2019mnl} and \citet{chen2017note} also  formulated dynamic assortment selection as an online regret minimization problem. However, these previous works are in non-contextual settings and assume that each item is associated with a unique parameter, hence generalization across different items is not considered.

\citet{ou2018multinomial} extend \citet{agrawal2019mnl} to linear utility, yet they still assume the utilities are fixed over time.
Recent work by \citet{chen2018dynamic} establishes $\Tilde{\mathcal{O}}(d\sqrt{T})$ regret bound for the MNL contextual bandit with changing context, the same setting as ours. Although their algorithm appears similar to our first algorithm, \textsc{UCB-MNL} (Algorithm~\ref{algo:UCB-MNL}),
there is a fundamental difference between \citet{chen2018dynamic} and our \textsc{UCB-MNL}. \citet{chen2018dynamic} enumerates the exponentially many ($N$ choose $K$) assortments and builds confidence bounds for each of them. In contrast, \textsc{UCB-MNL} maintains the confidence bound in the parameter space and subsequently computes the upper confidence bounds of utility for each of the $N$ items. 
\citet{chen2018dynamic} recognize this computational issue and propose an approximate optimization algorithm to somewhat remedy it; however, not completely. Consider the simple case where each item has unit revenue. In this case, assortment selection under \textsc{UCB-MNL} reduces to sorting items based upper-confidence bounds and the run time is independent of $K$, whereas \citet{chen2018dynamic} still have to consider all the ($N$ choose $K$) assortments. \citet{oh2019thompson} consider Thompson sampling \cite{thompson1933likelihood} approach to the MNL contextual bandits but their regret bound still has a gap with $d^{3/2}\sqrt{T}$ regret. Currently known lower bounds are $\Omega(d\sqrt{T}/K)$ \cite{chen2018dynamic} and $K$ independent lower bound with $\Omega(\sqrt{dT})$ which comes from $\Omega(\sqrt{NT})$ of non-contextual setting in \citet{chen2017note}. However, no previous work has closed the gap for either case. Table~\ref{sample-table} summarizes the regret bounds and settings of recent work in the MNL bandits. Our main results in this paper provide tightest results for the MNL bandit problem in most general and practical settings.

\section{Proof Outline of Theorem~\ref{thm:expected_regret_ucb-mnl}}

The proof of the regret bound in Theorem ~\ref{thm:expected_regret_ucb-mnl} involves bounding the parameter estimation error $\| \Hat{\theta}_{t} - \theta^* \|_{V_{t}}$ and $\sum_{t'=1}^t \sum_{i \in S_{t'}} \| x_{t'i} \|_{V^{-1}_{t'-1}}$.
Also, we need to ensure that the optimistic utility estimate is indeed optimistic, i.e., the optimistic  utility estimate is higher than the true utility. We present the following key lemmas.

The initialization duration $T_0$ is specified in Theorem~\ref{thm:expected_regret_ucb-mnl}, which is chosen to ensure that $\lambda_{\min}(V_{T_0})$ is large enough so that we can ensure $\| \hat{\theta}_t - \theta^*\| \leq 1$ for $t > T_0$. The following proposition allows us to find such $T_0$.

\begin{proposition}[\citealt{li2017provably}, Proposition 1]\label{prop:lowerbounding_lambda}
Let $x_{t' i}$ be drawn i.i.d. from some distribution $\nu$ with $\| x_{t' i} \| \leq 1$ and $\mathbb{E}[x_{t' i} x_{t' i}^\top] \geq \sigma_0$ (Assumption~\ref{assum:x_bound}). Define $V_{T_0} = \sum^{T_0}_{t'=1}\sum_{i \in S_{t'}} x_{t' i} x_{t' i}^\top$, where $T_0$ is the length of random initialization.
Suppose we run a random initialization with assortment size $K$ for duration $T_0$ which satisfies 
\begin{align*}
    T_0 \geq \frac{1}{K}\left( \frac{C_1\sqrt{d} + C_2 \sqrt{2\log T}}{\sigma_0} \right)^2 + \frac{2B}{ K \sigma_0}
\end{align*}
for some positive, universal constants $C_1$ and $C_2$.
Then, $\lambda_{\min}(V_{T_0}) \geq B$ with probability at least $1-\frac{1}{T^2}$.
\end{proposition}

The proposition implies that we can have $\lambda_{\min}(V_{T_0}) \geq K$ with a high probability if we run the initialization for $\mathcal{O}(\sigma_0^{-2}(d + \log T))$ rounds. 
Similar to \citet{filippi2010parametric} and \citet{li2017provably}, the i.i.d. assumption (in Assumption 1) on the context $x_{ti}$ is only needed to ensure that $V_{T_0}$  is invertible at the end of the initialization phase. In the rest of the regret analysis, we do not require this stochastic assumption. Hence, after the initialization, $x_{ti}$ can even be chosen adversarily as long as $\| x_{ti} \|$ is bounded. We also want $\lambda_{\min}(V_{T_0})$ to be large enough so that  $\| \hat{\theta}_t - \theta^*\| \leq 1$ for $t > T_0$. The following lemma specifies how large $\lambda_{\min}(V_{T_0})$ should be
\begin{lemma}[\citealt{kveton2019randomized}, Lemma 9]\label{lemma:theta_hat_close}
Let $T_0$ be any round such that
\begin{align*}
    \lambda_{\min}(V_{T_0}) \geq \max\left\{ \sigma^2 \kappa^{-2} \left( d \log(T/d) + 4\log T \right), K \right\}.
\end{align*}
Then for any $t \geq T_0$, $\mathbb{P}\left( \| \hat{\theta}_t - \theta^* \| > 1 \right) \leq \frac{1}{T^2}$.
\end{lemma}

Lemma~\ref{lemma:theta_V_bound} shows that the true parameter $\theta^*$ lies within
an ellipsoid centered at $\Hat{\theta}_t$ with a suitable confidence radius under $V_t$ weighted $\ell_2$ norm with a high probability.
Recall that Proposition~\ref{prop:lowerbounding_lambda} ensures that we have $\lambda_{\min}(V_{T_0})$ is sufficiently large at the end of the initialization phase with a suitable initialization duration (which is specified in the statement of Theorem~\ref{thm:expected_regret_ucb-mnl}) if we run the initialization with size $K$ assortments; hence the algorithm satisfies the condition of the following lemma.

\begin{lemma}\label{lemma:theta_V_bound}
Suppose $\| \hat{\theta}_t - \theta^* \| \leq 1$ for $t > T_0$. Then
\begin{equation}\label{eq:confidence_width}
\| \Hat{\theta}_t - \theta^* \|_{V_t} \leq  \frac{1}{2\kappa}\sqrt{2d \log\left( 1 + \frac{t}{d} \right) + 2\log t }
\end{equation}
holds for all  $t > T_0$ with a probability $1 - \mathcal{O}(t^{-2})$.
\end{lemma}
The condition $\| \hat{\theta}_t - \theta^* \| \leq 1$ can be ensured with a high probability by combining Lemma~\ref{lemma:theta_hat_close} and Proposition~\ref{prop:lowerbounding_lambda}.
Lemma~\ref{lemma:theta_V_bound} is a finite-sample normality-type estimation error bound for the MLE of the MNL model. This result suggests that we can construct the optimistic utility estimate using the confidence radius $\alpha_t = \frac{1}{2\kappa}\sqrt{2d \log\left( 1 + \frac{t}{d} \right) + 2\log t }$. 
The following lemma shows our optimistic utility estimate $z_{ti}$ is an upper confidence bound for the expected utility $x^\top_{ti} \theta^*$ if the true parameter $\theta^*$ is contained in the confidence ellipsoid centered at $\Hat{\theta}_t$.

\begin{lemma}\label{lemma:utility_bound}
Let $z_{ti} =  x_{ti}^\top \Hat{\theta}_{t-1} + \alpha_t \| x_{ti} \|_{V^{-1}_{t}}$. If \eqref{eq:confidence_width} holds, then we have
\begin{equation*}
     0 \leq z_{ti} - x^\top_{ti} \theta^* \leq 2 \alpha_t \| x_{ti} \|_{V^{-1}_{t}}.
\end{equation*}
\end{lemma}

The following lemma shows that the optimistic expected revenue $\Tilde{R}_t(S_t)$ is an upper bound of the true expected revenue of the optimal assortment $R_t(S^*_t, \theta^*)$. The lemma is an adaptation of Lemma 4.2 in \cite{agrawal2019mnl} which is shown for non-contextual setting.\\

\begin{lemma}\label{lemma:revenue_bound}
Suppose $S^*_t$ is the offline optimal assortment as defined in \eqref{eq:S_star}, and suppose $S_t = \arg\max_{S \subset \mathcal{S} } \Tilde{R}_t(S)$. If for every item $i \in S^*_t$, $z_{ti} \geq x^\top_i \theta^*$, then the revenues satisfy the following inequalities for all round $t$:
\begin{equation*}
    R_t(S^*_t, \theta^*) \leq \Tilde{R}_t (S^*_t) \leq \Tilde{R}_t (S_t).
\end{equation*}
\end{lemma}

It is important to note that Lemma~\ref{lemma:revenue_bound} does not claim that the expected revenue is generally a monotone function, but only the value of the expected revenue corresponding to the optimal assortment increases with an increase in the MNL parameters \citep{agrawal2019mnl}.

Then we show that the expected revenue has Lipschitz property and bound the immediate regret with the maximum variance over the assortment.

\begin{lemma}\label{lemma:optimistic_regret}
Suppose that $0 \leq z_{ti} - x^\top_{ti} \theta^* \leq 2 \alpha_t \| x_{ti} \|_{V^{-1}_{t}}$ holds for $i \in S_t$ where $S_t$ is the chosen assortment in round $t$. Then, we have
\begin{equation*}
     \Tilde{R}_t(S_t) - R_t(S_t, \theta^*)  \leq   2\alpha_t \max_{i \in S_t} \| x_{ti} \|_{V^{-1}_{t}}
\end{equation*}
\end{lemma}

The next technical lemma bounds the sum of weighted squared norms. Note that we later apply Cauchy-Schwarz inequality to eventually bound  $\sum_{t=1}^T \max_{i \in S_t} \| x_{ti} \|_{V^{-1}_{t}}$ by $\Tilde{ \mathcal{O}}(\sqrt{dT})$. 

\begin{lemma}\label{lemma:self_norm_MG_bound}
Define $V_{T_0} = \sum^{T_0}_{t=1} \sum_{i\in S_{t'}} x_{ti} x_{ti}^\top$ and $V_T = V_{T_0} + \sum_{t=V_0+1}^T \sum_{i\in S_t} x_{ti} x_{ti}^\top$. If $\lambda_{\min}(V_{T_0}) \geq K$, then we have
\begin{equation*}\label{eq:self_normalized_bound}
    \sum_{t=1}^T \max_{i \in S_t} \| x_{ti} \|^2_{V^{-1}_{t}} \leq 2d \log \left( T/d \right)
\end{equation*}
\end{lemma}
Hence, each of Lemma~\ref{lemma:theta_V_bound} and Lemma~\ref{lemma:self_norm_MG_bound} contributes  $\sqrt{d}$ factor separately to the overall regret, resulting in $d$ factor in Theorem~\ref{thm:expected_regret_ucb-mnl}.
Now we can combine the results to show the cumulative regret bound. First we define the joint high probability event for the concentration of the MLE and the random initialization.
\begin{definition}
Define the following event:
\begin{align*}
        \hat{\mathcal{E}} &:= \left\{\| \hat{\theta}_t - \theta^*\| \leq 1, \enskip \|\Hat{\theta}_t - \theta^* \|_{V_t} \leq  \alpha_t, \forall t \geq T_0  \right\}
\end{align*}
\end{definition}

Note that by Proposition~\ref{prop:lowerbounding_lambda} with $T_0 = \max\left\{ \frac{\sigma^2 \left( d \log(T/d) + 4\log T\right) }{K \sigma_0 \kappa^2 } ,  \frac{d + \log T}{\sigma_0^2} \right\}$, we can show
\begin{align*}
\lambda_{\min}(V_{T_0}) \geq \max\left\{ \sigma^2 \kappa^{-2} \left( d \log(T/d) + 4\log T \right), K \right\}   
\end{align*}
with a high probability, which in turn can ensure $\| \hat{\theta}_t - \theta^*\| \leq 1$ by Lemma~\ref{lemma:theta_hat_close}.
 We first break the regret into the initialization phase and the learning phase:
\begin{align*}
    \mathcal{R}_T &= \mathbb{E} \left[ \sum_{t=1}^{T_0} \left( R(S^*_t, \theta^*) - R(S_t, \theta^*) \right)  \right] + \mathbb{E} \left[ \sum_{t=T_0+1}^T \left( R(S^*_t, \theta^*) - R(S_t, \theta^*) \right)  \right]\\
    &\leq T_0 + \mathbb{E} \left[ \sum_{t=T_0+1}^T   \left ( \Tilde{R}_t(S_t) - R(S_t, \theta^*) \right) \right] 
\end{align*}
where the last inequality comes from optimistic revenue estimation by Lemma~\ref{lemma:revenue_bound}.
Now, we further decompose the regret of the learning phase further into two components -- when the high probability event holds in Lemma~\ref{lemma:theta_V_bound} and in Lemma~\ref{lemma:theta_hat_close} (i.e., $\hat{\mathcal{E}}$ holds) and when either of the events does not hold, (i.e. $\hat{\mathcal{E}}^c$).
\begin{align*}
    \mathcal{R}_T &\leq T_0 + \mathbb{E} \left[ \sum_{t=T_0+1}^T  \left ( \Tilde{R}_t(S_t) - R_t(S_t, \theta^*) \right) \mathbb{1}(\hat{\mathcal{E}}) \right]
     + \mathbb{E} \left[ \sum_{t=T_0+1}^T   \left ( \Tilde{R}_t(S_t) - R_t(S_t, \theta^*) \right) \mathbb{1}(\hat{\mathcal{E}}^c) \right]\\
    & \leq T_0 + \mathbb{E} \left[ \sum_{t=T_0 + 1}^T  \left ( \Tilde{R}_t(S_t) - R_t(S_t, \theta^*) \right) \mathbb{1}(\hat{\mathcal{E}}) \right]
    +  \sum_{t=1}^T  \mathcal{O}(t^{-2})\\
    & \leq T_0 + \sum_{t=1}^T 2\alpha_T \max_{i \in S_t}   \| x_{ti} \|_{V^{-1}_{t}} + \mathcal{O}(1)
\end{align*}
where the last inequality is from Lemma~\ref{lemma:optimistic_regret}.
Applying Cauchy-Schwarz inequality in the second term, it follows that
\begin{align*}
      \mathcal{R}_T & \leq T_0 + 2\alpha_T \sqrt{  T \sum_{t=1}^T \max_{i \in S_t}    \| x_{ti} \|_{V^{-1}_{t}}^2 }
      + \mathcal{O}(1).
\end{align*}
Applying Lemma~\ref{lemma:self_norm_MG_bound} for $\sum_{t=1}^T \max_{i \in S_t}    \| x_{ti} \|_{V^{-1}_{t}}^2$,
\begin{align*}
      \mathcal{R}_T &\leq  T_0 + 2\alpha_T \sqrt{  2 d T \log \left( T/d \right) }
      + \mathcal{O}(1).
\end{align*}
Finally, letting $\alpha_T = \frac{\sigma}{\kappa}\sqrt{2d \log\left( 1 + \frac{T}{d} \right) + \log T }$, we have
\begin{align*}
      \mathcal{R}_T &\leq  T_0  + \frac{ d}{\kappa} \sqrt{ T  \log \left( 1 + T/d  \right) \log(T/d)} + \frac{1}{\kappa}\sqrt{d \log T \log \left(  T/d \right)} +  \mathcal{O}(1).
\end{align*}
\section{Proofs of Lemmas for Theorem~\ref{thm:expected_regret_ucb-mnl}}
\subsection{Proof of lemma~\ref{lemma:theta_V_bound}}
\proof{}
We first define the following:
\begin{align*}
J_n(\theta) &= \sum_{t=1}^{n-1} \sum_{i \in S_t} \left( p_t(i|S_t,\theta) - p_t(i|S_t, \theta^*) \right)x_{ti}\\
Z_n &:= J_n(\Hat{\theta}) = \sum^n_{t=1} \sum_{i \in S_t} \epsilon_{ti} x_{ti} \,.     
\end{align*}
Then we follow the same arguments of the proof of Theorem~\ref{thm:normalityMLE} until (\ref{eq:theta_G_bound}) which states
\begin{equation*}
    \| Z_n \|_{V^{-1}_n} = \| J_n(\Hat{\theta}) \|_{V^{-1}_n} \geq \kappa^2 \| \Hat{\theta} - \theta^* \|^2_{V_n}
\end{equation*}
for any $\Hat{\theta} \in \{ \theta : \| \theta - \theta^* \| \leq 1 \}$. Then we are left to bound $\| Z_n \|^2_{V^{-1}_n}$.
We can use Theorem 1 in \cite{abbasi2011improved}, which states if the noise $\epsilon_{ti}$ is sub-gaussian with parameter $\sigma$, then
\begin{equation*}\label{eqn:selfnorm_thm}
    \| Z_n \|_{V^{-1}_n}^2  \leq 2\sigma^2 \log \left( \frac{\det (V_n)^{1/2}\det(V_{T_0})^{-1/2}}{\delta} \right)
\end{equation*}
with probability at least $1-\delta$. Then we combine with Lemma~\ref{lemma:det_trace}. So it follows that
\begin{align*}
    \| Z_n \|_{V^{-1}_n}^2 &\leq 2\sigma^2  \left[ \frac{d}{2} \log \left(  \frac{ \text{trace}(V_{T_0}) + n K}{d} \right)  - \frac{1}{2} \log \det (V_{T_0})   + \log  \frac{1}{\delta} \right].
\end{align*}
Let $\lambda_1, ..., \lambda_d$ be the eigenvalues of $V_{T_0}$ and let $\Bar{\lambda} = \frac{\sum_i \lambda_i}{d}$, then
\begin{align*}
        \| Z_n \|_{V^{-1}_n}^2 &\leq 2\sigma^2  \left[ \frac{d}{2} \log \left(\Bar{\lambda} + \frac{ n K}{d} \right) - \frac{d}{2}\log \Bar{\lambda} + \frac{d}{2}\log \Bar{\lambda}  - \frac{1}{2} \log \det (V_{T_0})   + \log  \frac{1}{\delta} \right]\\
        &= 2\sigma^2  \left[ \frac{d}{2} \log \left(1 + \frac{ n K}{d\Bar{\lambda}} \right)  + \frac{1}{2} \sum_i \log \frac{\Bar{\lambda}}{\lambda_i}   + \log  \frac{1}{\delta} \right]\\
        &\leq 2\sigma^2  \left[ \frac{d}{2} \log \left(1 + \frac{ n K}{d \lambda_{\min}(V_{T_0})} \right)  + \frac{d}{2} \log \frac{\Bar{\lambda}}{\lambda_{\min}(V_{T_0})}   + \log  \frac{1}{\delta} \right]\\
        &\leq 2\sigma^2  \left[ \frac{d}{2} \log \left(1 + \frac{ n }{d} \right)  + \frac{d}{2} \log \frac{\Bar{\lambda}}{K}   + \log  \frac{1}{\delta} \right]\\
        &\leq 2\sigma^2  \left[ d \log \left(1 + \frac{ n }{d} \right)   + \log  \frac{1}{\delta} \right]
\end{align*}
where the third inequality is by $\lambda_{\min}(V_{T_0}) \geq K$ and the last inequality is from $d \Bar{\lambda} = \text{trace}(V_{T_0}) \leq nK$.
Then, using the fact that $\sigma^2 = \frac{1}{4}$ in our problem, we have that
\begin{equation*}
    \| \Hat{\theta}_n - \theta^* \|_{V_n} \leq \frac{1}{2\kappa}\sqrt{2d \log\left( 1 + \frac{n}{d} \right) + \log \frac{1}{\delta} }.
\end{equation*}
with probability at least $1-\delta$.  
\endproof

\subsection{Proof of Lemma~\ref{lemma:self_norm_MG_bound}}

The proof of Lemma~\ref{lemma:self_norm_MG_bound} requires the following  technical lemmas. These lemmas follow from the proof of Lemma 6 in \cite{oh2019thompson} with a subtle difference in $V_{T_0}$. For completeness, we present the proof in this specific setting.



\begin{lemma}\label{lemma:squarednorm_bound}
Suppose $\| x_{ti} \| \leq 1$ for all $i$ and $t$. Define $\displaystyle V_t = V_{T_0} + \sum_{t'=T_0 + 1}^t \sum_{i\in S_{t'}} x_{t' i} x_{t' i}^\top$. If $\lambda_{\min}(V_{T_0}) \geq K$. Then
\begin{equation*}
    \sum_{t'=T_0 + 1}^t \sum_{i \in S_{t'}} \| x_{t' i} \|_{V^{-1}_{t'-1}}^2 \leq 2 \log \left( \frac{\det(V_t)}{\lambda_{\min}(V_{T_0})^d} \right)
\end{equation*}
\end{lemma}

\proof{}
Let $\lambda_1, \lambda_2, ..., \lambda_d$ be the eigenvalues of $\sum_{i=1}^n x_{ti} x_{ti}^\top$. Since $\sum_{i=1}^n x_{ti} x_{ti}^\top$ is positive semi-definite, $\lambda_j \geq 0$ for all $j$. Hence, we have
\begin{align}\label{eq:lowerbd_det_cov}
    \det \left( I + \sum_{i \in S_t} x_{ti} x_{ti}^\top \right) &= \prod_{j=1}^d \left( 1 + \lambda_j \right) \notag\\
    &\geq 1 + \sum_{j=1}^d \lambda_j
    = 1-d + \sum_{j=1}^d (1 + \lambda_j) \notag\\
    &= 1-d + \text{trace}\left( I +  \sum_{i \in S_t} x_{ti} x_{ti}^\top \right)
    = 1 + \sum_{i \in S_t} \|x_{ti}\|^2_2
\end{align}

Now, we lower-bound $\det(V_t)$.
\begin{align}
\det(V_t) &= \det \left(  V_t + \sum_{i \in S_t} x_{ti} x_{ti}^\top  \right) \notag\\
&= \det(V_t) \det \left(  I + \sum_{i \in S_t} V_t^{-1/2} x_{ti} (V_t^{-1/2} x_{ti})^\top  \right) \notag\\
&\geq \det(V_t) \left(  1 + \sum_{i \in S_t} \| x_{ti} \|_{V^{-1}_{t}}^2  \right) \notag\\
&\geq \det(V_{T_0})\prod_{t'=T_0 + 1}^t \left(  1 + \sum_{i \in S_t} \| x_{t' i} \|_{V^{-1}_{t'-1}}^2  \right)\label{eq:det_ratio}
\end{align}
The first inequality comes from \eqref{eq:lowerbd_det_cov}. The second inequality comes from applying the first inequality repeatedly.
Let $\lambda_{\min} (V_{t})$ be the minimum eigenvalue of $V_t$. Notice that
\begin{equation*}
    \| x_{ti} \|_{V^{-1}_{t'-1}}^2 \leq  \frac{\| x_{ti} \|^2}{\lambda_{\min} (V_{\tau-1})}   \leq \frac{1}{\lambda_{\min}(V_{T_0})} \leq \frac{1}{K}.
\end{equation*}
Hence $\sum_{i \in S_t} \| x_{ti} \|_{V^{-1}_{t'-1}}^2 \leq 1$ for all $t \geq T_0$. Then using the fact that $z \leq 2 \log(1 + z)$ for any $z \in [0,1]$, we have
\begin{align*}
    \sum_{t'=T_0 + 1}^t \sum_{i \in S_{t'}} \| x_{t' i} \|_{V^{-1}_{t'-1}}^2 &\leq 2 \sum_{t'=T_0 + 1}^t \log \left( 1 + \sum_{i \in S_{t'}} \| x_{t' i} \|_{V^{-1}_{t'-1}}^2 \right)\\
    &= 2 \log \prod_{t'=T_0 + 1}^t  \left( 1 + \sum_{i \in S_{t'}} \| x_{t' i} \|_{V^{-1}_{t'-1}}^2 \right)\\
    &\leq 2 \log \left( \frac{\det (V_t)}{\det (V_{T_0})}  \right)\\
    &\leq 2 \log \left( \frac{\det (V_t)}{\lambda_{\min}(V_{T_0})^d}  \right)
\end{align*}
The second inequality is from (\ref{eq:det_ratio}). 
\endproof

\begin{lemma}\label{lemma:det_trace}
Suppose $\| x_{ti} \| \leq 1$ for all $i$ and $t$. Then $\det(V_t)$ is increasing with respect to $t$ and
\begin{equation}
    \det(V_t) \leq \left( \frac{t K}{d} \right)^d
\end{equation}
\end{lemma}

\proof{}
For any symmetric positive definite matrix $\Tilde{V} \in \mathbb{R}^{d \times d}$ and column vector $x \in \mathbb{R}^d$, we have
\begin{align*}
    \det (\Tilde{V} + xx^\top) &= \det(V) \det \left( I + \Tilde{V}^{-1/2} x x^\top \Tilde{V}^{-1/2} \right)\\
    &= \det(\Tilde{V}) \det(1 + \| \Tilde{V}^{-1/2} x\|^2)\\
    &\geq \det(\Tilde{V}).
\end{align*}
The second equality above is due to Sylvester’s determinant theorem, which states that $\det (I + BA) = \det (I + AB)$.
Let $\lambda_1, ..., \lambda_d  > 0$ be the eigenvalues of $V_t$. Then
\begin{align*}
    \det(V_t) 
    &\leq \left( \frac{\lambda_1 + ... + \lambda_d}{d} \right)^d\\
    &= \left( \frac{\text{trace}(V_t)}{d} \right)^d\\
    &= \left( \frac{ \sum_{t'= 1}^t \sum_{i \in S_{t'}} \text{trace}(x_{t' i} x_{t' i}^\top)}{d} \right)^d\\
    &= \left( \frac{ \sum_{t'= 1}^t \sum_{i \in S_{t'}} \| x_{t' i} \|_2^2}{d} \right)^d\\
    &\leq \left( \frac{t K }{d} \right)^d.
\end{align*}
 
\endproof

\proof{Proof of Lemma~\ref{lemma:self_norm_MG_bound}}
Combining Lemma~\ref{lemma:squarednorm_bound} and Lemma~\ref{lemma:det_trace}, 
\begin{align*}
    \sum_{t'=1}^t \max_{i \in S_{t'}} \| x_{t' i} \|_{V^{-1}_{\tau}}^2 
    \leq 2 \log \left( \frac{\det (V_t)}{\det (V_{T_0})}  \right)
    \leq 2 \log \left( \frac{t K }{d \lambda_{\min}(V_{T_0})} \right)^d 
    \leq 2d \log \left( t/d\right).
\end{align*}
where the last inequality is by $\lambda_{\min}(V_{T_0}) \geq K$. Then we complete the proof.
\endproof

\subsection{Proof of Lemma~\ref{lemma:utility_bound}}
\proof{}

\begin{align*}
    |x_{ti}^\top \Hat{\theta}_{t-1} - x_{ti}^\top \theta^*| &= \left| \left[ V^{-1/2}_{t-1} (\Hat{\theta}_{t-1} - \theta^*) \right]^\top (V^{-1/2}_{t-1} x_{ti}) \right|\\
    &\leq \| V^{-1/2}_{t-1} (\Hat{\theta}_{t-1} - \theta^*) \|_2 \| (V^{-1/2}_{t-1} x_{ti}) \|_2 \\
    &= \| \Hat{\theta}_{t-1} - \theta^* \|_{V_t} \| x_{ti} \|_{V^{-1}_{t}}\\
    &\leq  \alpha  \| x_{ti} \|_{V^{-1}_{t}}
\end{align*}
where the first inequality is by H\"older's inequality. Hence, it follows that
\begin{align*}
    \left( x_{ti}^\top \Hat{\theta}_{t-1} + \alpha \| x_{ti} \|_{V^{-1}_{t}} \right) - x_{ti}^\top \theta^* \leq 2  \alpha  \| x_{ti} \|_{V^{-1}_{t}}.
\end{align*}
Also, From $|x_{ti}^\top \Hat{\theta}_{t-1} - x_{ti}^\top \theta^*| \leq \alpha  \| x_{ti} \|_{V^{-1}_{t}}$, we have
\begin{align*}
     x_{ti}^\top \Hat{\theta}_{t-1}  - x_{ti}^\top \theta^* \geq -\alpha  \| x_{ti} \|_{V^{-1}_{t}}
\end{align*}
Hence, we have $\left( x_{ti}^\top \Hat{\theta}_{t-1} + \alpha \| x_{ti} \|_{V^{-1}_{t}} \right) - x_{ti}^\top \theta^* \geq 0$
\endproof

\subsection{Proof of Lemma~\ref{lemma:optimistic_regret}}

\proof{}
Let  $u_{ti} \geq u'_{ti}$ for all $i$.
By the mean value theorem, there exists $\bar{u}_{ti} := (1-c)u_{ti} + cu'_{ti}$ for some $c \in (0,1)$ with
\begin{align*}
      &\frac{\sum_{i \in S} r_{ti}  \exp \left(u_{ti} \right)}{1 + \sum_{j \in S} \exp \left(u_{tj}\right)} - \frac{\sum_{i \in S} r_{ti}  \exp (u'_{ti} )}{1 + \sum_{j \in S} \exp (u'_{tj})}\\
      &=  \frac{ (\sum_{i \in S} r_{ti} \exp\{ \bar{u}_{ti} \} (u_{ti} - u'_{ti}) ) (1 + \sum_{i \in S} \exp\{ \bar{u}_{ti} \})  }{(1 + \sum_{i \in S} \exp\{ \bar{u}_{ti} \})^2}\\
      &\qquad - \frac{ (\sum_{i \in S} r_{ti} \exp\{ \bar{u}_{ti} \}) ( \sum_{i \in S} \exp\{ \bar{u}_{ti} \} (u_{ti} - u'_{ti})) }{(1 + \sum_{i \in S} \exp\{ \bar{u}_{ti} \})^2} \\
     &=  \sum_{i \in S} r_{ti} p_{ti}(S, \bar{u}_{t})(u_{ti} - u'_{ti}) - R_t(S, \bar{u}_{t})\cdot \sum_{i \in S} p_{ti}(S, \bar{u}_{t})(u_{ti} - u'_{ti}) \\
     &=  \sum_{i \in S}  \big(r_{ti} - R_t(S, \bar{u}_{t}) \big) p_{ti}(S, \bar{u}_{t})  (u_{ti} - u'_{ti})\\
     &\leq  \max_{i \in S}  | u_{ti} - u'_{ti} | = \max_{i \in S}  ( u_{ti} - u'_{ti} )
\end{align*}
where the inequality is from $| r_{ti} | \leq 1$, and $p_{ti}(S, \bar{u}_{t}) \leq 1$ is a multinomial probability. \endproof




\section{Online Parameter Update}\label{sec:online_update}
UCB-MNL is simple to implement and more practical compared to previously known methods in MNL bandit problems. The algorithm also enjoys a good theoretical property, in particular, good statistical efficiency shown in Theorem~\ref{thm:expected_regret_ucb-mnl}. 
Despite these advantages, however, \textsc{UCB-MNL} can be still computationally expensive. In each round $t$, the MLE $\hat{\theta}_t$ is computed using $\Theta(tK)$ samples, i.e., the per-round computational complexity grows at least linearly with $t$ for a  straightforward implementation of the algorithm. Note that this issue is not unique to \textsc{UCB-MNL}. \cite{chen2018dynamic} also suffers from the same issue in addition to its computationally expensive procedure of the upper confidence construction for all assortments which we discussed earlier.  In fact, this bottleneck makes many bandit algorithms including those in generalized linear bandits \citep{filippi2010parametric,li2017provably} inappropriate for online implementations in real-world applications since the entire learning history is stored in memory and used for parameter estimation in each round.

In this section, we discuss a modification of \textsc{UCB-MNL} which incorporate an efficient online update that effectively exploits particular structures of the MNL model. As mentioned earlier, computing the exact solution for MLE does not scale well in time and space complexity. Hence, we propose an online update scheme to find an approximate solution.
First, we define the per-round loss for the MNL model and its gradient.

\begin{definition}\label{def:MNL_loss_per_round} 
Define the per-round loss $f_t(\theta)$ and its gradient $G_t(\theta)$ as the following:
\begin{align*}
    f_t(\theta) &:= - \sum_{i \in S_t} y_{ti} \log p_{t}(i|S_t, \theta)
    = -\sum_{i \in S_t} y_{ti} x_{ti}^\top \theta  + \log\Big(1 + \sum_{j \in S_t} \exp(x_{tj}^\top \theta)\Big)\\
    G_t(\theta) &:= \nabla_\theta f_t(\theta) = \sum_{i \in S_t} \left(  p_{t}(i|S_t, \theta) - y_{ti} \right) x_{ti} 
\end{align*} 
\end{definition}
The important observation here is that the loss for the MNL model at each round $t$ is strongly convex over bounded domain, which enables us to apply a variant of the online Newton step \citep{hazan2007logarithmic}, in particular inspired by \citep{hazan2014logistic,zhang2016online} which proposed online algorithms for the logistic model. Specifically, we propose to find an approximate solution  by solving the following problem
\begin{align}\label{eq:online_update_obj}
    \hat{\theta}_{t} = \argmin_{\theta} \left\{ \frac{1}{2}\| \theta - \hat{\theta}_{t-1}\|^2_{V_{t}} + (\theta - \hat{\theta}_{t-1})^\top G_{t-1} (\hat{\theta}_{t-1}) \right\}
\end{align}
where $V_{t} = V_t + \frac{\kappa}{2} \sum_{i \in S_t} x_{ti}x_{ti}^\top$.

\begin{algorithm}
\caption{UCB-MNL with online parameter update}
\begin{algorithmic}[1]
    \STATE \textbf{Input}: total rounds $T$, initialization rounds $T_0$ and confidence radius $\tilde{\alpha}_t$
    \STATE \textbf{Initialization}: \textbf{for} $t \in [T_0]$
    \STATE \quad Randomly choose $S_t$ with $|S_t|=K$
    \STATE \quad $V_t \leftarrow V_{t-1} + \sum_{i \in S_t} x_{ti} x_{ti}^\top$
    \FOR{all $t = T_0 + 1$ to $T$}
        \STATE Compute $\tilde{z}_{ti} =  x_{ti}^\top \Hat{\theta}_{t-1} + \tilde{\alpha}_t \| x_{ti} \|_{V^{-1}_{t-1}}$ for all $i \in [N]$
        \STATE Compute $S_t = \argmax_{S \subset \mathcal{S}} \Tilde{R}_t(S)$ based on $\{ \tilde{z}_{ti} \}$
        \STATE Offer $S_t$ and observe $y_t$ (user choice at time $t$)
        \STATE Update $V_t \leftarrow V_{t-1} + \frac{\kappa}{2}\sum_{i \in S_t} x_{ti} x_{ti}^\top$
        \STATE Compute $\Hat{\theta}_t$ by solving the problem
        \begin{align*}
            \hat{\theta}_{t} = \argmin_{\theta} \left\{ \frac{1}{2}\| \theta - \hat{\theta}_{t-1}\|^2_{V_{t}} + (\theta - \hat{\theta}_{t-1})^\top G_{t-1} (\hat{\theta}_{t-1}) \right\}
        \end{align*}
    \ENDFOR
\end{algorithmic}
\label{algo:UCB-MNL-online-update}
\end{algorithm}

The modified algorithm is summarized in Algorithm~\ref{algo:UCB-MNL-online-update}. The key difference is the parameter update rule in \eqref{eq:online_update_obj} and the corresponding confidence radius. During the learning phase, the learning agent builds a upper confidence utility estimate $\tilde{z}_{ti}$ based on a new confidence radius $\tilde{\alpha}_t$ which is specified in Lemma~\ref{lemma:online_update} and Theorem~\ref{thm:expected_regret_ucb-mnl-online-update}.  For parameter estimation, only $\Theta(K)$ samples are needed (for both computation and space) per each round, compared to $\Theta(tK)$ in Algorithm~\ref{algo:UCB-MNL} which grows linearly with each round $t$.

\begin{lemma}\label{lemma:online_update}
If $\lambda_{\min}(V_{T_0}) \geq K$, then
\begin{align*}
    \| \hat{\theta}_{t} - \theta^* \|_{V_{t}}   
    &\leq \sqrt{ T_0 + \frac{8}{\kappa} d \log \left(  1 + \frac{t}{d } \right) + \left(\frac{8}{\kappa} + \frac{16}{3}\right) \log\left( \lceil 2\log_2 (tK/2)\rceil t^4 \right) + 4}
\end{align*}
holds for all  $t > T_0$ with a probability $1 - \mathcal{O}(t^{-2})$.
\end{lemma}

The proof relies on exploiting the structure of the MNL loss and concentration inequalities for
martingales. Since we use fewer samples (less information) per update in the modified online update compared to the MLE computation, one might expect the confidence bound to increase with the online update modification.
Nevertheless, Lemma~\ref{lemma:online_update} shows the confidence bound with $\mathcal{O}\big(\sqrt{d \log \left(  1 + t/d \right)} \big)$ which is of the same order as the bound shown in Lemma~\ref{lemma:theta_V_bound} -- although there are extra additive terms and potentially a larger constant. This suggests that the total regret bound for the modified \textsc{UCB-MNL} should be also of the same order as the original \textsc{UCB-MNL}. We present the regret bound for the \textsc{UCB-MNL} with online parameter update, which is an formal statement of Corollary~\ref{cor:online_update_regret}.

\begin{theorem}\label{thm:expected_regret_ucb-mnl-online-update}
There exists a universal constant $C_0 > 0$, such that if we run \textsc{UCB-MNL} with ``online parameter update'' (Algorithm~\ref{algo:UCB-MNL-online-update}) with confident radius $\tilde{\alpha}_t$ for total of $T$ rounds with $T_0 =  \left\lceil C_0\max\left\{\frac{d + \log T}{\sigma_0^2 K } ,\frac{1}{\sigma_0} \right\} \right\rceil$ assortment size constraint $K$, then the expected regret of the algorithm with is upper-bounded by
\begin{align*}
      \mathcal{R}_T &\leq  T_0 + \mathcal{O}(1) + \tilde{\alpha}_T  \sqrt{ dT \log \left( T/d \right)  }
      = \mathcal{O}\left(  d\sqrt{ T  \log \left(  1 + T/d \right) \log(T/d) }  \right)
\end{align*}
where $\tilde{\alpha}_t = \sqrt{ T_0 + \frac{8}{\kappa} d \log \left(  1 + \frac{t}{d } \right) + \left(\frac{8}{\kappa} + \frac{16}{3}\right) \log\left( \lceil 2\log_2 (tK/2)\rceil t^4 \right) + 4}$.
\end{theorem}

Theorem~\ref{thm:expected_regret_ucb-mnl-online-update} achieves a regret bound of $\tilde{\mathcal{O}}(d\sqrt{T})$ which matches the bound in Theorem~\ref{thm:expected_regret_ucb-mnl} for \textsc{UCB-MNL}. The proof of Theorem~\ref{thm:expected_regret_ucb-mnl-online-update} follows the similar steps as Theorem~\ref{thm:expected_regret_ucb-mnl} and is presented in the follwing section.  This result suggests that the modified \textsc{UCB-MNL} is appropriate for online implementation, achieving both statistical and computational efficiency.
In Section~\ref{sec:numerics}, we compare the numerical performances of \textsc{UCB-MNL} and its online update modification along with other benchmarks.

\section{Proofs for Lemma~\ref{lemma:online_update} and Theorem~\ref{thm:expected_regret_ucb-mnl-online-update}}

The proof of Lemma~\ref{lemma:online_update} depends on the few technical lemma we present here in this section.
Recall from Definition~\ref{def:MNL_loss_per_round}
for the per-round loss $f_t(\theta)$ and its gradient $G_t(\theta)$:
\begin{align*}
    f_t(\theta) &= - \sum_{i \in S_t \cup \{0\}} y_{ti} \log p_{t}(i|S_t, \theta)
    = -\sum_{i \in S_t} y_{ti} x_{ti}^\top \theta  + \log\Big(1 + \sum_{j \in S_t} \exp(x_{tj}^\top \theta)\Big)\\
    G_t(\theta) &= \nabla_\theta f_t(\theta) = \sum_{i \in S_t} \left( p_{t}(i|S_t, \theta) - y_{ti} \right) x_{ti} 
\end{align*} 
We will use these terms throughout this section. In addition to $f_t(\theta)$ and $G_t(\theta)$, we also define their conditional expectations which we will utilize in the proofs of this section.
\begin{definition}
 Define the conditional expectations over $y$ of $f_t(\theta)$ and its gradient $G_t(\theta)$.
\begin{align*}
    \bar{f}_t(\theta) := \mathbb{E}_{y} \left[ f_t(\theta) | \mathcal{F}_t\right] \qquad
    \bar{G}_t(\theta) := \mathbb{E}_y[G_t(\theta) | \mathcal{F}_t] = \mathbb{E}_y[\nabla f_t(\theta) | \mathcal{F}_t]
\end{align*}
\end{definition}

\begin{lemma}\label{lemma:f_taylor_exp}
For any $\theta_1, \theta_2$, we have
\begin{equation*}
    f_t(\theta_2) \geq f_t(\theta_1) + G_t(\theta_1)^\top (\theta_2 - \theta_1) + \frac{\kappa}{2} (\theta_2 - \theta_1)^\top \Big( \sum_{i\in S_t} x_{ti} x_{ti}^\top \Big)  (\theta_2 - \theta_1)
\end{equation*}
\end{lemma}

\proof{}
Using the Taylor expansion, with $\Bar{\theta} = c\theta_2 - (1-c)\theta_1$ for some $c \in (0,1)$
\begin{align}\label{eq:risk_taylor_exp}
    f_t(\theta_2) =  f_t(\theta_1) + G_t(\theta_1)^\top (\theta_2 - \theta_1) + \frac{1}{2}(\theta_2 - \theta_1)^\top H_{f}(\Bar{\theta})(\theta_2 - \theta_1)
\end{align}
where $H_{f}(\Bar{\theta})$ is the Hessian matrix at $\Bar{\theta}$. Following the proof of Theorem~\ref{thm:normalityMLE}, the Hessian matrix can be lower-bounded as follows
\begin{align*}
    H_{f}(\Bar{\theta}) &= \sum_{i \in S_t} p_{t}(i|S_t, \Bar{\theta}) x_{ti}x_{ti}^\top - \sum_{i \in S_t}\sum_{j \in S_t} p_{t}(i|S_t, \Bar{\theta}) p_{tj}(S_t, \Bar{\theta}) x_{ti}x_{tj}^\top \\
    &\succeq  \sum_{i \in S_t} p_{t}(i|S_t, \Bar{\theta}) p_{t0}(\Bar{\theta}) x_{ti}x_{ti}^\top 
\end{align*}
From Assumption~\ref{assum:prob_bound}, we have
\begin{align*}
    H_{f}(\Bar{\theta}) \succeq \kappa \sum_{i \in S_t}  x_{ti}x_{ti}^\top 
\end{align*}
Therefore, we have
\begin{align*}
    f_t(\theta_2) &=  f_t(\theta_1) + G_t(\theta_1)^\top (\theta_2 - \theta_1) + \frac{1}{2}(\theta_2 - \theta_1)^\top H_{f}(\Bar{\theta})(\theta_2 - \theta_1)\\
    &\geq  f_t(\theta_1) + G_t(\theta_1)^\top (\theta_2 - \theta_1) + \frac{\kappa}{2}(\theta_2 - \theta_1)^\top \Big( \sum_{i \in S_t}  x_{ti}x_{ti}^\top \Big)(\theta_2 - \theta_1).
\end{align*}
\hfill
\endproof

\begin{lemma}\label{lemma:taylor_1st_2nd_bound}
\begin{align*}
     2G_t (\hat{\theta}_t)^\top (\hat{\theta}_t - \theta^*) 
    \leq \|G_t(\theta_t)\|^2_{V^{-1}_{t+1}} + \| \hat{\theta}_t - \theta^* \|^2_{V_{t+1}} - \| \hat{\theta}_{t+1} - \theta^* \|^2_{V_{t+1}}
\end{align*}
\end{lemma}

\proof{}
Note that $\hat{\theta}_{t+1}$ is the optimal solution to the problem
\begin{align*}
    \hat{\theta}_{t+1} = \argmin_{\theta} \frac{1}{2}\| \theta - \hat{\theta}_t\|^2_{V_{t+1}} + (\theta - \hat{\theta}_t)^\top G_t (\hat{\theta}_t)
\end{align*}
Hence, from the first-order optimality condition, we have
\begin{align*}
    \left[  G_t (\hat{\theta}_t) + V_{t+1}(\hat{\theta}_{t+1} - \hat{\theta}_{t}) \right]^\top(\theta - \hat{\theta}_{t+1}) \geq 0, \forall \theta
\end{align*}
which gives
\begin{align*}
    \theta^\top V_{t+1}(\hat{\theta}_{t+1} - \hat{\theta}_t) \geq  \hat{\theta}_{t+1} ^\top V_{t+1}(\hat{\theta}_{t+1} - \hat{\theta}_t) -  G_t(\hat{\theta}_t) (\theta - \hat{\theta}_{t+1} ).
\end{align*}
Then we can write
\begin{align*}
    &\| \hat{\theta}_t - \theta^* \|^2_{V_{t+1}} - \| \hat{\theta}_{t+1} - \theta^* \|^2_{V_{t+1}}\\
    &= \hat{\theta}_t^\top V_{t+1} \hat{\theta}_t - \hat{\theta}_{t+1}^\top V_{t+1} \hat{\theta}_{t+1} + 2 {\theta^*}^\top V_{t+1}(\hat{\theta}_{t+1} - \hat{\theta}_t)\\
    &\geq \hat{\theta}_t^\top V_{t+1} \hat{\theta}_t - \hat{\theta}_{t+1}^\top V_{t+1} \hat{\theta}_{t+1} + 2 \hat{\theta}_{t+1} ^\top V_{t+1}(\hat{\theta}_{t+1} - \hat{\theta}_t) - 2 G_t(\hat{\theta}_t) (\theta^* - \hat{\theta}_{t+1} )\\
    &= \hat{\theta}_t^\top V_{t+1} \hat{\theta}_t + \hat{\theta}_{t+1}^\top V_{t+1} \hat{\theta}_{t+1} - 2 \hat{\theta}_{t+1} ^\top V_{t+1}\hat{\theta}_t - 2 G_t(\hat{\theta}_t) (\theta^* - \hat{\theta}_{t+1} )\\
    &=\| \hat{\theta}_t - \hat{\theta}_{t+1} \|^2_{V_{t+1}} + 2 G_t(\hat{\theta}_t) ( \hat{\theta}_{t+1} - \hat{\theta}_t) + 2 G_t(\hat{\theta}_t) ( \hat{\theta}_t - \theta^*)\\
    &\geq  -\|G_t(\theta_t)\|^2_{V^{-1}_{t+1}} + 2 G_t(\hat{\theta}_t) ( \hat{\theta}_t - \theta^*)
\end{align*}
where the last inequality is from the fact that
\begin{align*}
\| \hat{\theta}_t - \hat{\theta}_{t+1} \|^2_{V_{t+1}} + 2 G_t(\hat{\theta}_t) ( \hat{\theta}_{t+1} - \hat{\theta}_t) &\geq \min_\theta \left\{ \| \theta \|^2_{V_{t+1}} + 2 G_t(\hat{\theta}_t) ( \theta ) \right\} \\
&= -\|G_t(\theta_t)\|^2_{V^{-1}_{t+1}}.
\end{align*}
 \endproof

\begin{lemma}\label{lemma:f_opt_ineq}
For all $\theta \in \mathbb{R}^d$, we have $\bar{f}_t(\theta) \geq \bar{f}_t(\theta^*)$.
\end{lemma}

\proof{}
\begin{align*}
    \bar{f}_t(\theta) - \bar{f}_t(\theta^*)
    &= - \sum_{i \in S_t} p_{t}(i|S_t, \theta^*) \log p_{t}(i|S_t, \theta) + \sum_{i \in S_t} p_{t}(i|S_t, \theta^*) \log p_{t}(i|S_t, \theta^*)\\
    &= \sum_{i \in S_t} p_{t}(i|S_t, \theta^*) \left[\log p_{t}(i|S_t, \theta^*) - \log p_{t}(i|S_t, \theta) \right]\\
    &= \sum_{i \in S_t} p_{t}(i|S_t, \theta^*) \log \frac{p_{t}(i|S_t, \theta^*)}{p_{t}(i|S_t, \theta)}\\
    &\geq 0
\end{align*}
where $\sum_{i \in S_t} p_{t}(i|S_t, \theta^*) \log \frac{p_{t}(i|S_t, \theta^*)}{p_{t}(i|S_t, \theta)}$ is the Kullback-Leibler divergence between two distributions which is always non-negative.
\endproof

\begin{lemma}\label{lemma:f_grad_upperbound}
For any positive-semidefinte matrix $V$,
\begin{align*}
    \|G_t(\theta)\|^2_V \leq  4 \max_{i \in S_t}\| x_{ti}\|^2_V
\end{align*}
\end{lemma}

\proof{}
For any positive-semidefinte matrix $V$
\begin{align*}
    (z_i - z_j)^\top V(z_i - z_j)^\top = z_i^\top V z_i + z_j^\top V z_j - z_i^\top V z_j - z_j^\top V z_i \geq 0
\end{align*}
which implies $ z_i^\top V z_i + z_j^\top V z_j \geq z_i^\top V z_j + z_j^\top V z_i$. We let $z_i := \left(  p_{t}(i|S_t, \theta) - y_{ti} \right) x_{ti}$

\begin{align*}
    \|G_t(\theta)\|^2_V &=  \sum_{i \in S_t} \sum_{j \in S_t} \left(  p_{t}(i|S_t, \theta) - y_{ti} \right)\left(  p_{tj}(S_t, \theta) - y_{tj} \right) x_{ti}^\top V x_{tj}\\
    &= \sum_{i \in S_t} \left(  p_{t}(i|S_t, \theta) - y_{ti} \right)^2 x_{ti}^\top V x_{ti} \\
    &\quad + \frac{1}{2}\sum_{i \in S_t} \sum_{j \in S_t} \left(  p_{t}(i|S_t, \theta) - y_{ti} \right)\left(  p_{tj}(S_t, \theta) - y_{tj} \right) (x_{ti}^\top V x_{tj} + x_{tj}^\top V x_{ti})\\
    &\leq \sum_{i \in S_t} \left(  p_{t}(i|S_t, \theta) - y_{ti} \right)^2 x_{ti}^\top V x_{ti}\\
    &\quad + \frac{1}{2}\sum_{i \in S_t} \sum_{j \in S_t} \left[\left(  p_{t}(i|S_t, \theta) - y_{ti} \right)^2 x_{ti}^\top V x_{tj} + \left(  p_{tj}(S_t, \theta) - y_{tj} \right)^2 x_{tj}^\top V x_{ti} \right]\\
    &= \sum_{i \in S_t} \left(  p_{t}(i|S_t, \theta) - y_{ti} \right)^2 x_{ti}^\top V x_{ti} +  \sum_{i \in S_t} \left(  p_{t}(i|S_t, \theta) - y_{ti} \right)^2 x_{ti}^\top V x_{ti}\\
    &= 2\sum_{i \in S_t} \left(  p_{t}(i|S_t, \theta) - y_{ti} \right)^2 x_{ti}^\top V x_{ti}\\
    &\leq 4 \max_{i \in S_t} x_{ti}^\top V x_{ti}\\
    &= 4 \max_{i \in S_t}\| x_{ti}\|^2_V
\end{align*}
\endproof

\begin{lemma}\label{lemma:f_grad_MG_bound}
With a probability at least $1-\delta$,
\begin{align*}
    \sum_{t'=T_0+1}^t\left[ \bar{G}_{t'}(\hat{\theta}_{t'}) - G_{t'}(\hat{\theta}_{t'})\right]^\top (\hat{\theta}_{t'} - \theta^* ) \leq \frac{\kappa}{4}\sum_{t'=T_0+1}^t\|\theta^* - \hat{\theta}_{t'} \|^2_{W_{t'}} + \left(\frac{4}{\kappa} + \frac{8}{3}\right) \log\left( \frac{\lceil 2\log_2 \frac{tK}{2}\rceil t^2}{\delta} \right) + 2
\end{align*}
\end{lemma}
\proof{}
First, notice that $\left[ \bar{G}_{t'}(\hat{\theta}_{t'}) - G_{t'}(\hat{\theta}_{t'})\right]^\top (\hat{\theta}_{t'} - \theta^* )$ is a martingale difference sequence. Also, we have

\begin{align*}
    \left|\left[ \bar{G}_{t'}(\hat{\theta}_{t'}) - G_{t'}(\hat{\theta}_{t'})\right]^\top (\hat{\theta}_{t'} - \theta^* ) \right| 
    &\leq
    \left|\left[ \bar{G}_{t'}(\hat{\theta}_{t'}) \right]^\top (\hat{\theta}_{t'} - \theta^* ) \right| +  \left|\left[ G_{t'}(\hat{\theta}_{t'})\right]^\top (\hat{\theta}_{t'} - \theta^* ) \right|\\
    &\leq
    \left\| \bar{G}_{t'}(\hat{\theta}_{t'}) \right \| \left\|\hat{\theta}_{t'} - \theta^*  \right\| +  \left\| G_{t'}(\hat{\theta}_{t'}) \right\| \left\| \hat{\theta}_{t'} - \theta^* \right\|\\
    &\leq 2\sqrt{2} \| \hat{\theta}_{t'} - \theta^*  \|
\end{align*}
where the last inequality is from the fact that $\|G_t(\theta)\| = \|\sum_{i \in S_t} \left(  p_{t}(i|S_t, \theta) - y_{ti} \right) x_{ti}\| \leq \sqrt{2}$ for any $\theta$. Also, note that for large enough $t'$ (i.e. after the random initialization), we have $ \| \hat{\theta}_{t'} - \theta^*  \| \leq 1$. Hence, we have
\begin{align*}
    \left|\left[ \bar{G}_{t'}(\hat{\theta}_{t'}) - G_{t'}(\hat{\theta}_{t'})\right]^\top (\hat{\theta}_{t'} - \theta^* ) \right| \leq 2\sqrt{2}.
\end{align*}
We define the martingale $M_t := \sum_{t' = 1}^t \left[ \bar{G}_{t'}(\hat{\theta}_{t'}) - G_{t'}(\hat{\theta}_{t'})\right]^\top (\hat{\theta}_{t'} - \theta^* )$. And, we also define $\Sigma_t$ as
\begin{align*}
    \Sigma_t &:= \sum_{t' = 1}^t \mathbb{E}_{y_{t'}} \left[ \left( \left[ \bar{G}_{t'} (\hat{\theta}_{t'}) - G_{t'} (\hat{\theta}_{t'})\right]^\top (\hat{\theta}_{t'} - \theta^*) \right)^2 \right]\\
    &\leq \sum_{t' = 1}^t \mathbb{E}_{y_{t'}} \left[ \left(  G_{t'} (\hat{\theta}_{t'})^\top (\hat{\theta}_{t'} - \theta^*) \right)^2 \right]\\
    &\leq \sum_{t' = 1}^t \sum_{i \in S_{t'}}\left( x_{t' i}^\top(\hat{\theta}_{t'} - \theta^*) \right)^2 \\
    &=  \sum_{t' = 1}^t \| \hat{\theta}_{t'} - \theta^* \|^2_{W_{t'}} := B_t
\end{align*}
Note that $B_t$, the upper bound for $\Sigma_t$, is a random variable, so we cannot directly apply Bernstein's inequality to $M_t$. Instead, we consider two cases (i) $B_t \leq \frac{4}{tK}$ and (ii) $B_t > \frac{4}{tK}$.

\subsubsection*{Case (i)}
Let's assume
 $\displaystyle B_t = \sum_{t' = 1}^t \| \hat{\theta}_{t'} - \theta^* \|^2_{W_{t'}} \leq \frac{4}{tK}$. Then we have
\begin{align*}
    M_t &= \sum_{t' = 1}^t  \left[ \bar{G}_{t'}(\hat{\theta}_{t'}) - G_{t'}(\hat{\theta}_{t'})\right]^\top (\hat{\theta}_{t'} - \theta^* ) \\
     &= \sum_{t' = 1}^t  \sum_{i \in S_{t'}} (y_{t' i} - p(S_t, \theta^*)) x_{t' i}^\top  (\hat{\theta}_{t'} - \theta^* ) \\
    &\leq \sum_{t' = 1}^t \sum_{i \in S_{t'}} | x_{t' i}^\top ( \hat{\theta}_t - \theta^* ) | \\
    &\leq \sqrt{ t K \sum_{t' = 1}^t \sum_{i \in S_{t'}} \left( x_{t' i}^\top ( \hat{\theta}_t - \theta^* ) \right)^2}\\
    &\leq 2.
\end{align*}

\subsubsection*{Case (ii)}
Let's assume
 $\displaystyle B_t =  \sum_{t' = 1}^t \| \hat{\theta}_{t'} - \theta^* \|^2_{W_{t'}} > \frac{4}{tK}$. Note that we have both a lower and upper bounds for $B_t$, i.e., $\frac{4}{tK} < B_t \leq tK$. Then we can use the peeling process \citep{bartlett2005local}.
 \begin{align*}
     \mathbb{P}\left( M_t \geq 2 \sqrt{\eta_t B_t} + \frac{8\eta_t}{3}\right)
     &= \mathbb{P}\left( M_t \geq 2 \sqrt{\eta_t B_t} + \frac{8\eta_t}{3}, \frac{4}{tK} < B_t \leq tK\right)\\
     &= \mathbb{P}\left( M_t \geq 2 \sqrt{\eta_t B_t} + \frac{8\eta_t}{3}, \frac{4}{tK} < B_t \leq tK, \Sigma_t \leq B_t \right)\\
     &\leq \sum_{j=1}^m \mathbb{P}\left( M_t \geq 2 \sqrt{\eta_t B_t} + \frac{8\eta_t}{3}, \frac{4\cdot2^{j-1}}{tK} < B_t \leq \frac{4\cdot2^{j}}{tK}, \Sigma_t \leq B_t \right)\\
    &\leq \sum_{j=1}^m \mathbb{P}\left( M_t \geq \sqrt{\eta_t \frac{8\cdot2^{j}}{tK}} + \frac{8\eta_t}{3}, \Sigma_t \leq \frac{4\cdot2^{j}}{tK} \right)\\
    &\leq m \exp(-\eta_t)
 \end{align*}
where $m = \lceil 2\log_2 \frac{tK}{2} \rceil$, and the last inequality is from Bernstein's inequality for martingales. 
Combining with the result in Cases (i) and (ii), letting $\eta_t = \log\frac{mt^2}{\delta} = \log\frac{\lceil 2\log_2 \frac{tK}{2} \rceil t^2}{\delta}$ and taking the union bound over $t$, we have with probability at least $1-\delta$
\begin{align*}
    M_t = \sum_{t' = 1}^t \left[ \bar{G}_{t'}(\hat{\theta}_{t'}) - G_{t'}(\hat{\theta}_{t'})\right]^\top (\hat{\theta}_{t'} - \theta^* ) \leq 2 \sqrt{\eta_t \sum_{t'=T_0+1}^t\|\theta^* - \hat{\theta}_{t'} \|^2_{W_{t'}}} + \frac{8\eta_t}{3} + 2.
\end{align*}
Then we apply $uv \leq cu^2 + v^2/(4c)$ to the second term on the right hand side with $c = \frac{2}{\kappa}$.
\begin{align*}
    \sqrt{\eta_t \sum_{t'=T_0+1}^t\|\theta^* - \hat{\theta}_{t'} \|^2_{W_{t'}}}
    \leq \frac{2\eta_t}{\kappa} + \frac{\kappa}{8}\sum_{t'=T_0+1}^t\|\theta^* - \hat{\theta}_{t'} \|^2_{W_{t'}}
\end{align*}
Then we have
\begin{align*}
    \sum_{t'=T_0+1}^t\left[ \bar{G}_{t'}(\hat{\theta}_{t'}) - G_{t'}(\hat{\theta}_{t'})\right]^\top (\hat{\theta}_{t'} - \theta^* ) \leq \frac{\kappa}{4}\sum_{t'=T_0+1}^t\|\theta^* - \hat{\theta}_{t'} \|^2_{W_{t'}} + \left(\frac{4}{\kappa} + \frac{8}{3}\right)\eta_t + 2
\end{align*}
\hfill
\endproof

\subsection{Proof of Lemma~\ref{lemma:online_update}}
\proof{}
From Lemma~\ref{lemma:f_taylor_exp}, we have
\begin{align*}
   f_t(\hat{\theta}_t) \leq  f_t(\theta^*) + G_t(\hat{\theta}_t)^\top (\hat{\theta}_t - \theta^* ) - \frac{\kappa}{2} (\theta^* - \hat{\theta}_t)^\top \Big( \sum_{i\in S_t} x_{ti} x_{ti}^\top \Big)  (\theta^* - \hat{\theta}_t)
\end{align*}
Taking expectation over $y$ gives
\begin{align*}
    \bar{f}_t(\hat{\theta}_t) \leq  \bar{f}_t(\theta^*) +  \bar{G}_t(\hat{\theta}_t)^\top (\hat{\theta}_t - \theta^* ) - \frac{\kappa}{2} (\theta^* - \hat{\theta}_t)^\top \Big( \sum_{i\in S_t} x_{ti} x_{ti}^\top \Big)  (\theta^* - \hat{\theta}_t)
\end{align*}
Note that $\nabla \bar{f}_t(\theta) = \mathbb{E}_y[\nabla f_t(\theta)] = \bar{G}_t(\theta)$ by the Leibniz integral rule. Let $W_t :=  \sum_{i\in S_t} x_{ti} x_{ti}^\top$.
Since $\bar{f}_t(\theta) \geq \bar{f}_t(\theta^*)$ from Lemma~\ref{lemma:f_opt_ineq}, we have
\begin{align*}
        0&\leq \bar{f}_t(\hat{\theta}_t) - \bar{f}_t(\theta^*)\\
        &\leq   \bar{G}_t(\hat{\theta}_t)^\top (\hat{\theta}_t - \theta^* ) - \frac{\kappa}{2} \|\theta^* - \hat{\theta}_t \|^2_{W_t}\\
        &= G_t(\hat{\theta}_t)^\top (\hat{\theta}_t - \theta^* ) - \frac{\kappa}{2} \|\theta^* - \hat{\theta}_t \|^2_{W_t} + \left[ \bar{G}_t(\hat{\theta}_t) - G_t(\hat{\theta}_t)\right]^\top (\hat{\theta}_t - \theta^* )
\end{align*}

From Lemma~\ref{lemma:taylor_1st_2nd_bound}, we have $2G_t (\hat{\theta}_t)^\top (\hat{\theta}_t - \theta^*) 
    \leq \|G_t(\theta_t)\|^2_{V^{-1}_{t+1}} + \| \hat{\theta}_t - \theta^* \|^2_{V_{t+1}} - \| \hat{\theta}_{t+1} - \theta^* \|^2_{V_{t+1}}$. So we have
\begin{align*}
        0 
        &\leq  \frac{1}{2}\|G_t(\theta_t)\|^2_{V^{-1}_{t+1}} + \frac{1}{2}\| \hat{\theta}_t - \theta^* \|^2_{V_{t+1}} - \frac{1}{2}\| \hat{\theta}_{t+1} - \theta^* \|^2_{V_{t+1}} \\
        &\quad - \frac{\kappa}{2} \|\theta^* - \hat{\theta}_t \|^2_{W_t} + \left[ \bar{G}_t(\hat{\theta}_t) - G_t(\hat{\theta}_t)\right]^\top (\hat{\theta}_t - \theta^* ) \\
        &\leq  2\max_{i \in S_t}\| x_{ti}\|^2_{V^{-1}_{t+1}} + \frac{1}{2}\| \hat{\theta}_t - \theta^* \|^2_{V_{t+1}} - \frac{1}{2}\| \hat{\theta}_{t+1} - \theta^* \|^2_{V_{t+1}}\\
         &\quad- \frac{\kappa}{2} \|\theta^* - \hat{\theta}_t \|^2_{W_t} + \left[ \bar{G}_t(\hat{\theta}_t) - G_t(\hat{\theta}_t)\right]^\top (\hat{\theta}_t - \theta^* )
\end{align*}
where the last inequality is by Lemma~\ref{lemma:f_grad_upperbound}, $\|G_t(\theta)\|^2_{V^{-1}_{t+1}} \leq  4 \max_{i \in S_t}\| x_{ti}\|^2_{V^{-1}_{t+1}}$.
Note that since $V_{t+1} = V_{t} + \frac{\kappa}{2} \sum_{i \in S_t} x_{ti}x_{ti}^\top$, we have
\begin{align*}
    \| \hat{\theta}_t - \theta^* \|^2_{V_{t+1}} &= \| \hat{\theta}_t - \theta^* \|^2_{V_t} + \frac{\kappa}{2} (\hat{\theta}_t - \theta^*)^\top \left(\sum_{i \in S_t} x_{ti}x_{ti}^\top\right)(\hat{\theta}_t - \theta^*)\\
    &= \| \hat{\theta}_t - \theta^* \|^2_{V_t} + \frac{\kappa}{2}\| \hat{\theta}_t - \theta^* \|^2_{W_t}.
\end{align*}
Therefore, we can continue
\begin{align*}
    0&\leq 2\max_{i \in S_t}\| x_{ti}\|^2_{V^{-1}_{t+1}} + \frac{1}{2}\| \hat{\theta}_t - \theta^* \|^2_{V_t} + \frac{\kappa}{4}\| \hat{\theta}_t - \theta^* \|^2_{W_t} - \frac{1}{2}\| \hat{\theta}_{t+1} - \theta^* \|^2_{V_{t+1}} \\
    &\quad - \frac{\kappa}{2} \|\theta^* - \hat{\theta}_t \|^2_{W_t}
     + \left[ \bar{G}_t(\hat{\theta}_t) - G_t(\hat{\theta}_t)\right]^\top (\hat{\theta}_t - \theta^* )\\
    &= 2\max_{i \in S_t}\| x_{ti}\|^2_{V^{-1}_{t+1}} + \frac{1}{2}\| \hat{\theta}_t - \theta^* \|^2_{V_t}  - \frac{1}{2}\| \hat{\theta}_{t+1} - \theta^* \|^2_{V_{t+1}} 
        - \frac{\kappa}{4} \|\theta^* - \hat{\theta}_t \|^2_{W_t}\\
    &\quad + \left[ \bar{G}_t(\hat{\theta}_t) - G_t(\hat{\theta}_t)\right]^\top (\hat{\theta}_t - \theta^* )
\end{align*}
Hence, we have
\begin{align*}
    \| \hat{\theta}_{t+1} - \theta^* \|^2_{V_{t+1}}  
    &\leq \| \hat{\theta}_t - \theta^* \|^2_{V_t}  
    + 4\max_{i \in S_t}\| x_{ti}\|^2_{V^{-1}_{t+1}}  - \frac{\kappa}{2} \|\theta^* - \hat{\theta}_t \|^2_{W_t} \\
    &\quad + 2\left[ \bar{G}_t(\hat{\theta}_t) - G_t(\hat{\theta}_t)\right]^\top (\hat{\theta}_t - \theta^* ).
\end{align*}
Summing over $t$ gives
\begin{align*}
    \| \hat{\theta}_{t+1} - \theta^* \|^2_{V_{t+1}}   
    &\leq  \lambda_{\max}(V_{T_0}) + 4\sum_{t'=T_0+1}^t\max_{i \in S_{t'}}\| x_{t' i}\|^2_{V^{-1}_{\tau+1}} - \frac{\kappa}{2} \sum_{t'=T_0+1}^t\|\theta^* - \hat{\theta}_{t'} \|^2_{W_{t'}}\\
    &\quad + 2\sum_{t'=T_0+1}^t\left[ \bar{G}_{t'}(\hat{\theta}_{t'}) - G_{t'}(\hat{\theta}_{t'})\right]^\top (\hat{\theta}_{t'} - \theta^* )
\end{align*}
Now, we can use Lemma~\ref{lemma:f_grad_MG_bound} which shows with a probability at least $1-\delta$,
\begin{align*}
    &\sum_{t'=T_0+1}^t\left[ \bar{G}_{t'}(\hat{\theta}_{t'}) - G_{t'}(\hat{\theta}_{t'})\right]^\top (\hat{\theta}_{t'} - \theta^* )\\
    & \leq \frac{\kappa}{4}\sum_{t'=T_0+1}^t\|\theta^* - \hat{\theta}_{t'} \|^2_{W_{t'}} + \left(\frac{4}{\kappa} + \frac{8}{3}\right) \log\left( \frac{\lceil 2\log_2 \frac{tK}{2}\rceil t^2}{\delta} \right) + 2.
\end{align*}
We have with a probability at least $1-\delta$
\begin{align*}
    \| \hat{\theta}_{t+1} - \theta^* \|^2_{V_{t+1}}   
    &\leq  T_0 + 4\sum_{t'=T_0+1}^t\max_{i \in S_{t'}}\| x_{t' i}\|^2_{V^{-1}_{\tau+1}} + \left(\frac{8}{\kappa} + \frac{16}{3}\right) \log\left( \frac{\lceil 2\log_2 \frac{tK}{2}\rceil t^2}{\delta} \right) + 4\\
    &\leq  T_0 + \frac{8}{\kappa} d \log \left(  1 + \frac{T}{d } \right) + \left(\frac{8}{\kappa} + \frac{16}{3}\right) \log\left( \frac{\lceil 2\log_2 \frac{tK}{2}\rceil t^2}{\delta} \right) + 4
\end{align*}
where we apply Lemma~\ref{lemma:self_norm_MG_bound} to bound $\sum_{t'=1}^t\max_{i \in S_{t'}}\| x_{t' i}\|^2_{V^{-1}_{\tau+1}}$ in the last inequality. Note that $V_t$ in Algorithm~\ref{algo:UCB-MNL} and $V_t$ in Algorithm~\ref{algo:UCB-MNL-online-update} are different by the factor of $\frac{\kappa}{2}$, which results in additional $\frac{2}{\kappa}$ factor for the bound of $\sum_{t'=1}^t\max_{i \in S_{t'}}\| x_{t' i}\|^2_{V^{-1}_{\tau+1}}$.
\endproof

\subsection{Proof of Theorem~\ref{thm:expected_regret_ucb-mnl-online-update}}

\proof{}
Similar to the proof of Theorem~\ref{thm:expected_regret_ucb-mnl}, we first define the high probability event
\begin{definition}
Define the following event:
\begin{align*}
        \tilde{\mathcal{E}} &= \left\{\lambda_{\min}(V_{T_0}) \geq K, \enskip \|\Hat{\theta}_t - \theta^* \|_{V_t} \leq  \tilde{\alpha}_t, \forall t \leq T  \right\}
\end{align*}
where $\tilde{\alpha}_t$ is defined as Theorem~\ref{thm:expected_regret_ucb-mnl-online-update}.
\end{definition}
Then following steps equivalent the first few steps in the proof of Theorem~\ref{thm:expected_regret_ucb-mnl}, we have
\begin{align*}
    \mathcal{R}_T &\leq T_0 + \mathbb{E} \left[ \sum_{t=T_0+1}^T  \left ( \Tilde{R}_t(S_t) - R_t(S_t, \theta^*) \right) \mathbb{1}(\tilde{\mathcal{E}}) \right]
     + \mathbb{E} \left[ \sum_{t=T_0+1}^T   \left ( \Tilde{R}_t(S_t) - R_t(S_t, \theta^*) \right) \mathbb{1}(\tilde{\mathcal{E}}^c) \right]\\
    & \leq T_0 + \mathbb{E} \left[ \sum_{t=T_0 + 1}^T  \left ( \Tilde{R}_t(S_t) - R_t(S_t, \theta^*) \right) \mathbb{1}(\tilde{\mathcal{E}}) \right]
    +  \sum_{t=1}^T  \mathcal{O}(t^{-2})\\
    & \leq T_0 + \sum_{t=1}^T 2\tilde{\alpha}_T \max_{i \in S_t}   \| x_{ti} \|_{V^{-1}_{t}} + \mathcal{O}(1)
\end{align*}
Applying Cauchy-Schwarz inequality and Lemma~\ref{lemma:self_norm_MG_bound} for $\sum_{t=1}^T \max_{i \in S_t}    \| x_{ti} \|_{V^{-1}_{t}}^2$, we have
\begin{align*}
      \mathcal{R}_T &\leq  T_0 + 2\tilde{\alpha}_T \sqrt{  2 d T \log \left( T/d \right) }
      + \mathcal{O}(1)
\end{align*}
where $\tilde{\alpha}_T = \sqrt{ T_0 + \frac{8}{\kappa} d \log \left(  1 + \frac{T}{d } \right) + \left(\frac{8}{\kappa} + \frac{16}{3}\right) \log\left( \lceil 2\log_2 (TK/2)\rceil t^4 \right) + 4}$. 
\endproof

\vspace{1cm}

\section{Proof of Theorem~\ref{thm:normalityMLE}}

In this section, we present a finite-sample version of the asymptotic normality of the MLE for the MNL model. It is a generalization of Theorem 1 in \citep{li2017provably} to a multinomial setting.

\proof{}
Recall that the gradient of the negative log-likelihood of the MNL model is given by
\begin{equation*}
    \nabla_{\theta} \ell_n(\theta) = \sum_{t=1}^n \sum_{i \in S_t} (p_{t}(i|S_t, \theta)  - y_{ti}) x_{ti}
\end{equation*}
We define its conditional expectation $J_n(\theta)$ and will use this term throughout this section
\begin{definition}
Define the conditional expectation $\nabla_{\theta} \ell(\theta)$ as
\begin{align*}
J_n(\theta) := \mathbb{E}_y\left[ \nabla_{\theta} \ell_n(\theta) | \mathcal{F}_t\right]
= \sum_{t=1}^{n} \sum_{i \in S_t} \left( p_{t}(i|S_t, \theta) - p_{t}(i|S_t, \theta^*) \right)x_{ti}.
\end{align*}
\end{definition}
Notice that $J_n(\Hat{\theta}) = \sum^{n}_{t=1} \sum_{i \in S_t} \epsilon_{ti} x_{ti}$ since the choice of $\Hat{\theta}$ is given by the MLE. In other words, $\Hat{\theta}$ is given by the solution to the following:
\begin{equation*}
    \sum_{t = 1}^{n} \sum_{i \in S_t} \left(  p_{t}(i|S_t, \Hat{\theta}) - y_{ti} \right) x_{ti} = 0
\end{equation*}
Hence it follows that
\begin{align*}
    J_n(\Hat{\theta}) &= \sum_{t=1}^{n} \sum_{i \in S_t} \left( p_{t}(i|S_t, \Hat{\theta}) - p_{t}(i|S_t, \theta^*) \right)x_{ti}\\ 
     &= \sum_{t=1}^{n} \sum_{i \in S_t} \left( p_{t}(i|S_t, \Hat{\theta}) - y_{ti}\right)x_{ti} + \sum_{t=1}^{n} \sum_{i \in S_t} \left(y_{ti} - p_{t}(i|S_t, \theta^*) \right)x_{ti}\\
      &= 0 + \sum^{n}_{t=1} \sum_{i \in S_t} \epsilon_{ti} x_{ti}
\end{align*}
For convenience, define $Z_n := J_n(\Hat{\theta})$. 
For brevity, 
we will denote $p_{ti}(\theta) := p_{t}(i|S_t, \theta)$ when it is clear that $S_t$ is the assortment chosen at round $t$.

\subsection{Consistency of MLE}
In this section, we show the consistency of MLE $\hat{\theta}$.
For any $\theta_1, \theta_2 \in \mathbb{R}^d$, the mean value theorem implies that there exists $\Bar{\theta} = c\theta_1 + (1-c)\theta_2$ with $c \in (0,1)$. 
\begin{align*}
J_n(\theta_1) - J_n(\theta_2) &= \left[ \sum^{n}_{t=1} \sum_{i \in S_t} \sum_{j \in S_t} \nabla_j p_{ti}(\Bar{\theta}) x_{ti}x_{tj}^\top \right] (\theta_1 - \theta_2)\\
&= \sum^n_{t=1} \left[ \sum_{i \in S_t} p_{ti}(\Bar{\theta}) x_{ti}x_{ti}^\top - \sum_{i \in S_t}\sum_{j \in S_t} p_{ti}(\Bar{\theta}) p_{tj}(\Bar{\theta}) x_{ti}x_{tj}^\top \right](\theta_1 - \theta_2)
\end{align*}
Let $H_t := \sum_{i \in S_t} p_{ti}(\Bar{\theta}) x_{ti}x_{ti}^\top - \sum_{i,j \in S_t} p_{ti}(\Bar{\theta}) p_{tj}(\Bar{\theta}) x_{ti}x_{tj}^\top$. Notice $H_t$ is a Hessian of a negative log-likelihood which is convex. Hence, $H_t$ is positive semidefinite. Also note that
\begin{align*}
    (x_i - x_j)(x_i - x_j)^\top = x_i x_i^\top + x_j x_j^\top - x_i x_j^\top - x_j x_i^\top \succeq 0
\end{align*}
which implies $ x_i x_i^\top + x_j x_j^\top \succeq x_i x_j^\top + x_j x_i^\top$. Therefore, it follows that 
\begin{align*}
    H_t &= \sum_{i \in S_t} p_{ti}(\Bar{\theta}) x_{ti}x_{ti}^\top - \sum_{i \in S_t}\sum_{j \in S_t} p_{ti}(\Bar{\theta}) p_{tj}(\Bar{\theta}) x_{ti}x_{tj}^\top\\
    &= \sum_{i \in S_t} p_{ti}(\Bar{\theta}) x_{ti}x_{ti}^\top - \frac{1}{2}\sum_{i \in S_t}\sum_{j \in S_t} p_{ti}(\Bar{\theta}) p_{tj}(\Bar{\theta}) \left(x_{ti}x_{tj}^\top + x_{tj}x_{ti}^\top \right)\\
    &\succeq \sum_{i \in S_t} p_{ti}(\Bar{\theta}) x_{ti}x_{ti}^\top - \frac{1}{2}\sum_{i \in S_t}\sum_{j \in S_t} p_{ti}(\Bar{\theta}) p_{tj}(\Bar{\theta}) \left(x_{ti}x_{ti}^\top + x_{tj}x_{tj}^\top \right)\\
    &=\sum_{i \in S_t} p_{ti}(\Bar{\theta}) x_{ti}x_{ti}^\top - \sum_{i \in S_t}\sum_{j \in S_t} p_{ti}(\Bar{\theta}) p_{tj}(\Bar{\theta}) x_{ti}x_{ti}^\top \\
    &=\sum_{i \in S_t} p_{ti}(\Bar{\theta}) \left(1-\sum_{j \in S_t}  p_{tj}(\Bar{\theta}) \right) x_{ti}x_{ti}^\top\\
    &=\sum_{i \in S_t} p_{ti}(\Bar{\theta}) p_{t0}(\Bar{\theta}) x_{ti}x_{ti}^\top 
\end{align*}
where $p_{t0}(\Bar{\theta})$ is the probability of choosing the no purchase option under parameter $\Bar{\theta}$. Define $\mathcal{H}_n(\theta) :=  \sum^n_{t=1} \sum_{i \in S_t} p_{ti}(\Bar{\theta}) p_{t0}(\Bar{\theta}) x_{ti}x_{ti}^\top $.
Then, we can write
\begin{align}\label{eq:G_H_lower_bound}
    J_n(\theta_1) - J_n(\theta_2) &= \left[  \sum^n_{t=1} H_t  \right] (\theta_1 - \theta_2) \notag\\
    &\geq \left[  \sum^n_{t=1} \sum_{i \in S_t} p_{ti}(\Bar{\theta}) p_{t0}(\Bar{\theta}) x_{ti}x_{ti}^\top  \right] (\theta_1 - \theta_2) \notag\\
    &= \mathcal{H}_n(\Bar{\theta}) (\theta_1 - \theta_2) 
\end{align}

If $\Bar{\theta} \in \mathcal{B}_\eta := \{ \theta : \|\theta - \theta^*\| \leq \eta \}$ with some $\eta > 0 $, then $p_{ti}(\Bar{\theta}) p_{t0}(\Bar{\theta}) \geq \kappa_\eta$, where $\kappa_\eta$ is defined as $\kappa_\eta := \inf_{\theta \in \mathcal{B}_\eta, i \in S, S \in \mathcal{S}} p_{ti}(\theta)p_{t0}(\theta) > 0$. Then since  $\mathcal{H}_n(\Bar{\theta}) \succeq \kappa_\eta V_n$, we have
\begin{align*}
    (\theta_1 - \theta_2)^\top ( J_n(\theta_1) - J_n(\theta_2)) \geq (\theta_1 - \theta_2)^\top (\kappa_\eta V_n) (\theta_1 - \theta_2) > 0
\end{align*}
for any $\theta_1 \neq \theta_2$. Therefore, $ J_n(\theta)$ is an injection from $\mathbb{R}^d$ to $\mathbb{R}^d$.
Note that $\mathcal{B}_\eta$ is a convex set. Hence, if $\theta_1, \theta_2 \in \mathcal{B}_\eta$, then also $\Bar{\theta} \in \mathcal{B}_\eta$.
Also, by the definition of $J_n(\theta)$, we have $J_n(\theta^*) = 0$. Then, for any $\theta \in \mathcal{B}_\eta$, it follows that
\begin{align}
    \| J_n(\theta) \|_{V^{-1}_n}^2 &= \| J_n(\theta) - J_n(\theta^*) \|_{V^{-1}_n}^2 \notag\\
    &\geq (\theta - \theta^*)^\top \mathcal{H}_n(\Bar{\theta}) V^{-1}_n \mathcal{H}_n(\Bar{\theta}) (\theta - \theta^*) \notag\\
    &\geq \kappa^2_\eta \lambda_{\min}(V_n)\| \theta - \theta^* \|^2 \label{eq:theta_G_bound}
\end{align}
where the first inequality is due to \eqref{eq:G_H_lower_bound} and the second inequality is again from the fact that $\mathcal{H}_n(\Bar{\theta}) \succeq \kappa_\eta V_n$. Now, we need an upper-bound for $ \| J_n(\theta) \|_{V^{-1}_n}$.
From Lemma~\ref{lemma:G_bound}, we have
\begin{align}\label{eq:J_upper_bound}
    \| J_n(\hat{\theta}) \|_{V^{-1}_n} \leq 2\sqrt{2d + \log \frac{1}{\delta}}
\end{align}
with probability at least $1-\delta$. Then, we combine with \eqref{eq:theta_G_bound} and have
\begin{equation*}
    \| \Hat{\theta} - \theta^* \| \leq \frac{2}{\kappa_\eta} \sqrt{\frac{2d + \log(1/\delta)}{\lambda_{\min}(V_n)}} 
\end{equation*}
Then since $\kappa = \kappa_1$ where $\kappa \leq \min_{\|\theta - \theta^*\| \leq 1} p_{ti}(S, \theta)p_{t0}(S, \theta) $ defined in Assumption~\ref{assum:prob_bound}, we have
\begin{equation}\label{eq:theta_bound}
    \| \Hat{\theta} - \theta^* \| \leq \frac{2}{\kappa} \sqrt{\frac{2d + \log(1/\delta)}{\lambda_{\min}(V_n)}} \leq 1
\end{equation}
as long as $\lambda_{\min}(V_n) \geq \frac{4}{\kappa^2} (2d + \log\frac{1}{\delta})$. 

\subsection{Normality of MLE}
In this section, we show the normality result of MLE $\hat{\theta}$.
For the rest of the section, we assume \eqref{eq:J_upper_bound} holds.
First, we define $F, L$ and $E$ which are defined as:
\begin{align*}
&F(\theta) := \sum_{t=1}^n \sum_{i \in S_t} p_{ti}(\theta) x_{ti} x_{ti}^\top -\sum_{t=1}^n \sum_{i \in S_t}\sum_{j \in S_t} p_{ti}(\theta) p_{tj}(\theta) x_{ti} x_{tj}^\top\\
&L := F(\theta^*) = \sum_{t=1}^n \sum_{i \in S_t} p_{ti}(\theta^*) x_{ti} x_{ti}^\top -\sum_{t=1}^n \sum_{i \in S_t}\sum_{j \in S_t} p_{ti}(\theta^*) p_{tj}(\theta^*) x_{ti} x_{tj}^\top\\
&E := F(\Tilde{\theta}) - F(\theta^*)
\end{align*}
where $\Tilde{\theta} := c\theta^* + (1-c)\Hat{\theta}$ for some constant $c \in (0, 1)$.
Then, it follows that
\begin{align*}
    Z_n &= J_n(\Hat{\theta}) = J_n(\Hat{\theta}) - J_n(\theta^*)\\
    &= (L + E)(\Hat{\theta} - \theta^*).
\end{align*}
Hence, for any $x \in \mathbb{R}^2$, we can write
\begin{align}\label{eq:x_theta_diff}
    x^\top (\Hat{\theta} - \theta^*) &= x^\top (L + E)^{-1} Z_n \notag\\
    &= x^\top L^{-1}Z_n - x^\top L^{-1} E (L+E)^{-1}Z_n.
\end{align}
Note that  $(L+E)$ is a non-singular matrix, hence $(L+E)$ is invertible. Here, the key element is controlling the matrix $E$. Note that if $\Hat{\theta}$ and $\theta^*$ are close (so $\tilde{\theta}$ and $\theta^*$ are also close), elements in $E$ are small. 

\subsection*{Bounding Matrix $E$}
First, we further decompose $E$ into two summations, $E_1$ and $E_2$
\begin{align}\label{eq:E_definition}
    E &= \underbrace{\sum_{t=1}^n \sum_{i \in S_t} \left( p_{ti}(\Tilde{\theta}) - p_{ti}(\theta^*) \right) x_{ti} x_{ti}^\top}_{E_1} - 
    \underbrace{\sum_{t=1}^n \sum_{i \in S_t}\sum_{j \in S_t} \left( p_{ti}(\Tilde{\theta}) p_{tj}(\Tilde{\theta}) - p_{ti}(\theta^*) p_{tj}(\theta^*) \right) x_{ti} x_{tj}^\top}_{E_2}
\end{align}
We first bound the first summation $E_1$. Note that 
\begin{align*}
    E_1 &= \sum_{t=1}^n \sum_{i \in S_t} \left( p_{ti}(\Tilde{\theta}) - p_{ti}(\theta^*) \right) x_{ti} x_{ti}^\top\\
    &= \sum_{t=1}^n \sum_{i \in S_t} \sum_{j \in S_t} \nabla_j p_{ti}(\theta_1) x_{tj}^\top (\Hat{\theta} - \theta^*) x_{ti} x_{ti}^\top\\
    &= \sum_{t=1}^n \sum_{i \in S_t}  p_{ti}(\theta_1) x_{ti}^\top (\Hat{\theta} - \theta^*) x_{ti} x_{ti}^\top - \sum_{t=1}^n \sum_{i \in S_t} \sum_{j \in S_t} p_{ti}(\theta_1) p_{tj}(\theta_1) x_{tj}^\top (\Hat{\theta} - \theta^*) x_{ti} x_{ti}^\top
\end{align*}
where the second equality is by the mean value theorem for some $\theta_1 := c_1 \theta^* + (1 - c_1)\Hat{\theta}$ with $c_1 \in (0,1)$. Note that the mean value theorem is applied to $\Tilde{\theta}$ and $\theta^*$, and since $\Tilde{\theta}$ is a convex combination of $\Hat{\theta}$ and $\theta^*$, we can find such $c_1$.
Then it follows that
\begin{align*}
    E_1  &= \sum_{t=1}^n \sum_{i \in S_t}  p_{ti}(\theta_1) \left( x_{ti}^\top (\Hat{\theta} - \theta^*) - \sum_{j \in S_t} p_{tj}(\theta_1) x_{tj}^\top (\Hat{\theta} - \theta^*) \right) x_{ti} x_{ti}^\top\\
    &\leq \sum_{t=1}^n \sum_{i \in S_t}  p_{ti}(\theta_1) \left\| x_{ti} - \sum_{j \in S_t} p_{tj}(\theta_1) x_{tj} \right\| \| \Hat{\theta} - \theta^* \| x_{ti} x_{ti}^\top\\
    &\leq \sum_{t=1}^n \sum_{i \in S_t}  2 p_{ti}(\theta_1)  \| \Hat{\theta} - \theta^* \| x_{ti} x_{ti}^\top
\end{align*}
where we have used the assumption that $\| x_{ti}\| < 1$ for all $i$ and $t$ for the last inequality.
Then, for any $x \in \mathbb{R}^d \setminus \{0\}$, we have
\begin{align*}
    x^\top L^{-1/2} E_1 L^{-1/2} x 
    &\leq  \sum_{t=1}^n \sum_{i \in S_t}  2 p_{ti}(\theta_1)  \| \Hat{\theta} - \theta^* \| \| x^\top L^{-1/2} x_{ti} \|^2\\
    &\leq  \sum_{t=1}^n \sum_{i \in S_t}  2 \| \Hat{\theta} - \theta^* \| \| x^\top L^{-1/2} x_{ti} \|^2\\
    &\leq 2\| \Hat{\theta} - \theta^* \| \left(  x^\top L^{-1/2} \left( \sum_{t=1}^n \sum_{i \in S_t}   x_{ti} x_{ti}^\top  \right) L^{-1/2} x \right)\\
    &\leq \frac{2}{\kappa}  \| \Hat{\theta} - \theta^* \|  \| x \|^2 
\end{align*}
where the third inequality follows from the fact that $p_{ti}(\theta_1) \leq 1$. Therefore, combining with (\ref{eq:theta_bound}) it follows that
\begin{equation}\label{eq:HE1H_bound}
    \|L^{-1/2} E_1 L^{-1/2}\| \leq \frac{2}{\kappa}  \| \Hat{\theta} - \theta^* \| \leq \frac{4}{\kappa^2} \sqrt{\frac{2d + \log(1/\delta)}{\lambda_{\min}(V_n)}}.
\end{equation}

Similarly, we can bound the second summation $E_2$ in \eqref{eq:E_definition}. Again by the mean value theorem, for some $\theta_2 := c_2 \theta^* + (1 - c_2)\Hat{\theta}$ with $c_2 \in (0,1)$ we have
\begin{align*}
    E_2 &= \sum_{t=1}^n \sum_{i \in S_t}\sum_{j \in S_t} \left( p_{ti}(\Tilde{\theta}) p_{tj}(\Tilde{\theta}) - p_{ti}(\theta^*) p_{tj}(\theta^*) \right) x_{ti} x_{tj}^\top\\
    &= \sum_{t=1}^n \sum_{i \in S_t} \sum_{j \in S_t} \sum_{k \in S_t} \nabla_k [p_{ti}(\theta_2) p_{tj}(\theta_2)] x_{t,k}^\top (\Hat{\theta} - \theta^*) x_{ti} x_{ti}^\top.
\end{align*}
Let $p_{ti} = p_{ti}(\theta_2)$ for brevity. Then, it follows that
\begin{align*}
    E_2 &= \sum_{t=1}^n \sum_{i \in S_t} \sum_{j \in S_t} \sum_{k \in S_t} \nabla_k [p_{ti} p_{tj}] x_{t,k}^\top (\Hat{\theta} - \theta^*) x_{ti} x_{tj}^\top\\
    &= \sum_{t=1}^n \sum_{i \in S_t} \sum_{j \in S_t} \left[ p_{tj} \left(p_{ti} x_{ti} - \sum_{k \in S_t} p_{ti} p_{t,k} x_{t,k} \right) + p_{ti} \left(p_{tj} x_{tj} - \sum_{k \in S_t} p_{tj} p_{t,k} x_{t,k} \right) \right]^\top (\Hat{\theta} - \theta^*) x_{ti} x_{tj}^\top \notag\\
    &= \sum_{t=1}^n \sum_{i \in S_t} \sum_{j \in S_t} p_{ti} p_{tj} \left[ (x_{ti}+x_{tj}) -2 \sum_{k \in S_t} p_{t,k} x_{t,k} \right]^\top (\Hat{\theta} - \theta^*) x_{ti} x_{tj}^\top\\
    &\leq \sum_{t=1}^n \sum_{i \in S_t} \sum_{j \in S_t} p_{ti} p_{tj} \left\| (x_{ti}+x_{tj}) -2 \sum_{k \in S_t} p_{t,k} x_{t,k} \right\| \|\Hat{\theta} - \theta^*\| x_{ti} x_{tj}^\top\\
    &\leq \sum_{t=1}^n \sum_{i \in S_t} \sum_{j \in S_t} 4 p_{ti} p_{tj} \| \Hat{\theta} - \theta^* \|  x_{ti} x_{tj}^\top\\
    &= \sum_{t=1}^n \sum_{i \in S_t} 4 p_{ti} \left(1- p_{t0}\right) \| \Hat{\theta} - \theta^* \|  x_{ti} x_{ti}^\top
\end{align*}
where $p_{t0} = p_{t0}(\theta_2)$ is a probability of choosing an outside option.
Then, for any $x \in \mathbb{R}^d \setminus \{0\}$, we have
\begin{align*}
    x^\top L^{-1/2} E_2 L^{-1/2} x 
    &\leq  \sum_{t=1}^n \sum_{i \in S_t}  4 p_{ti}(\theta_2) \left(1- p_{t0}(\theta_2)\right)  \|\Hat{\theta} - \theta^*\| \| x^\top L^{-1/2} x_{ti} \|^2\\
    &\leq  \sum_{t=1}^n \sum_{i \in S_t}  4 \|\Hat{\theta} - \theta^*\| \| x^\top L^{-1/2} x_{ti} \|^2\\
    &\leq 4\| \Hat{\theta} - \theta^* \| \left(  x^\top L^{-1/2} \left( \sum_{t=1}^n \sum_{i \in S_t}   x_{ti} x_{ti}^\top  \right) L^{-1/2} x \right)\\
    &\leq \frac{4}{\kappa}  \| \Hat{\theta} - \theta^* \|  \| x \|^2 
\end{align*}

Similarly, combining with (\ref{eq:theta_bound}) it follows that
\begin{equation}\label{eq:HE2H_bound}
    \|L^{-1/2} E_2 L^{-1/2}\| \leq \frac{4}{\kappa}  \| \Hat{\theta} - \theta^* \| \leq \frac{8}{\kappa^2} \sqrt{\frac{2d + \log(1/\delta)}{\lambda_{\min}(V_n)}}.
\end{equation}

Hence, combining (\ref{eq:HE1H_bound}) and (\ref{eq:HE2H_bound}), we have with $\lambda_{\min}(V_n) \geq \frac{24^2}{\kappa^4} (d + \log\frac{1}{\delta})$
\begin{align}\label{eq:HEH_bound}
\|L^{-1/2} E L^{-1/2}\| &= \|L^{-1/2} (E_1 - E_2) L^{-1/2}\| \notag\\
 &\leq \|L^{-1/2} E_1 L^{-1/2}\| + \|L^{-1/2} E_2 L^{-1/2}\| \notag\\
 &\leq \frac{12}{\kappa^2} \sqrt{\frac{2d + \log(1/\delta)}{\lambda_{\min}(V_n)}}
 \leq \frac{1}{2}.  
\end{align}

\subsection*{Bounding the Prediction Error $x^\top (\Hat{\theta} - \theta^*)$}
Recall from \eqref{eq:x_theta_diff} that the prediction error for any $x \in \mathbb{R}^2$ can be written as
\begin{align*}
    x^\top (\Hat{\theta} - \theta^*) &= x^\top L^{-1}Z_n - x^\top L^{-1} E (L+E)^{-1}Z_n.
\end{align*}
First, we bound the first term $x^\top L^{-1}Z_n$ in (\ref{eq:x_theta_diff}). 
We start with providing the following definitions for the ease of our presentation:
\begin{alignat*}{2}
    X_{t} &:= [x_{t1}; x_{t2}; ...; x_{t|S_t|}]^\top \in \mathbb{R}^{|S_t| \times d}\\
    D &:= [X_{1}; X_{2}; ...; X_{n}]^\top \in \mathbb{R}^{ (\sum_t|S_t|) \times d}\\
    \mathcal{E}_t &:= [\epsilon_{t1}, \epsilon_{t2}, ..., \epsilon_{t|S_t|}]^\top \in \mathbb{R}^{|S_t|} 
\end{alignat*}

Then we use the notations above to see $|x^\top L^{-1} Z_n| = \left| \sum_t x^\top L^{-1} X_t^\top \mathcal{E}_t \right|$. 
For independent samples, $X_t$ and $\mathcal{E}_t$ are independent. Therefore, for each $t$
\begin{align*}
    \mathbb{E}\left[ x^\top L^{-1} X_t^\top \mathcal{E}_t \right] 
    = \mathbb{E}\left[ \sum_{i \in S_t} x^\top L^{-1} x_{ti} \epsilon_{ti} \right]
    = \sum_{i \in S_t}  \mathbb{E}\left[ x^\top L^{-1} x_{ti} \right] \mathbb{E}[\epsilon_{ti} ]
    &= 0
\end{align*}
since $\mathbb{E}[\epsilon_{ti}] = 0$ for all $t, i$. Also, we have
\begin{align*}
    \left| x^\top L^{-1} X_t^\top \mathcal{E}_t \right| \leq \| x^\top L^{-1} X_t^\top \| \| \mathcal{E}_t\| \leq \sqrt{2}\| x^\top L^{-1} X_t^\top \|
\end{align*}
where we use $\| \mathcal{E}_t\| \leq \sqrt{2}$. We also know $\| x^\top L^{-1} X_t^\top \|$ is bounded since both $X_t$ and $x$ are bounded. Hence, each $x^\top L^{-1} X_t^\top \mathcal{E}_t$ is therefore a bounded random variable. 
This allows us to apply Hoeffding inequality for bounded random variables in Lemma~\ref{lemma:hoeffding}. 
\begin{align}
    \mathbb{P} \left( |x^\top L^{-1} Z_n| \geq \nu \right ) 
    &= \mathbb{P} \left(\left| \sum_{t=1}^n x^\top L^{-1} X_t^\top \mathcal{E}_t \right| \geq \nu \right) \notag\\
    &\leq 2 \exp \left\{ -\frac{2 \nu^2}{\sum_{t=1}^n \left(2\sqrt{2}\| x^\top L^{-1} X^\top_t \| \right)^2} \right\} \notag\\
    &= 2 \exp \left\{ -\frac{\nu^2}{ 4\| x^\top L^{-1} D^\top \|^2} \right\} \notag\\
    &\leq 2 \exp \left\{ -\frac{\kappa^2 \nu^2}{ 4\| x \|^2_{V^{-1}_n}} \right\} \label{eq:xHZ_bound}
\end{align}
where the second equality follows from the definition of $D_t$, i.e.,
\begin{align*}
    \sum_{t=1}^n \| x^\top L^{-1} X^\top_t \|^2 = \sum_{t=1}^n  x^\top L^{-1} X^\top_t X_t L^{-1} x = x^\top L^{-1} D^\top D L^{-1} x = \| x^\top L^{-1} D^\top \|^2 \,.
\end{align*}
And, the last inequality follows from the fact that $L \succeq \kappa V = \kappa D^\top D$ and combining it with the following:
\begin{align*}
     \| x^\top L^{-1} D^\top \|^2 = x^\top L^{-1} D^\top D L^{-1} x \leq \frac{1}{\kappa^2} \| x \|^2_{V^{-1}_n} \,. 
\end{align*}
Then, letting the right-hand side of (\ref{eq:xHZ_bound}) be $2\delta$ and solving for $\nu$, we obtain that with probability at least $1 - 2\delta$,
\begin{equation}\label{eq:pred_err_1}
    |x^\top L^{-1}Z| \leq \frac{2\sqrt{\log(1/\delta)}}{\kappa} \|x\|_{V^{-1}_n}. 
\end{equation}
Then, the rest of the proof for the theorem follows the proof of Theorem~1 in \cite{li2017provably}. For the sake of completeness, we present the full proof.
\begin{align}\label{eq:secondterm_bound}
    | x^\top L^{-1} E(L+E)^{-1}Z_n| &\leq \|x\|_{L^{-1}} \| L^{-1/2} E(L+E)^{-1}Z_n \| \notag\\
    &\leq \|x\|_{L^{-1}} \| L^{-1/2} E(L+E)^{-1}L^{1/2} \| \| Z_n\|_{L^{-1}} \notag\\
    &\leq \frac{1}{\kappa}  \|x\|_{V^{-1}_n} \| L^{-1/2} E(L+E)^{-1}L^{1/2} \| \| Z_n\|_{V^{-1}_n}
\end{align}
where the last inequality is from $L \succeq \kappa V_n$. Then it follows that
\begin{align*}
     \| L^{-1/2} E(L+E)^{-1} L^{1/2} \| &=  \| L^{-1/2} E (L^{-1} - L^{-1}E(L+E)^{-1} ) L^{1/2} \|\\
     &= \| L^{-1/2} E L^{-1/2} - L^{-1/2} EL^{-1}E(L+E)^{-1}  L^{1/2} \|\\
     &\leq \| L^{-1/2} E L^{-1/2}\| + \| L^{-1/2} E L^{-1/2}\| \| L^{-1/2}E(L+E)^{-1}  L^{1/2} \|
\end{align*}
By solving this inequality, we get
\begin{align*}
    \| L^{-1/2} E (L+E)^{-1} L^{1/2} \| &\leq \frac{\| L^{-1/2} E L^{-1/2}\|}{1-\| L^{-1/2} E L^{-1/2}\|}\\
    &\leq 2 \| L^{-1/2} E L^{-1/2}\|\\
    &\leq  \frac{24}{\kappa^2} \sqrt{\frac{d + \log(1/\delta)}{\lambda_{\min}(V_n)}}
\end{align*}
where the second inequality is from (\ref{eq:HEH_bound}) and the third inequality is from combining with (\ref{eq:HEH_bound}). Combining with (\ref{eq:secondterm_bound}) and $\| Z_n \|_{V^{-1}_n} \leq 2\sqrt{2d + \log \frac{1}{\delta}}$ from Lemma~\ref{lemma:G_bound} (which we assume to hold in this section), we have
\begin{align}
    | x^\top L^{-1}E(L+E)^{-1}Z_n | &\leq \frac{1}{\kappa}  \|x\|_{V^{-1}_n} \| L^{-1/2} E(L+E)^{-1}L^{1/2} \| \| Z_n\|_{V^{-1}_n} \notag\\
    &\leq  \frac{ 48 \left( 2d + \log\frac{1}{\delta} \right)}{\kappa^3\sqrt{\lambda_{\min}(V_n)}} \| x\|_{V^{-1}_n} \label{eq:pred_err_2}
\end{align}
Then combining the results from \eqref{eq:pred_err_1} and \eqref{eq:pred_err_2}, we have
\begin{align*}
    | x^\top (\Hat{\theta}_n - \theta^*) | &\leq  |x^\top L^{-1}Z|  + | x^\top L^{-1}E(L+E)^{-1}Z_n |\\
    &\leq \frac{\sqrt{\log\frac{1}{\delta}}}{\kappa} \|x\|_{V^{-1}_n} + \frac{ 48 \left( 2d + \log\frac{1}{\delta} \right)}{\kappa^3\sqrt{\lambda_{\min}(V_n)}} \| x\|_{V^{-1}_n}.
\end{align*}
Then it follows that $| x^\top (\Hat{\theta}_n - \theta^*) | \leq \frac{5}{\kappa} \sqrt{\log\frac{1}{\delta} } \|x\|_{V^{-1}_n}$ holds as long as $\lambda_{\min}(V_n) \geq \frac{144}{\kappa^4}\left( 4d^2 + \log\frac{1}{\delta} \right)$ holds.

\hfill
\endproof

\begin{lemma}\label{lemma:G_bound}
For any $\delta > 0$, with probability at least $1-\delta$, we have
\begin{equation}\label{eq:G_bound}
     \| J_n(\hat{\theta}) \|_{V^{-1}_n} \leq 4\sqrt{2d + \log \frac{1}{\delta}}  .
\end{equation}

\end{lemma}
\proof{}
This lemma is an extension of Lemma 7 in \cite{li2017provably}.
For convenience, let $Z = J_n(\Hat{\theta})$ and $V = V_n$.
Let $\hat{\mathbb{B}}$ be a 1/2-net of the unit ball $\mathbb{B}^d$. Then $|\hat{\mathbb{B}}| \leq 6^d$ (\citealt{pollard1990empirical}, Lemma 4.1), and for any $x \in \mathbb{B}^d$, there is a $\hat{x} \in \hat{\mathbb{B}}$ such that $\| x - \hat{x}\| \leq \frac{1}{2}$. Therefore, we have
\begin{align*}
    x^\top V^{-1/2} Z  &=  \hat{x}^\top V^{-1/2} Z  +  (x - \hat{x})^\top V^{-1/2} Z \\
    &=  \hat{x}^\top V^{-1/2} Z  +  \| x - \hat{x}\|  \cdot \frac{1}{\| x - \hat{x}\|} (x - \hat{x})^\top V^{-1/2} Z \\
    &\leq \hat{x}^\top V^{-1/2} Z  + \frac{1}{2}\sup_{z \in \mathbb{B}^d}  z^\top V^{-1/2} Z \,.
\end{align*}

Taking supremum on both sides, we get
\begin{equation*}
    \sup_{x \in \mathbb{B}^d}  x^\top V^{-1/2} Z  \leq 2 \max_{\hat{x} \in \hat{\mathbb{B}}}  \hat{x}^\top V^{-1/2} Z \,.
\end{equation*}
Also, note that $\|Z\|_{V^{-1}} = \| V^{-1/2} Z \|_2 = \sup_{\|x\|_2 \leq 1} x^\top V^{-1/2} Z$. Recall that
$Z = \sum_{t=1}^n X^\top_t \mathcal{E}_t$.
Then, it follows that
\begin{align*}
    \mathbb{P}\left( \| Z \|_{V^{-1}} \geq \nu \right) &\leq \mathbb{P}\left( \max_{\hat{x} \in \hat{\mathbb{B}}}  \hat{x}^\top V^{-1/2} Z  > \frac{\nu}{2} \right)\\
    &\leq \sum_{\hat{x} \in \hat{\mathbb{B}}} \mathbb{P}\left(   \hat{x}^\top V^{-1/2} Z  > \frac{\nu}{2} \right)\\
    &= \sum_{\hat{x} \in \hat{\mathbb{B}}} \mathbb{P}\left( \sum_{t=1}^n \hat{x}^\top V^{-1/2} X^\top_t \mathcal{E}_t \geq \frac{\nu}{2} \right) \,.
\end{align*}
Noting that $|\hat{x}^\top V^{-1/2} X^\top_t \mathcal{E}_t| \leq \sqrt{2}\|\hat{x}^\top V^{-1/2} X^\top_t\| $, we again apply Hoeffding inequality (Lemma~\ref{lemma:hoeffding}) to a sum of bounded random variables $\hat{x}^\top V^{-1/2} X^\top_t \mathcal{E}_t $ as done in \eqref{eq:xHZ_bound}. Then, it follows that
\begin{align*}
    \mathbb{P}\left( \| Z \|_{V^{-1}} \geq \nu \right) 
    &\leq  \sum_{\hat{x} \in \hat{\mathbb{B}}} \exp \left\{ -\frac{2\nu^2}{32 \sum_{t=1}^n \| \hat{x}^\top V^{-1/2} X^\top_t \|^2} \right\} \notag\\
    &=   \sum_{\hat{x} \in \hat{\mathbb{B}}}  \exp \left\{ -\frac{\nu^2}{16 \| \hat{x}^\top V^{-1/2} D^\top \|^2} \right\} \notag\\
    &\leq   \exp \left\{ -\frac{ \nu^2}{16} + d \log 6 \right\} 
\end{align*}

where the second inequality is by a union bound and the forth inequality is from Hoeffding inequality. The last inequality comes from the fact that $ V =  D^\top D$ and also from $|\hat{\mathbb{B}}| \leq 6^d$. If we let $\nu = 4\sqrt{2d + \log(1/\delta)}$, then we have
\begin{align*}
     \mathbb{P}\left( \| Z \|_{V^{-1}} \geq 4\sqrt{2d + \log(1/\delta)} \right) \leq \exp \left\{ -\frac{ 32d + 16\log(1/\delta)}{16} + d \log 6 \right\} \leq \delta.
\end{align*}
\endproof

\section{Generating Independent Samples using \textsc{supCB-MNL}}


\begin{algorithm}
\caption{\textsc{baseCB-MNL}}
\begin{algorithmic}[1]
    \STATE \textbf{Input}: confidence radius $\alpha$, index set $\Psi$,  set $A$, features $\{x_{ti}\}$
    \STATE Compute MLE $\Hat{\theta}_t$ by solving the equation
        \begin{equation*}
            \sum_{t' \in \Psi} \sum_{i \in S_{t'}} \left(  p_{t'}(i|S_{t'}, \theta) - y_{t' i} \right) x_{t' i} = 0
        \end{equation*}

        \STATE Update $V_{\Psi} = \sum_{t' \in \Psi} \sum_{i \in S_{t'}} x_{t' i} x_{t' i}^\top$
        \STATE Compute the following:
        \begin{align*}
            w_{ti} &= \alpha \|x_{ti}\|_{V^{-1}_{\Psi}}  \text{ for all $i \in \mathcal{I}$}\\
            \mathcal{W}_t &= 2  \max_{i\in \mathcal{I}} w_{ti}
        \end{align*}
        where $\mathcal{I} = \{i \in S : S \in A\}$
\end{algorithmic}
\label{algo:baseCB-MNL}
\end{algorithm}

To overcome the issue of dependent samples, we design a method which consists of two parts:
(i) a subroutine algorithm \textsc{baseCB-MNL} (Algorithm~\ref{algo:baseCB-MNL}) to compute MLE and maximum standard deviation of utility among the items in the candidate set (assuming statistical independence among the samples), and (ii) a master algorithm \textsc{supCB-MNL} (Algorithm~\ref{algo:supCB-MNL}) to ensure the independence assumption holds. As mentioned in Section~\ref{sec:optimality_v_practicality}, this technique is inspired by the decomposition of the algorithm introduced in \citet{auer2002using} and also adopted in many followup works, e.g., \cite{chu2011contextual, li2017provably, zhou2019learning}.
supCB-MNL operates on the radius of the confidence bound, independent of expected mean utility, to perform exploration. 
supCB-MNL maintains $\{\Psi_\ell\}_{\ell=0}^L$, the sets of time indices which are the partitions of the entire planning horizon $\{1, 2, ..., T\}$. The purpose of this partitioning is to ensure that the choice responses $y_t$ in each index set $\Psi_\ell$ are independent, so that we can apply the normality result of Theorem~\ref{thm:normalityMLE} to samples in each $\Psi_\ell$ seperately.

\begin{algorithm}
\caption{\textsc{supCB-MNL}}
\begin{algorithmic}[1]
    \STATE \textbf{Input}: $T$, initialization $T_0$, confidence radius $\alpha$    
    \STATE \textbf{Initialization}: \textbf{for} $t \in [T_0]$
        \STATE \quad randomly choose $S_t$  with $|S_t|=K$
    \STATE set $L = \lfloor \frac{1}{2}\log_2 T  \rfloor$, and $\Psi_0 = \cdots = \Psi_L = \emptyset$.
    \FOR{all $T_0 = \tau + 1$ to $T$}
        \STATE Initialize $A_1 = \mathcal{S}$ and $\ell=1$
        \WHILE{$S_t$ is empty}
            \STATE (a). Run Algorithm~\ref{algo:baseCB-MNL} with $A_\ell$, $\alpha$ and $\Psi_\ell \cup [T_0]$ to compute $\Hat{\theta}^{(\ell)}_t$, $w^{(\ell)}_{ti}$,  $\mathcal{W}^{(\ell)}_t$
            \STATE (b). \textbf{If } $\mathcal{W}^{(\ell)}_t \leq \frac{1}{\sqrt{T}}$,
            \STATE \qquad set $S_t = \argmax_{S \in A_\ell} R_t(S, \Hat{\theta}^{(\ell)}_t)$
            \STATE \qquad update $\Psi_0 = \Psi_0 \cup \{t\}$
            \STATE (c). \textbf{Else if } $\mathcal{W}^{(\ell)}_t >  2^{-\ell}$,
            \STATE \qquad set $S_t = \argmax_{S \subseteq A_{\ell}}\sum_{i\in S} w_{ti}^{(\ell)}$
            \STATE \qquad update $\Psi_\ell = \Psi_\ell \cup \{t\}$
            \STATE (d). \textbf{Else if } $\mathcal{W}^{(\ell)}_t \leq  2^{-\ell}$,
            \STATE \qquad compute $ \mathcal{M}_{t}^{(\ell)} = \max_{S\in A_{\ell}} R_t(S, \Hat{\theta}^{(\ell)}_t)$
            \STATE \qquad $ \displaystyle A_{\ell+1} = \left\{ S \in A_\ell : R_t(S,\Hat{\theta}^{(\ell)}_t) \geq \mathcal{M}_{t}^{(\ell)} - 2^{-\ell+1}  \right\}$
            \STATE \qquad $\ell \leftarrow \ell + 1$
        \ENDWHILE
    \ENDFOR
\end{algorithmic}
\label{algo:supCB-MNL}
\end{algorithm}


In each round of Algorithm~\ref{algo:supCB-MNL}, the learning agent screens the candidate assortments based on the value of $w_{ti} = \alpha \|x_{ti}\|_{V^{-1}_t}$ for items in assortments in $A_\ell$ through epochs $\ell = 1, ..., L$ until an assortment $S_t$ is chosen.

\begin{itemize}
    \item \textit{Sub-routine:} in step (a), we run \textsc{baseCB-MNL} (Algorithm~\ref{algo:baseCB-MNL}) which uses the normality result to compute $w^{(\ell)}_{ti}$ for all $i$, $\mathcal{W}^{(\ell)}_t$, and $\Hat{\theta}^{(\ell)}_t$. We can utilize the normality result here since $\{ y_t, t \in \Psi_\ell \}$'s are independent given the feature vectors in each $\Psi_\ell$ (see Lemma~\ref{lemma:indepedent_samples}).
    \item \textit{Exploitation:} in step (b), if the maximal confidence interval of an assortment is very small, smaller than $\frac{1}{K\sqrt{T}}$, for all possible candidate sets, then we perform pure exploitation. This step's contribution to the total regret will be small.
    \item \textit{Exploration:} in step (c), if there is a set that has large confidence interval (larger than $2^{-\ell}$), then we choose that set as $S_t$. Then we update the index set $\Psi_\ell$ to include the timestamp $t$.
    \item \textit{Pruning:} finally, step (d) is a pruning step, where we remove clearly sub-optimal sets and keep the sets which are possibly optimal.
\end{itemize}


If the algorithm does not choose $S_t$ in epoch $\ell$, then it moves on to the next epoch $\ell+1$ and repeat the process until $S_t$ is chosen either through exploitation action in (b) or exploration action in (c). Note that when maximizing the expected revenue $R_t(S, \Hat{\theta})$ in step (b) or in step (d), it uses the expected revenue defined in (\ref{eq:expected_revenue}) replacing $\theta^*$ with the current estimator $\Hat{\theta}^{(\ell)}_t$ --- note that we use the expected revenue $R_t(S)$ in \textsc{supCB-MNL}, not the optimistic expected revenue $\tilde{R}_t(S)$ used in \textsc{UCB-MNL} (Algorithm~\ref{algo:UCB-MNL}).

Adapted from Lemma 14 of \citet{auer2002using} and Lemma 4 of \citet{li2017provably}, the following result shows that the samples collected from Algorithm~\ref{algo:supCB-MNL} in each index set $\Psi_\ell$ are independent.

\begin{lemma}\label{lemma:indepedent_samples}
For all $\ell \in [L]$ and $t \in [T]$, given the set of feature vectors in index set $\Psi_\ell$, $\{ [x_{ti}]_{i\in S_t}, t \in \Psi_\ell \}$, the corresponding choice responses $\{ y_t, t \in \Psi_\ell \}$ are independent random variables.
\end{lemma}

\subsection{Regret Bound for \textsc{supCB-MNL}}

Independent samples ensured by the master algorithm \textsc{supCB-MNL} and Lemma~\ref{lemma:indepedent_samples} enable us to apply the non-asymptotic normality result in Theorem~\ref{thm:normalityMLE} separately to samples in each index set $\Psi_\ell$. We present the following regret bound of \textsc{supCB-MNL} (Algorithm~\ref{algo:supCB-MNL}), which is a formal statement of Theorem~\ref{thm:expected_regret_supCb-mnl}

\textbf{Theorem~\ref{thm:expected_regret_supCb-mnl}} (Formal statement).
\textit{
Suppose Assumptions~\ref{assum:x_bound} and \ref{assum:prob_bound}, and we run Algorithm~\ref{algo:supCB-MNL} with $T_0 = \frac{C_0}{\kappa^4}\max\left\{ \frac{\sqrt{dT}}{\sigma_0}, \frac{d + 2\log (TN \log_2 T) }{\sigma_0^2}  \right\}$ with a universal constant $C_0$ and $\alpha = \frac{5}{\kappa} \sqrt{ 2\log(TN \log_2 T)}$ for $T \geq \tilde{T}$ rounds, where
\begin{equation}\label{eq:T0}
        \tilde{T} =  \Omega \left( \max\left\{ \frac{\log^2\left( T N \log_2 T \right)}{d}, d^3 \right\} \right).
\end{equation}
Then, the algorithm's expected regret is upper-bounded by
\begin{align*}
    \mathcal{R}_T 
    &= \mathcal{O}\left(  \sqrt{dT \log(T/d) \log (TN \log_2 T)  \log_2 T} \right).
\end{align*}
}


\textbf{Discussion of Theorem~\ref{thm:expected_regret_supCb-mnl}.}
We establish $\Tilde{\mathcal{O}}(\sqrt{dT})$ regret bound for \textsc{supCB-MNL} algorithm. 
\citet{chen2017note} provide a lower bound of $\Omega(\sqrt{NT})$ in the non-contextual setting which is free of $K$. This lower bound can be translated to $\Omega(\sqrt{dT})$ if each item is represented as one-hot encoding. Hence the regret bound in Theorem~\ref{thm:expected_regret_supCb-mnl} matches the lower bound for the MNL bandit problem with finite actions.
To the best of our knowledge, \textsc{supCB-MNL} is the first algorithm which achieves the rate of $\Tilde{\mathcal{O}}(\sqrt{dT})$ regret in MNL contextual bandits. Comparing with Theorem~\ref{thm:expected_regret_ucb-mnl} for \textsc{UCB-MNL} (Algorithm~\ref{algo:UCB-MNL}) as well as its online update variant (Algorithm~\ref{algo:UCB-MNL-online-update}) --- which are near-optimal in the case of infinitely large item set (or exponentially large $N$) --- the improvement of $\sqrt{d}$ factor comes from directly controlling the utility estimation error using Theorem~\ref{thm:normalityMLE}. Note that the regret bound in Theorem~\ref{thm:expected_regret_supCb-mnl} has logarithmic dependence on $N$, therefore \textsc{supCB-MNL} is not applicable to a case where there are an infinite number of total items. However, when $N$ is not exponentially large (i.e., $N \ll e^{d}$), the rate of \textsc{supCB-MNL} is optimal.

\subsection{Proof Sketch of Theorem~\ref{thm:expected_regret_supCb-mnl}}
Note that we want to have the concentration result of the prediction error in Theorem~\ref{thm:normalityMLE} to hold for all items $i \in [N]$ and for all rounds $t \in [T]$ including the inner loop (epochs) in Algorithm~\ref{algo:supCB-MNL}; hence for all $\ell$ up to $L = \mathcal{O}(\log_2 T)$. Hence, we choose the confidence radius to be $\alpha = \frac{5}{\kappa}\sqrt{2\log\left( T N \log_2 T \right)}$. Then with probability at least $1 - \frac{3}{T N \log_2 T }$, we would have
\begin{equation*}
    | x_{ti}^\top (\Hat{\theta}_t - \theta^*) | \leq  \alpha \|x_{ti}\|_{V^{-1}_{\Psi_\ell}}
\end{equation*}
for each $t \in \Psi_\ell$ if the independence and minimum eigenvalue conditions are satisfied. Then we can use the union bound to show this concentration holds jointly for all items and all rounds with a high probability.
Now, we know that the independence requirement is satisfied by \textsc{supCB-MNL} and Lemma~\ref{lemma:indepedent_samples}. For the minimum eigenvalue condition, we need to ensure that
\begin{equation}\label{eq:minimum_eigenvalue_condition}
    \lambda_{\min}(V_t) = \Omega \left(\frac{ d^2 + \log\left( T N \log_2 T \right) }{\kappa^4}\right).
\end{equation}
Hence for $T \geq \tilde{T}$ where $\tilde{T} = \Omega \left( \max\left\{ \frac{\log^2\left( T N \log_2 T \right)}{d}, d^3 \right\} \right)$, using Proposition~\ref{prop:lowerbounding_lambda}, we can run the random initialization for $\mathcal{O}(\sqrt{dT})$ to ensure \eqref{eq:minimum_eigenvalue_condition} holds with a high probability.
Given this concentration result, we decompose the regret into to two parts -- the regret incurred when an assortment is chosen for exploitation (step (b) in Algorithm~\ref{algo:supCB-MNL}) and the regret for exploration (step (c) in Algorithm~\ref{algo:supCB-MNL}). We show the regret coming from step (b) is small since the utility estimates are already accurate in that case. We also show that even when we take an exploratory action in step (c), the regret incurred by such an action is not too large due to the concentration result as well as the pruning procedure in step (d).

\section{Proof of Theorem~\ref{thm:expected_regret_supCb-mnl}}\label{sec:supCb-mnl_regret_proof}
We first present two lemmas to help bound the cumulative expected regret. The first lemma ensures that normality results (Theorem~\ref{thm:normalityMLE}) holds with given confidence radius $\alpha$ for all items.

\begin{lemma}\label{lemma:normality_bound_all_items}
Let $T_0 = \frac{C_0}{\kappa^4}\max\left\{ \frac{\sqrt{dT}}{\sigma_0}, \frac{d + 2\log (TN \log_2 T) }{\sigma_0^2}  \right\}$ and $\alpha = \frac{5}{\kappa} \sqrt{ 2\log(TN \log_2 T)}$. Suppose $T \geq \tilde{T}$ where $\tilde{T}$ is defined as (\ref{eq:T0}). Define the following event:
\begin{equation}\label{eq:utility_event}
    \mathcal{E}_t := \left\{  | x^{\top}_{ti} \Hat{\theta}^{(\ell)}_{t-1} - x^{\top}_{ti} \theta^* | \leq w^{(\ell)}_{ti}, \enskip \forall i \in [N], \forall \ell \in [L] \right\}
\end{equation}
Then, event $\mathcal{E}_t$ holds with probability at least $1-\mathcal{O}(T^{-2})$ for all $t \geq T_0$ 
\end{lemma}

The next lemma bounds the immediate regret of \textsc{supCB-MNL}, breaking down to two choice scenarios --- when an assortment is chosen for exploitation (step (b)) or for exploration (step (c)) in Algorithm~\ref{algo:baseCB-MNL}. Intuitively, the regret coming from step (b) is very small since the utility estimates are accurate in that scenario. The challenge is to show that even when we take an exploratory action in step (c), the regret incurred by such an action is not too large.

\begin{lemma}\label{lemma:regret_bound_for_psi}
Suppose that event $\mathcal{E}_t$ in (\ref{eq:utility_event}) holds, and that in round $t$, the assortment $S_t$ is chosen at stage $\ell_t$. Then $S_t^* \in A_\ell$ for all $\ell \leq \ell_t$. Furthermore, we have
\begin{align*}
    R_t(S_t^*, \theta^*) - R_t (S_t, \theta^*)
    \leq 
    \begin{dcases}
    \frac{2}{\sqrt{T}},& \text{if } S_t \text{ chosen in step } (b) \\
    \frac{8}{2^{\ell_t}},& \text{if } S_t \text{ chosen in step } (c) \\
\end{dcases}
\end{align*}
\end{lemma}




Then, we follow the similar arguments of \citet{li2017provably} to show the cumulative expected regret bound.
First, define $V_{\ell, t} = \sum_{t \in \Psi_\ell} \sum_{i \in S_t} x_{ti} x_{ti}^\top$, then by Lemma~\ref{lemma:self_norm_MG_bound} and Cauchy-Schwarz inequality, we have
\begin{align*}
    \sum_{t \in \Psi_\ell}  \max_{i \in S_t} w_{ti}^{(\ell)} &= \sum_{t \in \Psi_\ell}  \max_{i \in S_t} \alpha \|x_{ti}\|_{V^{-1}_{\ell,t}}\\
    &\leq \alpha \sqrt{2|\Psi_\ell| d \log(T/d) } .
\end{align*}
However, from the choices made at exploration steps (step (c)) of Algorithm~\ref{algo:supCB-MNL}, we know 
\begin{align*}
    2^{-\ell} |\Psi_\ell| \leq 2 \sum_{t \in \Psi_\ell} \max_{i \in S_t} w_{ti}^{(\ell)}
\end{align*}
for $\ell \in \{1, ..., L\}$. Now, we combine the two inequalities above. Then it follows that
\begin{align}\label{eq:psi_bound}
    |\Psi_\ell| \leq 2^{\ell+1}\alpha \sqrt{2 |\Psi_\ell| d \log(T/d) }.
\end{align}
Note that each index set $\Psi_\ell$ is a disjoint set with $\cup^L_{\ell=0} \Psi_\ell = \{t+1, ..., T\}$.
Then,  we break the regret into three components -- when  event $\mathcal{E}_t$ in \eqref{eq:utility_event} holds, i.e., the concentration result holds, and when the event does not hold ($\mathcal{E}^c_t$), and the random initialization phase with length $T_0$.
Note that we need the minimum eigenvalue of $V_{T_0}$ to be larger than the case in \textsc{UCB-MNL} but we can still use Proposition 1 to ensure such case with a high probability.
\begin{align*}
    \mathcal{R}_T 
    &\leq T_0 + \mathbb{E} \left[ \sum_{t=T_0 + 1}^T   \left( R(S^*, \theta^*) - R(S_t, \theta^*) \right) \mathbb{1}\left(\mathcal{E}_t \right) \right]
    + \mathbb{E} \left[ \sum_{t=T_0+1}^T   \left( R(S^*, \theta^*) - R(S_t, \theta^*) \right) \mathbb{1}\left(\mathcal{E}^c_t\right) \right]
\end{align*}    

We further decompose the regret into the disjoint stages recorded by $\Psi_\ell$.
\begin{align*}
    \mathcal{R}_T &\leq T_0 + \mathbb{E} \left[ \sum_{t \in \Psi_0}   \left( R(S^*, \theta^*) - R(S_t, \theta^*) \right) \mathbb{1}\left( \mathcal{E}_t\right) \right]
    + \mathbb{E} \left[ \sum^L_{\ell=1} \sum_{t \in \Psi_\ell}   \left( R(S^*, \theta^*) - R(S_t, \theta^*) \right) \mathbb{1}\left(\mathcal{E}_t \right) \right] + \mathcal{O}(1) \\
    &\leq T_0 + \frac{2}{\sqrt{T}}|\Psi_0| + \sum^L_{\ell=1} \frac{8}{2^\ell} |\Psi_\ell| 
    + \mathcal{O}(1)\\
    &\leq T_0 + 2\sqrt{T} + \sum^L_{\ell=1} 16\alpha \sqrt{2 |\Psi_\ell| d \log(T/d) }
    + \mathcal{O}(1)\\
    &\leq T_0 + 2\sqrt{T} + 16\alpha \sqrt{ 2d L T \log(T/d) } 
    + \mathcal{O}(1)
\end{align*}
where the third inequality uses (\ref{eq:psi_bound}) and the last inequality is by Cauchy-Schwartz inequality. 
Now, with our choices of $\alpha = \frac{5}{\kappa} \sqrt{ 2\log(TN \log_2 T)}$, $T_0 =  \frac{C_0}{\kappa^4}\max\left\{ \frac{\sqrt{dT}}{\sigma_0}, \frac{d + 2\log (TN \log_2 T) }{\sigma_0^2}  \right\}$ and $L = \lfloor \frac{1}{2}\log_2 T \rfloor \leq \frac{1}{2}\log_2 T$, then we complete the proof

\vspace{0.5cm}

\section{Proofs of Lemmas for Theorem~\ref{thm:expected_regret_supCb-mnl}}

\subsection{Proof of Lemma~\ref{lemma:indepedent_samples}}

\proof{}
Since a timestamp $t$ can only be added to $\Psi_\ell, \ell \geq 1$ in step (c) of Algorithm~\ref{algo:supCB-MNL}, the event $\{t \in \Psi_\ell\}$ only depends on the results of trials $t' \in \cup_{\ell' < \ell} \Psi_{\ell'}$ and on $\Bar{w}^{(\ell)}_{ti}$. From the definition of $\Bar{w}^{(\ell)}_{ti}$, we know it only depends on the sets of feature vectors $\{x_{u,i}\}_{i \in S_u}, u \in \Psi_\ell$ and on $\{x_{ti}\}_{i \in S_t}$.
\endproof

\subsection{Proof of Lemma~\ref{lemma:normality_bound_all_items}}

\proof{}
With $T_0 = \frac{C_0}{\kappa^4}\max\left\{ \frac{\sqrt{dT}}{\sigma_0}, \frac{d + 2\log (TN \log_2 T) }{\sigma_0^2}  \right\}$ and $T \geq \tilde{T}$, at the end of initialization we have 
\begin{align*}
    \lambda_{\min}(V_{T_0}) \geq C \sqrt{dT} =   \Omega \left( \max\left\{ \log\left( T N \log_2 T \right), d^2 \right\} \right),
\end{align*}
with probability at least $1 - \frac{1}{(TN \log_2 T)^2}$ using Proposition~\ref{prop:lowerbounding_lambda}. Then, the condition on the minimum eigenvalue of $V_t$ for $t \geq T_0$ is satisfied since $\lambda_{\min}(V_{t}) \geq \lambda_{\min}(V_{T_0})$.
Therefore, applying Theorem~\ref{thm:normalityMLE}, we have
\begin{equation*}
    | x_{ti}^\top (\Hat{\theta}_t - \theta^*) | \leq  \alpha \|x_{ti}\|_{V^{-1}_t}
\end{equation*}
 with probability at least $1 - \frac{3}{(T N \log_2 T )^2}$. Note that we are applying Theorem~\ref{thm:normalityMLE} to $x_{ti}$ for all $i$ which are i.i.d by definition (Assumption~\ref{assum:x_bound}). Now, applying the union bound over all items and epochs, we complete the proof.
\endproof

\subsection{Proof of Lemma~\ref{lemma:regret_bound_for_psi}}

\proof{}
Combining Lemma~\ref{lemma:revenue_bound} and Lemma~\ref{lemma:optimistic_regret}, we have
\begin{align*}
    \left| R_t(S, \theta^*) - R_t(S, \Hat{\theta}^{(\ell)})  \right| \leq \left| \tilde{R}_t(S, \Hat{\theta}^{(\ell)}) - R_t(S, \Hat{\theta}^{(\ell)})  \right| \leq 2 \max_{i \in S} w^{(\ell)}_{ti} \leq \mathcal{W}^{(\ell)}_{t}.
\end{align*}
We first show the optimal assortment $S_t^* \in A_\ell$ for all $\ell$. We prove this by induction. For $\ell  = 1$, the lemma automatically holds. As an inductive step, suppose $S^*_t \in A_\ell$ and we want to prove $S^*_t \in A_{\ell+1}$. Since the algorithm proceed to stage $\ell+1$, we know from step (c) in Algorithm~\ref{algo:supCB-MNL} that
\begin{align*}
    \left| R_t(S, \theta^*) - R_t(S, \Hat{\theta}^{(\ell)})  \right|  \leq \mathcal{W}^{(\ell)}_{t} \leq 2^{-\ell}
\end{align*}
for all $S \in A_\ell$. In particular, it holds for $S = S^*_t$ since $S^*_t \in A_\ell$ by the inductive step. Then the optimality of $S_t^*$ implies
\begin{align*}
    R_t(S^*_t, \Hat{\theta}^{(\ell)}) &\geq R_t(S^*_t, \theta^*) -  2^{-\ell} \geq R_t(S, \theta^*) -  2^{-\ell} \geq R_t(S, \Hat{\theta}^{(\ell)}) - 2 \cdot 2^{-\ell}
\end{align*}
for $S \in A_\ell$. Hence, it follows that
\begin{align*}
    R_t(S^*_t, \Hat{\theta}^{(\ell)}) \geq \max_{S \in A_\ell} R_t(S, \Hat{\theta}^{(\ell)}) - 2 \cdot 2^{-\ell} = \mathcal{M}_t^{(\ell)} - 2 \cdot 2^{-\ell}.
\end{align*}
Therefore, we have $S^*_t \in A_{\ell+1}$ according to step (d).

If $S_t$ is selected in step (b), that it implies $R_t(S_t, \Hat{\theta}^{(\ell_t)}) \geq R_t(S^*_t, \Hat{\theta}^{(\ell_t)})$. Then if follows that
\begin{align*}
    R_t(S_t, \theta^*) &\geq R_t(S_t, \Hat{\theta}^{(\ell_t)}) - \frac{1}{\sqrt{T}} \geq R_t(S^*_t, \Hat{\theta}^{(\ell_t)}) - \frac{1}{\sqrt{T}} \geq R_t(S^*_t, \theta^*) - \frac{2}{\sqrt{T}}.
\end{align*}

Suppose $S_t$ is chose at stage $\ell_t$ in step (c) in Algorithm~\ref{algo:supCB-MNL}. The lemma holds automatically for $\ell_t = 1$ since $R_t(S,\theta^*) \in [0,1]$ for all $S$ and $t$. If $\ell_t > 1$, $S_t$ must have passed through steps (c) and (d) in the previous stage, $\ell_t - 1$. Also note that we have already shown that the optimal assortment $S^*_t \in A_{\ell_t}$. Hence, $S^*_t$ also must have passed through steps (c) and (d) in stage $\ell_t - 1$. Therefore, passing through step (c) at stage $\ell_t - 1$ implies that
\begin{align*}
    \left| R_t(S, \Hat{\theta}^{(\ell_t-1)}) - R_t(S, \theta^*) \right| \leq \mathcal{W}^{(\ell_t-1)}_{t} \leq   2^{-(\ell_t-1)}
\end{align*}
for $S = S_t$ and $S = S^*_t$. Also, for step (d) at stage $\ell_t - 1$ implies that
\begin{align*}
     R_t(S^*_t, \Hat{\theta}^{(\ell_t-1)}) -  R_t(S_t, \Hat{\theta}^{(\ell_t-1)}) \leq 2 \cdot 2^{-(\ell_t-1)}
\end{align*}
Combining these inequalities above, we have
\begin{align*}
    R_t(S_t, \theta^*) &\geq R_t(S_t, \Hat{\theta}^{(\ell_t-1)}) -  2^{-(\ell_t-1)}\\
    &\geq R_t(S^*_t, \Hat{\theta}^{(\ell_t-1)}) - 3 \cdot 2^{-(\ell_t-1)}\\
    &\geq R_t(S^*_t, \theta^*) - 4 \cdot 2^{-(\ell_t-1)}.
\end{align*}
 \endproof

\section{Proof of Theorem~\ref{thm:expected_regret_DBL-MNL}}

We first make the formal statement of Theorem~\ref{thm:expected_regret_DBL-MNL}.

\textbf{Theorem~\ref{thm:expected_regret_DBL-MNL}} (Formal statement)\textbf{.} \textit{Suppose Assumptions~\ref{assum:x_bound}-\ref{assum:rho} hold, $r_i \equiv r$ is uniform for all $i$, $K \leq \frac{18}{\kappa^4}$, and we run \textsc{DBL-MNL} 
with $\alpha_k = \frac{5}{\kappa}\sqrt{\log(\tau_k^2 N/4)}$ and
$q_k = \frac{288}{ K \sigma_0 \kappa^4}(4d^2 + \log(\tau_k^2 N/4))$.
Then the expected regret of \textsc{DBL-MNL}
over horizon  $T$  is  upper-bounded by
\begin{align*}
\Rcal_T = \Ocal\big( \sqrt{d T \log \left( T/d\right) \log(T N) \log_2 T } \big) \,.    
\end{align*}
}

\begin{remark}
We emphasize that the assumption $K \leq \frac{18}{\kappa^4}$  is not restrictive. In fact, we can instead use Proposition~\ref{prop:lowerbounding_lambda} to show that
\begin{align*}
q_k = \frac{C}{K} \max\left\{ \frac{d^2 + \log(\tau_k^2 N/4) }{\sigma_0 \kappa^4}, \frac{d + 2\log (\tau_k/2) }{\sigma_0^2}  \right\}    
\end{align*}
for some constant $C$ satisfies the threshold on $\lambda_{\min}(V_{\tau_k})$ without assuming $K \leq \frac{18}{\kappa^4}$. However, we would like to provide a specific value of $q_k$ which does not depend on an unknown constant since $q_k$ is an input to the algorithm. Furthermore, in many real-world applications, $K$ is typically small; hence $K \leq \frac{18}{\kappa^4}$ (recall that $\kappa \in (0,1)$) is a  reasonable assumption.
\end{remark}

Since the length of episode grows exponentially, the number of episodes by round $T$  is logarithmic in horizon $T$. In particular, round $T$ belongs to the $L$-th episode with $L = \lfloor \log_2 T \rfloor + 1$. Let $\text{Regret}(k)$ denote cumulative regret of the $k$-th episode. Hence,
\begin{align*}
\mathcal{R}_T \leq \sum_{k=1}^L \text{Regret}(k) \,.
\end{align*}
Let $\Tcal_k := \{\tau_{k-1}+1, ..., \tau_k\}$ denote a set of rounds that belong to the $k$-th episode.
Note that the length of the $k$-th episode is $|\Tcal_k| = \tau_k/2$. 
Now, for each episode $k$, we consider the following two cases.
\begin{enumerate}[(i)]
    \item $|\Tcal_k| \leq q_k$: In this case, the length of an episode is not large enough to ensure the concentration of the prediction error due to the failure to ensure the lower bound on $\lambda_{\min}(V_t)$. Therefore, we use a crude upper bound on the regret in this case. However, the number of such rounds is only logarithmic in $T$, hence contributing minimally to the total regret.
    \item $|\Tcal_k| > q_k$: We can apply Theorem~\ref{thm:normalityMLE} in this case if the lower bound on $\lambda_{\min}(V_t)$ is guaranteed. $\lambda_{\min}(V_t)$ grows linearly as $t$ increases in each episode (with high probability) since samples are independent of each other.
    In case of $\lambda_{\min}(V_t)$ not growing as fast as the rate we require, we perform random sampling to satisfy this criterion towards the end of each episode. Therefore, with high probability, the lower bound on $\lambda_{\min}(V_t)$ becomes satisfied.
\end{enumerate}

For case (i), clearly $q_k \leq 2C_0 (4d^2 + \log(T^2 N))$ for any $k$ where $C_0 = \frac{144}{ K \sigma_0 \kappa^4}$. 
$|\Tcal_k|$ eventually grows to be larger than $2C_0(4d^2 + \log(T^2 N))$. Let $k'$ be the first episode such that $|\Tcal_{k'}| \geq 2C_0(4d^2 + \log(T^2 N))$. Hence, $|\Tcal_{k'}| \leq 4C_0(2d^2 + \log(T^2 N))$. Then cumulative
 regret due to case (i)  is at most 
\begin{align*}
    \sum_{k=1}^{k'-1} \text{Regret}(k) \leq \sum_{k=1}^{k'-1} |\Tcal_k| = |\Tcal_{k'}| \leq 4C_0 \left(4d^2 + \log(T^2 N)\right) \,.
\end{align*}

For case (ii), it suffices to show random sampling ensures the growth of $\lambda_{\min}(V_t)$. 
Lemma~\ref{lemma:dbl_mnl_min_eigen} shows that random sampling with duration $q_k$ specified in Theorem~\ref{thm:expected_regret_DBL-MNL} ensures the lower bound of $\lambda_{\min}(V_t)$, i.e., $\lambda_{\min}(V_t) \geq C_0(4d^2 + \log(\tau_k^2 N/4))$ with high probability.

\begin{lemma}\label{lemma:dbl_mnl_min_eigen}
Suppose $K \leq \frac{18}{\kappa^4}$ and $q_k = 2C_0(4d^2 + \log(\tau_k^2 N/4))$.
For the $k$-th episode, with probability at least $1 - \frac{4}{\tau_k^2 N}$, we have
\begin{align}\label{eq:dbl_mnl_min_eigen}
\lambda_{\min}(V_{\tau_k}) \geq C_0(4d^2 + \log(\tau_k^2 N/4))
\end{align}
where $C_0 = \frac{144}{ K \sigma_0 \kappa^4}$.
\end{lemma}


We then apply Theorem~\ref{thm:normalityMLE} to prediction error in the $k$-th episode which requires samples in the $k-1$-th episode are independent and $\lambda_{\min}(V_{\tau_{k-1}})$ at the end of the $k-1$-th episode is large enough.
With a lower bound guarantee on $\lambda_{\min}(V_{\tau_{k-1}})$ from Lemma~\ref{lemma:dbl_mnl_min_eigen} and the fact that samples are independent of each other within each episode, we have with probability at least $1 - \frac{3}{|\Tcal_k|^2 N}$
\begin{align*}
    | x_{ti}^\top(\hat{\theta}_k - \theta^*)| \leq \alpha_k \|x_{ti}\|_{W^{-1}_{k-1}}
\end{align*}
where $\alpha_k = \frac{5}{\kappa}\sqrt{\log(\tau_k^2 N/4)}$.
Recall that $W_{k-1} = V_{\tau_{k-1}} = \sum_{t'= \tau_{k-1}+1}^{\tau_{k-1}} \sum_{i \in S_{t'}} x_{t' i} x_{t' i}^\top $ is the Gram matrix at the end of the $k-1$-th episode.
Then, we can use the union bound to show this concentration result for all times and all round within the episode. Hence, it folows that with  probability at least $1 - \frac{3}{|\Tcal_k|}$,
\begin{align}\label{eq:dbl_mnl_mle_error_bound}
    | x_{ti}^\top(\hat{\theta}_k - \theta^*)| \leq \alpha_k \|x_{ti}\|_{W^{-1}_{k-1}}, \enskip \forall i \in [N], \forall t \in \Tcal_k \,.
\end{align}
Let $\tilde{\Ecal}_k$ denote the event that both the minimum eigenvalue condition in \eqref{eq:dbl_mnl_min_eigen} (at the end of the $k-1$-th episode) and the MLE concentration result in \eqref{eq:dbl_mnl_mle_error_bound} hold.
\begin{align*}
    \tilde{\Ecal}_{k,1} &:= \left\{ \lambda_{\min}(V_{\tau_{k-1}}) \geq C_0\left(4d^2 + \log\Big(\frac{\tau_{k-1} N}{2}\Big) \right) \right\}\\
    \tilde{\Ecal}_{k,2} &:= \left\{ | x_{ti}^\top(\hat{\theta}_k - \theta^*)| \leq \alpha_k \|x_{ti}\|_{W^{-1}_{k-1}}, \forall i \in [N], \forall t \in \Tcal_k \right\}\\
    \tilde{\Ecal}_k &:= \tilde{\Ecal}_{k,1} \cap \tilde{\Ecal}_{k,2} \,.
\end{align*}

On this event $\tilde{\Ecal}_k$, by the definition of the upper confidence bound of an utility estimate $\tilde{z}_{ti}$ and following the same arguments as Lemma~\ref{lemma:utility_bound}, we have
\begin{align*}
     0 \leq \tilde{z}_{ti} - x_{ti}^\top \theta^* \leq 2\alpha_k \|x_{ti}\|_{W^{-1}_{k-1}} \,.
\end{align*}
Therefore, the optimistic expected revenue $\Tilde{R}_t(S)$ based on $\{\tilde{z}_{ti}\}$ is computed the same way as \eqref{eq:ucb_revenue}.
It is important to note that while the formation of the optimistic revenue $\Tilde{R}_t(S)$ is identical to \eqref{eq:ucb_revenue},
the actual values of $\Tilde{R}_t(S)$ are different for the two algorithms. 
In particular, when feature dimension $d$ is large, $\Tilde{R}_t(S)$ of \textsc{DBL-MNL} can be much tighter than that of \textsc{UCB-MNL} since the confidence width $\tilde{\alpha}_k$ for \textsc{DBL-MNL} does not have dependence on $d$.

Let $S_t = \argmax_{S \in \Scal}\Tilde{R}_t(S)$. Then, it follows that  $\Tilde{R}_t(S_t) \geq R(S_t^*, \theta^*)$ following from Lemma~\ref{lemma:revenue_bound}.
Thus, we can bound the regret in the $k$-th episode as follows:
\begin{align*}
    \text{Regret}(k) 
    &= \sum_{t \in \Tcal_k} \left( R(S_t^*, \theta^*) - R(S_t, \theta^*)\right)\mathbb{1}( \tilde{\Ecal}_k)\\
    &\leq \sum_{t \in \Tcal_k} \left( \tilde{R}(S_t) - R(S_t, \theta^*)\right)\mathbb{1}( \tilde{\Ecal}_k)
\end{align*}
 Then, by the Lipschitz property of the expected revenue of the MNL model shown in Lemma~\ref{lemma:optimistic_regret}, it follows that
\begin{align*}
    \sum_{t \in \Tcal_k} \left( \tilde{R}(S_t) - R(S_t, \theta^*)\right)\mathbb{1}( \tilde{\Ecal}_k)
    &\leq \sum_{t \in \Tcal_k} \sum_{i \in S_t} \left| x_{ti}^\top ( \hat{\theta}_k - \theta^* ) + \alpha_k \|x_{ti}\|_{W^{-1}_{k-1}}\right|\\
    &\leq  2\alpha_k \sum_{t \in \Tcal_k} \sum_{i \in S_t} \|x_{ti}\|_{W^{-1}_{k-1}}
\end{align*}
where the last inequality is from \eqref{eq:dbl_mnl_mle_error_bound}. Then we use Lemma~\ref{lemma:dbl_mnl_gram_mat_conversion} to bound using the norm using the current Gram matrix. This result utilizes the fact that the minimum eigenvalue of the Gram matrix grows linearly within each episode since the samples are independent from each other, allowing us to use the matrix Chernoff inequality to the sum of independent matrices. Furthermore, the fact that episode length difference is two-fold for adjacent episodes allows us to bound the difference between the Gram matrices.

\begin{lemma}\label{lemma:dbl_mnl_gram_mat_conversion}
For $t \in \Tcal_k$, 
\begin{align*}
    \sum_{t \in \Tcal_k} \sum_{i \in S_t} \|x_{ti}\|_{W^{-1}_{k-1}}
    \leq C_1\sum_{t \in \Tcal_k}  \sum_{i \in S_t} \|x_{ti}\|_{V^{-1}_{t-1}}
\end{align*}
with probability at least $1-d e^{-C_2 (t - \tau_{k-1} )}$.
\end{lemma}

Let $\tilde{\Ecal}_{k,3} := \left\{ \sum_{t \in \Tcal_k} \sum_{i \in S_t} \|x_{ti}\|_{W^{-1}_{k-1}}
    \leq C_1\sum_{t \in \Tcal_k}  \sum_{i \in S_t} \|x_{ti}\|_{V^{-1}_{t-1}}, \forall t \in \Tcal_k \right\}$ denote the event that Lemma~\ref{lemma:dbl_mnl_gram_mat_conversion} holds for the $k$-th episode.
Under this event along with , it follows that
\begin{align*}
    \sum_{t \in \Tcal_k} \left( \tilde{R}(S_t) - R(S_t, \theta^*)\right)\mathbb{1}( \tilde{\Ecal}_k \cap \tilde{\Ecal}_{k,3})
    &\leq 2C_1 \alpha_k \sum_{t \in \Tcal_k}  \sum_{i \in S_t} \|x_{ti}\|_{V^{-1}_{t-1}}\\
    &\leq 2C_1 \alpha_k \sqrt{\frac{\tau_{k}}{2} \sum_{t \in \Tcal_k}  \sum_{i \in S_t} \|x_{ti}\|^2_{V^{-1}_{t-1}}}\\
    &\leq  2C_1 \alpha_k \sqrt{\tau_{k}d \log \left( \frac{\tau_k}{2d}\right) }
\end{align*}
where we use the Cauchy-Schwarz inequality in the second inequality and apply the bound on the self-normalized process in Lemma \ref{lemma:self_norm_MG_bound} in the last inequality. Thus, when events $\tilde{\Ecal}_k$ and $\tilde{\Ecal}_{k,3}$ hold, the regret in the $k$-th episode is bounded by
\begin{align*}
    \sum_{t \in \Tcal_k} \left( R(S_t^*, \theta^*) - R(S_t, \theta^*)\right)\mathbb{1}( \tilde{\Ecal}_k \cap \tilde{\Ecal}_{k,3}) = \Ocal\left( \sqrt{d\tau_{k} \log \left( \tau_{k}/d\right) \log(\tau_{k} N) } \right)
\end{align*}
On the other hand, the cumulative regret for the episode under the failure events of $\tilde{\Ecal}_k$ and $\tilde{\Ecal}_{k,3}$ are 
\begin{align*}
    \sum_{t \in \Tcal_k} \left( R(S_t^*, \theta^*) - R(S_t, \theta^*)\right)\mathbb{1}( \tilde{\Ecal}_k^c ) &= \Ocal(1)\\
    \sum_{t \in \Tcal_k} \left( R(S_t^*, \theta^*) - R(S_t, \theta^*)\right)\mathbb{1}( \tilde{\Ecal}_{k,3}^c) &= \Ocal(d) \,.
\end{align*}
Therefore, summing over all episodes, the cumulative expected regret is given by
\begin{align*}
    \Rcal_T = \Ocal\left( \sqrt{dT \log \left( T/d\right) \log(T N) \log_2 T } \right)
\end{align*}

\subsection{Proof of Lemma~\ref{lemma:dbl_mnl_min_eigen}}
\proof{}
By the design of Algorithm~\ref{algo:DBL-MNL}, it suffices to show that the random sampling for duration $q_k$ provides sufficient growth of $\lambda_{\min}(V_{\tau_k})$. 
Let $\tilde{\Tcal}_k$ be the set of rounds in the $k$-th episode that random sampling is performed. 
Without loss generality, assume that the random initialization is invoked for the full duration $q_k$ 
(note that Algorithm~\ref{algo:DBL-MNL} may not invoke random sampling at all if the minimum eigenvalue condition is already satisfied). Hence, $\tilde{\Tcal}_k = \{\tau_k - q_k + 1, \tau_k \}$ in this case.
First, under random sampling of $S_t$, we have
\begin{align*}
    \lambda_{\min}\left(\sum_{t \in \tilde{\Tcal}_k} \sum_{i \in S_t} \EE[x_{ti}x_{ti}^\top] \right)
    &\geq \sum_{t \in \tilde{\Tcal}_k} \sum_{i \in S_t}  \lambda_{\min}\left(\EE[x_{ti}x_{ti}^\top] \right)\\
    &= K q_k \sigma_0 \\
    &= \frac{288}{\kappa^4}(4d^2 + \log(\tau_k^2 N/4))
\end{align*}
where the inequality is from the fact that the minimum eigenvalue function $\lambda_{\min}(\cdot)$ is concave over positive semi-definite matrices.
Also, since $\| x_{ti} \| \leq 1$ is bounded,
\begin{align*}
    \lambda_{\max}\left(\sum_{i \in S_t} \EE[x_{ti}x_{ti}^\top] \right) \leq K
\end{align*}
for all $t$. Therefore, we can use the Matrix Chernoff inequality shown in Lemma~\ref{lemma:matrix_concentration_iid} (Corollary 5.2 of \cite{tropp2012user})

\begin{align*}
    &\PP \left\{ \lambda_{\min}\Big(\sum_{t \in \tilde{\Tcal}_k} \sum_{i \in S_t} x_{ti}x_{ti}^\top \Big) \leq \frac{144}{\kappa^4}(4d^2 + \log(\tau_k^2 N/4)) \right\}\\
    &\leq \PP \left\{ \lambda_{\min}\Big(\sum_{t \in \tilde{\Tcal}_k} \sum_{i \in S_t} x_{ti}x_{ti}^\top \Big) \leq \frac{1}{2} \cdot \lambda_{\min}\Big(\sum_{t \in \tilde{\Tcal}_k} \sum_{i \in S_t} \EE[x_{ti}x_{ti}^\top] \Big) \right\}\\
    &\leq d \cdot \exp\left\{ - \frac{1}{4} \cdot \lambda_{\min}\Big(\sum_{t \in \tilde{\Tcal}_k} \sum_{i \in S_t} \EE[x_{ti}x_{ti}^\top] \Big)/(2K)  \right\}\\
    &\leq d \cdot \exp\left\{ - \frac{18(4d^2 + \log(\tau_k^2 N/4))}{K \kappa^4} \right\}\\
    &= \exp\left\{ \log d - \frac{72 d^2 }{K \kappa^4}  - \frac{18 \log(\tau_k^2 N/4)}{K \kappa^4} \right\}\\
    &\leq \exp\left\{  - \frac{18 \log(\tau_k^2 N/4)}{K \kappa^4} \right\}\\
    &\leq \left(\frac{4}{\tau^2_k N}\right)^{18/(K\kappa^4)}\\
    &\leq \frac{4}{\tau^2_k N} \,.
\end{align*}
Since $  \lambda(V_{\tau_k}) \succcurlyeq \sum_{t \in \tilde{\Tcal}_k} \sum_{i \in S_t} x_{ti}x_{ti}^\top $, this completes the proof.
\endproof

\subsection{Proof of Lemma~\ref{lemma:dbl_mnl_gram_mat_conversion}}

\proof{}
Recall that $W_{k-1}$ is the Gram matrix at the end of the $k-1$-th episode, i.e., $V_{\tau_{t-1}}$ before it resets at the beginning of the $k$-th episode.
Since $V_t$ resets at the beginning of each episode, we focus on how $V_t$ grows in the $k$-th episode relative to $W_{k - 1}$, the Gram matrix at the end of the previous episode.
Clearly, if $C W_{k - 1} \succcurlyeq V_t$, for all $t \in \{\tau_{k - 1}+1,  \tau_k\}$ for some constant $C$, then the claim holds. 
Then it suffices to show $\lambda_{\min}(V_t)$ grows linearly as $t$ increases during the $k-1$-th episode. In fact, since $\Xcal$ is time-invariant, we show the $\lambda_{\min}(V_t)$ grows linearly with $t$ in all episodes.

Let $\tilde{\theta}_{k,t}$ be the parameter corresponding to the upper confidence reward at round $t$, $\max_{S \in \Scal} \tilde{R}_t(S)$. Note that $\tilde{\theta}_{k,t}$ is not the same as the MLE $\hat{\theta}_k$. Since we take an UCB action in Algorithm~\ref{algo:DBL-MNL}, this is equivalent to taking some optimistic parameter within the confidence ellipsoid centered at $\hat{\theta}_k$. It is important to note that since we do not update the MLE and confidence bound within each episode, the samples $y_t$'s are still independent from each other in the same episode.
Consider $(i_1, ..., i_N)$, a set of all permutations of integers $\{1,,,N\}$. Without loss of generality, assume $N$ is divisible by $K$. Then we can write
\begin{align*}
    \EE \left[X_{ti} X_{ti}^\top \right] 
    &= \frac{1}{N}\EE \left[X_{t1} X_{t1}^\top + ... + X_{tN} X_{tN}^\top \right]\\
    &= \frac{1}{N} \sum_{(i_1,...,i_N)} \EE \left[ (X_{t,i_1} X_{t,i_1}^\top + ... + X_{t,i_N} X_{t,i_N}^\top) \mathbb{1}\{ X_{t,i_1}^\top \tilde{\theta}_{k,t} < \cdots < X_{t,i_N}^\top \tilde{\theta}_{k,t} \} \right]\\
    &\preccurlyeq \frac{1}{N} \sum_{(i_1,...,i_N)} \frac{N}{K} C_X \EE \left[ (\Vb_{t,\min}(\Ical) + \Vb_{t,\max}(\Ical)) \mathbb{1}\{ X_{t,i_1}^\top \tilde{\theta}_{k,t} < \cdots < X_{t,i_N}^\top \tilde{\theta}_{k,t} \} \right]
\end{align*}
where $\Vb_{t,\min}(\Ical)$ and $\Vb_{t,\max}(\Ical)$ are the first and last $K$ sums respectively under ordering $\Ical = (i_1,...,i_N)$. That is,
\begin{align*}
    \Vb_{t,\min}(\Ical) &= \Vb_{t,\min}(i_1,...,i_N) := X_{t,i_1} X_{t,i_1}^\top + ... + X_{t,i_K} X_{t,i_K}^\top\\
    \Vb_{t,\max}(\Ical) &= \Vb_{t,\max}(i_1,...,i_N) := X_{t,i_{N-K+1}} X_{t,i_{N-K+1}}^\top + ... + X_{t,i_N} X_{t,i_N}^\top
\end{align*}
Note that the last inequality holds since $C_X (\Vb_{\min}(\Ical) + \Vb_{\max}(\Ical))$ dominates any $K$ sum in $\{X_{t,i_1}X_{t,i_1}^\top, ..., X_{t,i_N}X_{t,i_N}^\top\}$ which follows from applying Lemma~\ref{lemma:boundary_dominance} repeatedly from $k'=1$ to $k'=K$.
\begin{align*}
    \EE \left[X_{ti} X_{ti}^\top \right] 
    &\preccurlyeq \frac{C_X}{K} \sum_{(i_1,...,i_N)} \EE \left[ (\Vb_{t,\min}(\Ical) + \Vb_{t,\max}(\Ical)) \mathbb{1}\{ X_{t,i_1}^\top \tilde{\theta}_{k,t} < \cdots < X_{t,i_N}^\top \tilde{\theta}_{k,t} \} \right]\\
    &\preccurlyeq \frac{C_X \rho_0}{K} \sum_{(i_1,...,i_N)} \EE \left[ \Vb_{t,\max}(\Ical) \mathbb{1}\{ X_{t,i_1}^\top \tilde{\theta}_{k,t} < \cdots < X_{t,i_N}^\top \tilde{\theta}_{k,t} \} \right]\\
    &= \frac{C_X \rho_0}{K} \EE \left[ \sum_{X_{ti} \in \Xcal_t} X_{ti} X_{ti}^\top \mathbb{1}\big(X_{ti} \in S_t \big)\right]
\end{align*}
where the second inequality is comes from utilizing the relaxed symmetry (Assumption~\ref{assum:rho}) and the proof of Lemma~2 in \cite{oh2020sparsity}.
The last eqaulity follow from the fact that $S_t = \argmax_{S \in \Scal} \tilde{R}_t(S)$.
Therefore, 
\begin{align*}
    \EE \left[ \sum_{X_{ti} \in \Xcal_t} X_{ti} X_{ti}^\top \mathbb{1}\big(X_{ti} \in S_t \big)\right]
    \succcurlyeq \frac{K}{C_X \rho_0} \EE \left[X_{ti} X_{ti}^\top \right] \,.
\end{align*}
Now, for $t \in \Tcal_k$, we define 
\begin{align*}
    \Sigma_{k,t} := \sum_{t'= \tau_{k-1}+1}^t \EE \left[  \sum_{X_{t'i} \in \Xcal_{t'}} X_{t'i} X_{t'i}^\top \mathbb{1}\big(X_{t'i} \in S_{t'} \big)\right] \,.
\end{align*}
Then, since the minimum eigenvalue function $\lambda_{\min}(\cdot)$ is concave over positive semi-definite matrices, we have
\begin{align}\label{eq:min_eigen_lowerbound_adapted_cov}
 \lambda_{\min}\left(\Sigma_{k,t} \right) 
 &= \lambda_{\min}\left( \sum_{t'= \tau_{k-1}+1}^t \EE \left[  \sum_{X_{t'i} \in \Xcal_{t'}} X_{t'i} X_{t'i}^\top \mathbb{1}\big(X_{t'i} \in S_{t'} \big)\right] \right) \notag\\
 &\geq \sum_{s=\tau_{k-1}+1}^t \lambda_{\min}\left(\EE \left[  \sum_{X_{t'i} \in \Xcal_{t'}} X_{t'i} X_{t'i}^\top \mathbb{1}\big(X_{t'i} \in S_{t'} \big)\right]\right) \notag\\
 &\geq  \frac{K(t - \tau_{k-1} )\sigma_0}{\rho_0 C_\mathcal{X}} > 0 \,.
\end{align}
Now, to apply the matrix concentration inequality,
we need to show an upper bound on the maximum eigenvalue of $\EE \left[  \sum_{X_{t'i} \in \Xcal_{t'}} X_{t'i} X_{t'i}^\top \mathbb{1}\big(X_{t'i} \in S_{t'} \big)\right]$. We use the fact that $\| X_{t' i}\| \leq 1$ is bounded. Hence, we have for all $\tau$
\begin{align*}
    \lambda_{\max}\left( \EE \left[  \sum_{X_{t'i} \in \Xcal_{t'}} X_{t'i} X_{t'i}^\top \mathbb{1}\big(X_{t'i} \in S_{t'} \big)\right]\right) \leq K \,.
\end{align*}
Then we can apply Corollary~5.2 in \cite{tropp2012user} to the finite sequence of independent matrices $V_t$ for $t \in \Tcal_k$.
\begin{align*}
    \mathbb{P}\left( \lambda_{\min}(V_t) \leq \frac{K(t - \tau_{k-1} )\sigma_0}{2 \rho_0 C_\mathcal{X}} \right) 
    &\leq d \left( \frac{e^{-1/2}}{ 0.5^{1/2}}\right)^{\frac{(t - \tau_{k-1} )\sigma_0}{\rho_0 C_\mathcal{X} }}\\
    &= d \exp \left\{\frac{(t - \tau_{k-1} )\sigma_0}{\rho_0 C_\mathcal{X}}\log \left( \frac{e^{-1/2}}{ 0.5^{1/2}}\right) \right\}\\
    &\leq d \exp\left\{-\frac{(t - \tau_{k-1} )\sigma_0 }{10 \rho_0 C_\mathcal{X}} \right\}
\end{align*}
 where the last inequality uses $-\frac{1}{2} - \frac{1}{2}\log\frac{1}{2} \leq -\frac{1}{10}$. 
Therefore, $\lambda_{\min}(V_t)$ grows linearly as $t$ grows within the episode with probability at least $1-d \exp\left\{-(t - \tau_{k-1} )\sigma_0/(10 \rho_0 C_\mathcal{X}) \right\}$. This completes the proof.
\endproof

\begin{remark}
Since our primary focus here is to show $\lambda_{\min}(V_t)$ grows linearly in every episode, we only show a very crude bound for  $C_\mathcal{X}$ for which we show a finite value
Note that exact value of $C_\mathcal{X}$ is characterized by the distribution of feature vector. For example, multivariate Gaussian and uniform distributions, it can be shown that $C_\mathcal{X} = \Ocal(1)$.
\end{remark}

\begin{lemma}\label{lemma:boundary_dominance}
Consider i.i.d. arbitrary distribution $p_\mathcal{X}$.
Fix some vector $\theta \in \mathbb{R}^d$. For a given integer $k \in \{k',..., N-k'+1\}$,
\begin{align*}
&\mathbb{E}\left[ X_k X_k^\top  \mathbb{1}\{ X_1^\top \theta < \cdots <  X_k^\top \theta <\cdots < X_N^\top \theta \} \right]\\
&\preccurlyeq  C_k \mathbb{E}\left[ ( X_{k'} X_{k'}^\top + X_{N-k'+1} X_{N-k'+1}^\top) \mathbb{1}\{ X_1^\top \theta < \cdots < X_N^\top \theta \} \right]
\end{align*}
where $C_k = \frac{(k'-1)!(N-k')!}{(k-1)!(N-k)!}$.
\end{lemma}

\proof{}
First notice that
\begin{align*}
&\mathbb{E}\left[ X_k X_k^\top  \mathbb{1}\{ X_1^\top \theta < \cdots <  X_k^\top \theta <\cdots < X_N^\top \theta \} \right]\\
&= \mathbb{E}_V\left[ V V^\top  \mathbb{E}_{X_{1:N}/X_k}\left[ \mathbb{1}\{ X_1^\top \theta < \cdots  < X_{k-1}^\top \theta  <  V^\top \theta < X_{k+1}^\top \theta < \cdots < X_N^\top \theta \} \mid V\right] \right]
\end{align*}
where $X_{1:N}/X_k$ denotes $X_1, ..., X_{k-1}, X_{k+1}, ..., X_N$.
Let $\psi(y) := \mathbb{P}(X^\top \theta \leq y)$ denote the CDF of $X^\top \theta$. Then
\begin{align*}
    &\mathbb{P}\left( X_1^\top \theta < \cdots  < X_{k-1}^\top \theta  <  V^\top \theta < X_{k+1}^\top \theta <\cdots < X_N^\top \theta \right)\\
    &= \prod_{i=1}^{k-1} \mathbb{P}\left( X_i^\top \theta \leq V^\top \theta \right) \frac{1}{(k-1)!} \prod_{i=k+1}^{N} \mathbb{P}\left( X_i^\top \theta \geq V^\top \theta \right) \frac{1}{(N-k)!}\\
    &= \frac{1}{(k-1)!(N-k)!} \psi(V^\top \theta)^{k-1} \left(1 - \psi(V^\top \theta) \right)^{N-k} .
\end{align*}
Then, we need to show there exists $C$ such that
\begin{align*}
&\mathbb{P}\left( X_1^\top \theta < \cdots  < X_{k-1}^\top \theta  <  V^\top \theta < X_{k+1}^\top \theta <\cdots < X_N^\top \theta \right)\\
&\leq C \mathbb{P}\left( X_1^\top \theta < ... < X_{k'-1}^\top \theta < V^\top \theta < X_{k'+1}^\top \theta < ... < X_N^\top \theta \right)\\
&\quad + C \mathbb{P}\left( X_1^\top \theta < ... < X_{N -k'}^\top \theta < V^\top \theta < X_{N -k'+2}^\top \theta < ... < X_N^\top \theta \right)
\end{align*}
That is,
\begin{align*}
     &\frac{1}{(k-1)!(N-k)!} \psi(V^\top \theta)^{k-1} \left(1 - \psi(V^\top \theta) \right)^{N-k}\\ 
     &\leq  \frac{C}{(k'-1)!(N-k')!} \left[ \psi(V^\top \theta)^{k'-1} \left(1 - \psi(V^\top \theta) \right)^{N-k'} +  \psi(V^\top \theta)^{N - k'} \left(1 - \psi(V^\top \theta) \right)^{k'-1}  \right]
\end{align*}
Hence,
\begin{align*}
    C \geq \frac{(k'-1)!(N-k')!}{(k-1)!(N-k)!} \cdot \frac{\psi(V^\top \theta)^{k-1} \left(1 - \psi(V^\top \theta) \right)^{N-k}}{ \psi(V^\top \theta)^{k'-1} \left(1 - \psi(V^\top \theta) \right)^{N-k'} +  \psi(V^\top \theta)^{N - k'} \left(1 - \psi(V^\top \theta) \right)^{k'-1} }
\end{align*}
Since $\psi(V^\top \theta) \in [0, 1]$, we have
\begin{align*}
    \frac{\psi(V^\top \theta)^{k-1} \left(1 - \psi(V^\top \theta) \right)^{N-k}}{ \psi(V^\top \theta)^{k'-1} \left(1 - \psi(V^\top \theta) \right)^{N-k'} +  \psi(V^\top \theta)^{N - k'} \left(1 - \psi(V^\top \theta) \right)^{k'-1}  } \leq 1
\end{align*}
for all $N$, $k$, and $k'$.
Hence, for $C = \frac{(k'-1)!(N-k')!}{(k-1)!(N-k)!}$, the claim holds.
\endproof

\section{Other Lemmas}

\begin{proposition}\label{prop:epsilon_subgaussian}
For each $\mathcal{E}_{t} = [\epsilon_{t1}, \epsilon_{t2}, ..., \epsilon_{t|S_t|}]^\top$, $\|\mathcal{E}_{t}\| \leq \sqrt{2}$.
\end{proposition}

\proof{}
Note that by the definition of $\epsilon_{ti}$, we have 
\begin{align}
\epsilon_{t1} + \epsilon_{t2} + ... + \epsilon_{t|S_t|} = 0, \quad \text{and} \quad \epsilon_{ti} \in [-1,1].\label{eq:hyperplane}
\end{align}
Hence the vector $\mathcal{E}_{t}$ lies within the bounded hyperplane in \eqref{eq:hyperplane}. Therefore, the $\ell_2$ norm $\|\mathcal{E}_{t}\| = \sqrt{\epsilon_{t1}^2 + \epsilon_{t2}^2 + ... + \epsilon_{t|S_t|}^2}$ is maximized at the corners of this bounded hyperplane, i.e., for some $i,j \in S_t$, $i \neq j$
\begin{align*}
    \epsilon_{ti} = 1, \epsilon_{tj} =  -1 \quad \text{and} \quad \epsilon_{tk} = 0, \text{ for all } k \neq i, k \neq j,
\end{align*}
which gives $\|\mathcal{E}_{t}\| \leq \sqrt{2}$. 
\endproof

\begin{lemma}[Hoeffding’s inequality]\label{lemma:hoeffding}
Let $X_1, ..., X_n$ be $n$ independent random variables such that $\mathbb{E}[X_i] = 0$ and almost surely, $X_i \in [a_i, b_i]$, for all $i$. Then for any $nu > 0$,
\begin{align*}
    \mathbb{P}\left( \left|\sum_{i=1}^n X_i \right| > \nu \right) \leq 2 \exp\left( -\frac{2 \nu^2}{ \sum_{i=1}^n (b_i - a_i)^2 } \right) \,.
\end{align*}
\end{lemma}


\begin{lemma}[\citet{tropp2012user}, Corollary~5.2]\label{lemma:matrix_concentration_iid}
Consider a finite sequence $\{\Yb_k\}$ of independent, random, self-adjoint matrices such that each $\Yb_k$ is positive semi-definite and $\lambda_{\max}(\Yb_k) \leq R$ almost surely.
Compute the minimum and maximum eigenvalues of the sum of expectations,
\begin{align*}
    \mu_{\min} := \lambda_{\min}\left(\sum_k \EE[\Yb_k]\right) \enskip \text{ and } \enskip \mu_{\max} := \lambda_{\max}\left(\sum_k \EE[\Yb_k]\right)
\end{align*}
Then
\begin{align*}
    &\PP \left\{ \lambda_{\min}\left(\sum_k \EE[\Yb_k]\right)
    \leq (1-\delta)\mu_{\min} \right\} 
    \leq d \cdot \left( \frac{e^{-\delta}}{(1-\delta)^{1-\delta}} \right)^{\mu_{\min}/R} \enskip \text{ for } \delta \in [0,1] \text{ and }\\
    &\PP \left\{ \lambda_{\max}\left(\sum_k \EE[\Yb_k]\right)
    \leq (1+\delta)\mu_{\max} \right\} 
    \leq d \cdot \left( \frac{e^{\delta}}{(1+\delta)^{1+\delta}} \right)^{\mu_{\max}/R} \enskip \text{ for } \delta \geq 0 \,.
\end{align*}
\end{lemma}

\section{Practical Extensions}

In this section, we briefly discuss some of the widely used problem settings in real-world applications, to which our proposed algorithms can be efficiently extended or reduced.

\subsection{Position Dependent Offering}\label{sec:display_position}
In many real-world applications, the choices of items are affected by not only their utilities but also the positions where they are displayed in the offered assortment \citep{ghose2014examining}. For example, in a brick-and-mortar store, items displayed in upper-shelf positions often receive more attention than those displayed in lower-shelf positions. Similarly, in an online store, items displayed at the top of the web page are more likely to be clicked or purchased than those displayed at the bottom. The effect of the display positions is usually unknown a priori.

In our proposed framework, we can easily incorporate display position effect by including a categorical variable indicating the display position. Hence, we need to estimate parameters corresponding to each display position. Suppose there are $K$ distinct display positions. Let $z_{tik}$ denote the upper confidence utility for item $i$ in round $t$ in display position $k \in [K]$ and let $w_{tik} := \exp(z_{tik})$. Then the optimal assortment choice $S_t = \{ (i,k) \in [N]\times[K] : \phi_{tik} = 1 \}$ can be given by the solutions of the following optimization problem:
\begin{equation}
\begin{aligned}
\max &\sum_{i \in [N], k \in [K]} \frac{r_{ti} w_{tik} \phi_{tik}}{1+ \sum_{ik} w_{tik} \phi_{tik}}\\
\textrm{s.t.} \quad & \sum_i \phi_{tik} \leq 1 \qquad \forall k \in [N] \\
  &\sum_k \phi_{tik} \leq 1   \qquad \forall i \in [N]\\
  & \phi_{tik} \in \{0,1\} \qquad \forall i \in [N], k \in [K]
\end{aligned}
\end{equation}
where $\phi_{tik}$ is the decision variable indicating item $i$ is displayed at position $k$ at round $t$. Note that the constraints satisfy that each position displays at most $1$ item, and each item is displayed at most once. 
\begin{proposition}[\citealt{davis2013assortment}]
The optimal position dependent assortment can be computed by solving an LP.
\end{proposition}
The proposition states that our algorithms can still use the LP solution for this position dependent extension of the combinatorial optimization problem. To see this, we first define the preference weight $w_{ti}(\theta) = \exp( x_{ti}^\top \theta)$ under some parameter $\theta$. Recall that, in the optimization step, we are indifferent of what parameter we use, i.e. the optimization step gives the optimizer set $S_t \subset [N]$ which maximizes the expected revenue given any parameter. Therefore, for the rest of this section we will use the notation $w_{ti}$  for brevity to denote  the preference weight given some parameter at round $t$.

For optimization procedure, we define the decision variable $\phi_{ti} \in \{0, 1\}$ such that $\phi_{ti} = 1$ if item $i$ is offered at round $t$, otherwise $\phi_{ti} = 0$. Under MNL, if the item offer decisions are given by the vector $\phi_t = \{\phi_{ti} : i \in [N]\} \in \{0,1\}^{N}$, then the user purchases item $i$ with probability $p_{i}(\phi_t) = \frac{w_{ti} \phi_{ti}}{1 + \sum_{j \in [N]} w_{tj}\phi_{tj}}$ where 1 in the denominator again represents the no-purchase option. Then we can rewrite the expected revenue as
\begin{equation*}
    R_t(\phi_t) = \sum_{i \in [N]} r_{ti} p_{i}(\phi_t) = \frac{\sum_{i \in [N]} r_{ti} w_{ti} \phi_{ti} }{1 + \sum_{j \in [N]} w_{tj}\phi_{tj}}
\end{equation*}
where $r_{ti}$ is the revenue parameter for item $i$ at round $t$. Based on the cardinality constraint on the assortment, The feasible set of assortment decisions are given by $\mathcal{F} = \{ \phi_{t} \in \{0,1\}^{N} : \sum_{i \in [N]} \phi_{ti} \leq K  \}$. Note that $\mathcal{F}$ defined here is a special case of totally unimodular constraint matrix for which \cite{davis2013assortment} show the LP formulation.
Then our goal is to find a set of feasible items to offer so as to maximize the
expected revenue:
\begin{equation}\label{eq:mnl_obj}
    R^*_t = \argmax_{\phi_t \in \mathcal{F}} R_t (\phi_t)
\end{equation}
where from $\phi^*_t = \argmax_{\phi_t \in \mathcal{F}} R_t (\phi_t)$ we can get the assortment $S_t = \{ i \in [N] : \phi_{ti} = 1\}$.
Note that problem \eqref{eq:mnl_obj} has a nonlinear objective function and integrality requirements on its decision variables. 
Theorem 1 in \cite{davis2013assortment} shows that problem \eqref{eq:mnl_obj} is equivalent to the following LP problem:

\begin{equation}
\begin{aligned}
\max &\sum_{i \in [N]} r_{ti} \rho_{ti}\\
\textrm{s.t.} \quad & \sum_{i \in [N]} \rho_{ti} + \rho_{t0} = 1\\
  &\sum_{i \in [N]} \frac{ \rho_{ti}}{ w_{ti}} \leq K  \rho_{t0}    \\
  & 0 \leq \frac{ \rho_{ti}}{ w_{ti}} \leq  \rho_{t0}
\end{aligned}
\end{equation}
where the decision variables are $\{ \rho_{ti} : i \in [N] \cup \{0\} \}$. In this LP problem, we can interpret the decision variable $\rho_{ti}, i \neq 0$ as the probability that the user purchases item $i$ in round $t$ and $\rho_{t0}$ as the probability that the user makes no purchase. The first constraint ensures that in each round a user purchases at most 1 item in the assortment, i.e., either purchases an item in the given assortment or purchase none.

For the position dependent offering, we rewrite the maximization problem in \eqref{eq:mnl_obj} by redefining the decision variable  $\phi_{tik} \in \{0,1\}$ as a binary variable indicating item $i$ is displayed at position $k$ at round $t$. 
\begin{equation}
\begin{aligned}
\max &\sum_{i \in [N], k \in [K]} \frac{r_{ti} w_{tik} \phi_{tik}}{1+ \sum_{i,k} w_{tik} \phi_{tik}}\\
\textrm{s.t.} \quad & \sum_i \phi_{tik} \leq 1 \qquad \forall k \in [N] \\
  &\sum_k \phi_{tik} \leq 1   \qquad \forall i \in [N]\\
  & \phi_{tik} \in \{0,1\} \qquad \forall i \in [N], k \in [K]
\end{aligned}
\end{equation}
Note that the constraints satisfy that each position displays at most $1$ item, and each item is displayed at most once.

\textbf{Comparisons with previous methods on position-dependent offering.}
The non-contextual setting in \cite{agrawal2016near,agrawal2017thompson} 
can be extended to incorporate position dependence; however, unlike in the setting here, the agent must offer every item in each position to learn the effect of display position. Therefore, the extension would create at least linearly increased amount of learning to their algorithm that is already not scalable for large $N$.
On the other hand, our proposed methods are able to learn the position effect across items. In \cite{chen2018dynamic}, it is possible to include a categorical variable corresponding to display position as part of  context vector; however, this will result in a further exponential increase in computational complexity. Moreover, their method cannot exploit that fact that the assortment optimization problem is an LP~(see the discussion on Section~\ref{sec:related_work}).

\subsection{Top-$K$ Selection with User Choice Consideration}
The top-$K$ selection problem \citep{cao2015top} is not necessarily an extension but rather a special case of the MNL bandit problem where the revenue parameters are uniform. Hence, our problem reduces to finding $K$ items which have the highest utility values. Note that this special case still differs from other variants of combinatorial bandits such as semi-bandits and cascading bandits in that top-$K$ offering \textit{may} still take the substitution effect into account. This special case is particularly important because of its wide range of applications. For example, the decision-making agent may want to maximize the click-through rate (CTR) on a website where each click is weighted uniformly. A notable aspect of the top-$K$ selection problem is that the combinatorial assortment selection step reduces to a sorting task based on estimated utilities in our proposed algorithms, making the assortment selection procedure much more computationally efficient. However, \cite{chen2018dynamic} still has to enumerate all $N$ choose $K$ many assortments and construct upper the confidence bounds of utilities for each of the assortments to choose the items even in this setting.

\section{Experiment Details and Additional Results}

We consider two multivariate distributions for feature vectors: a multivariate Gaussian distribution and uniform on a unit sphere. For a a multivariate Gaussian distribution, we draw each $x_{ti}$ i.i.d. from $\mathcal{N}(\vec{0}_d, I_d)$. Since the vanilla version of \textsc{MLE-UCB} is an exponential-time algorithm. We use their greedy heuristic version which does not provide a performance guarantee. 
For efficient evaluations, we consider uniform revenues, i.e., $r_{ti} = r$ for all $i$. Therefore, the combinatorial optimization step reduces to sorting items according to its utility estimate.

For each instance, we generate the true parameter $\theta^*$ from a uniform distribution in $[0,1]^d$ and simulate accordingly.
For each case with different experimental configurations, we conducted 20
independent runs for each instance, and report the average of the cumulative regret for each of the algorithms. The error bars represent the standard deviations. Note that each instance is generated using different random seeds.

\begin{figure*}
\begin{subfigure}[b]{0.32\textwidth}
    \includegraphics[width=\textwidth]{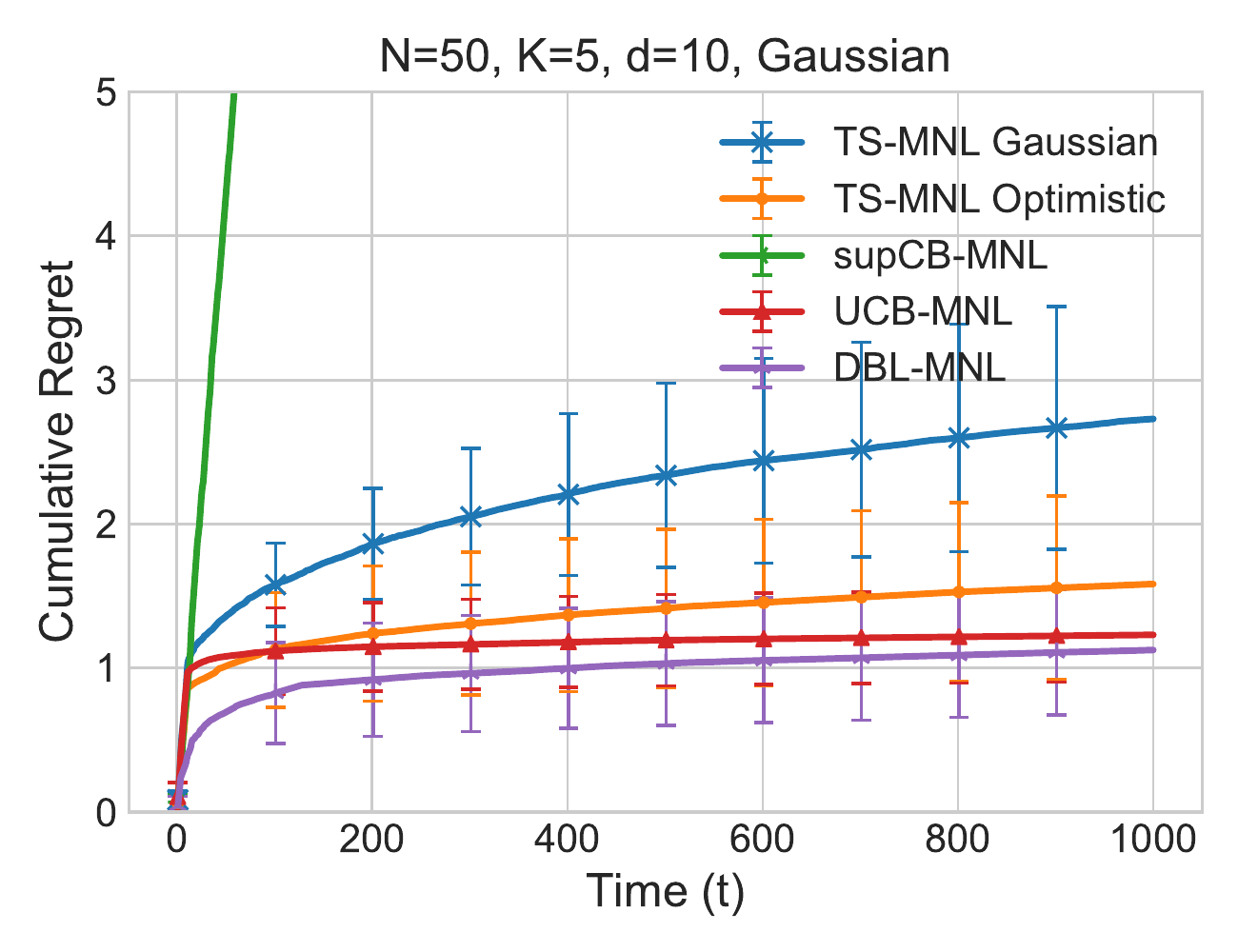}
\end{subfigure}
\begin{subfigure}[b]{0.32\textwidth}
    \includegraphics[width=\textwidth]{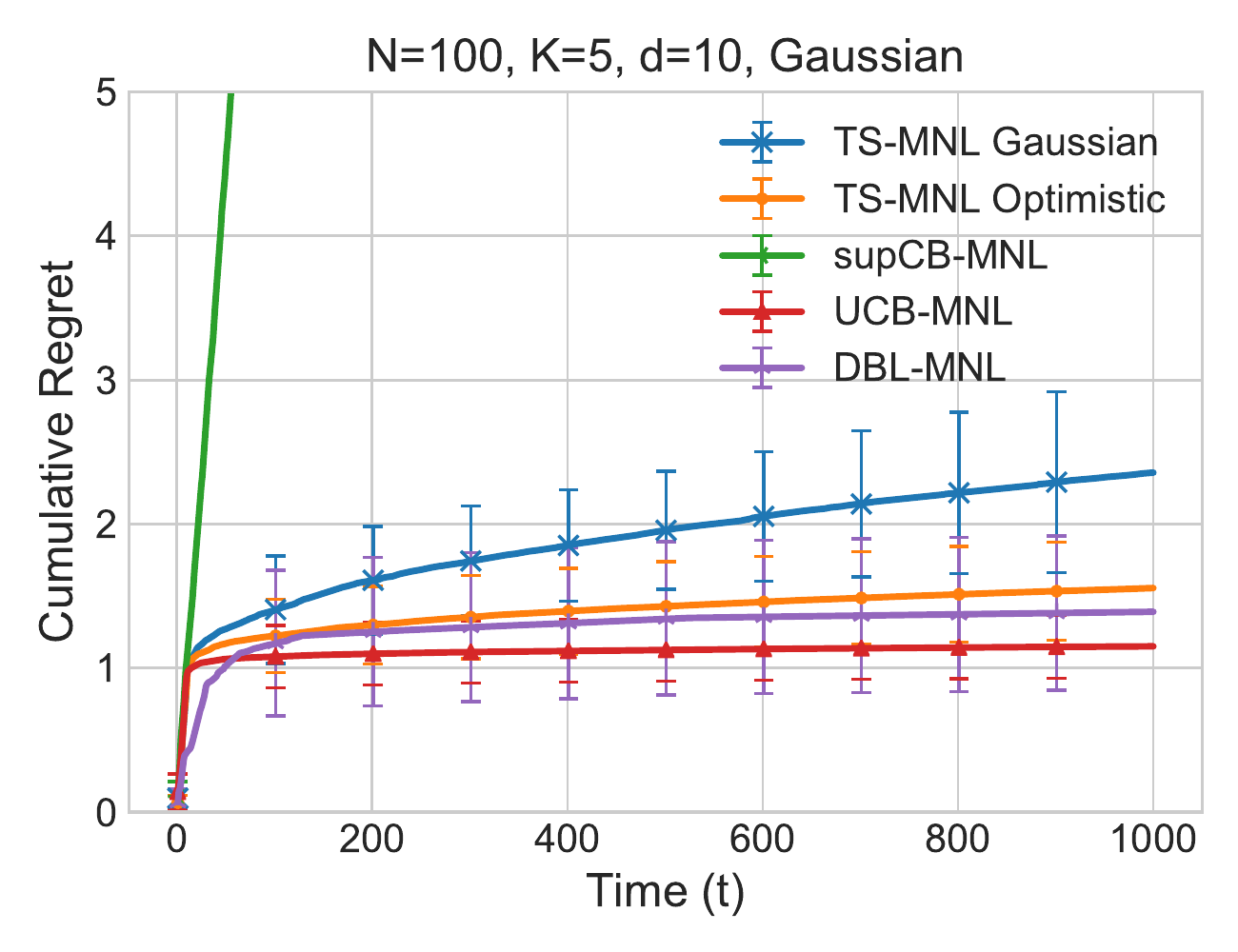}
\end{subfigure}
\begin{subfigure}[b]{0.32\textwidth}
    \includegraphics[width=\textwidth]{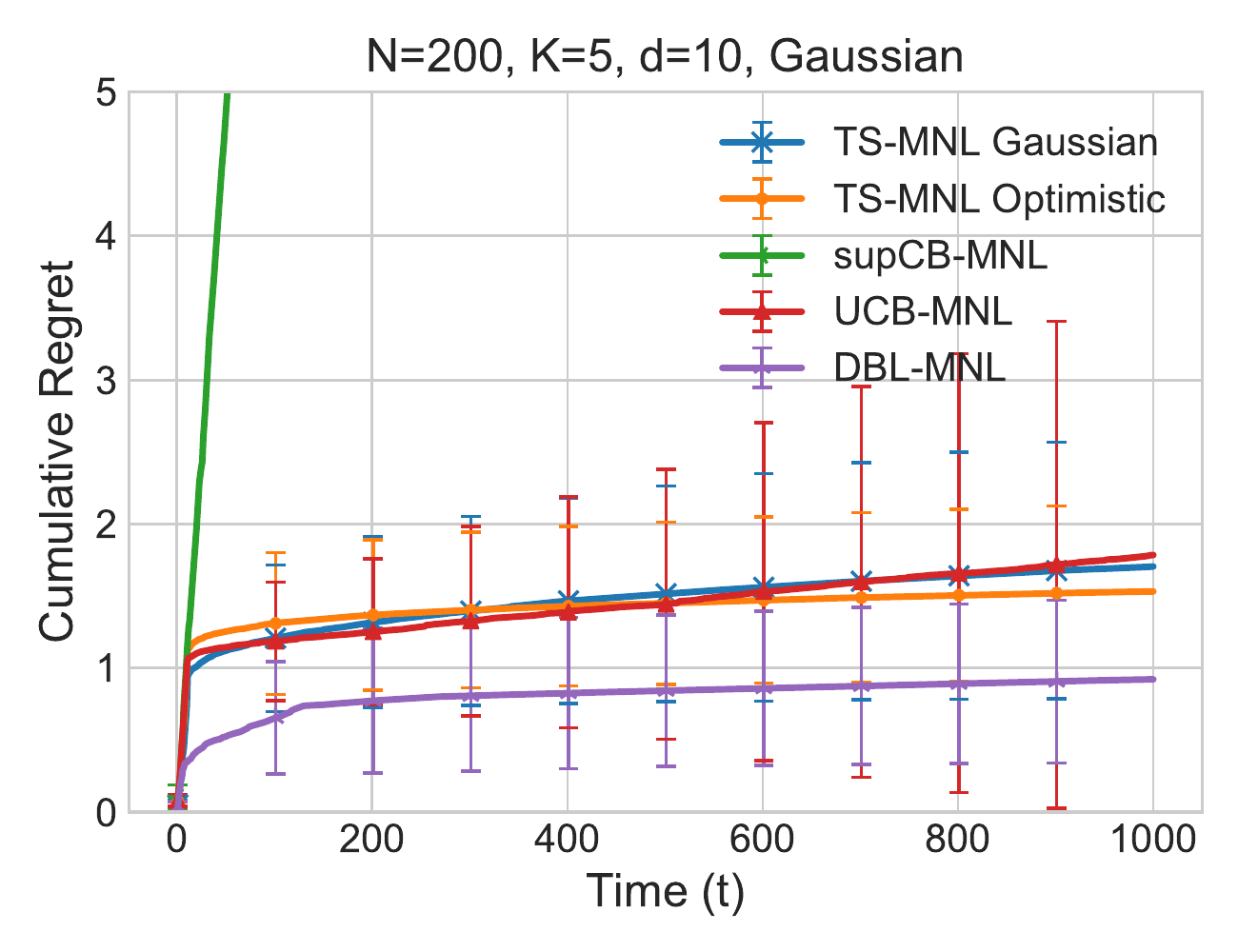}
\end{subfigure}\\
\begin{subfigure}[b]{0.32\textwidth}
    \includegraphics[width=\textwidth]{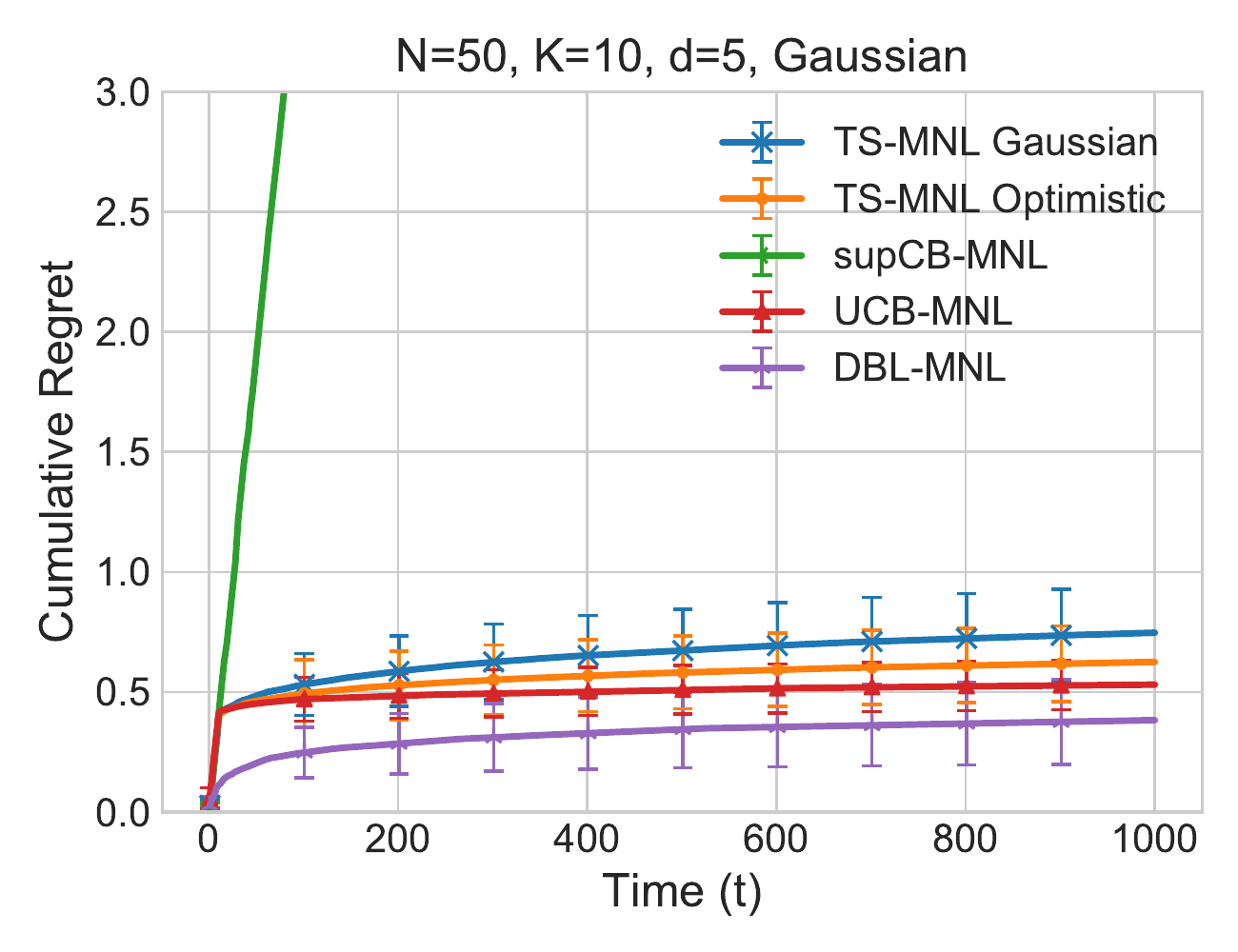}
\end{subfigure}
\begin{subfigure}[b]{0.32\textwidth}
    \includegraphics[width=\textwidth]{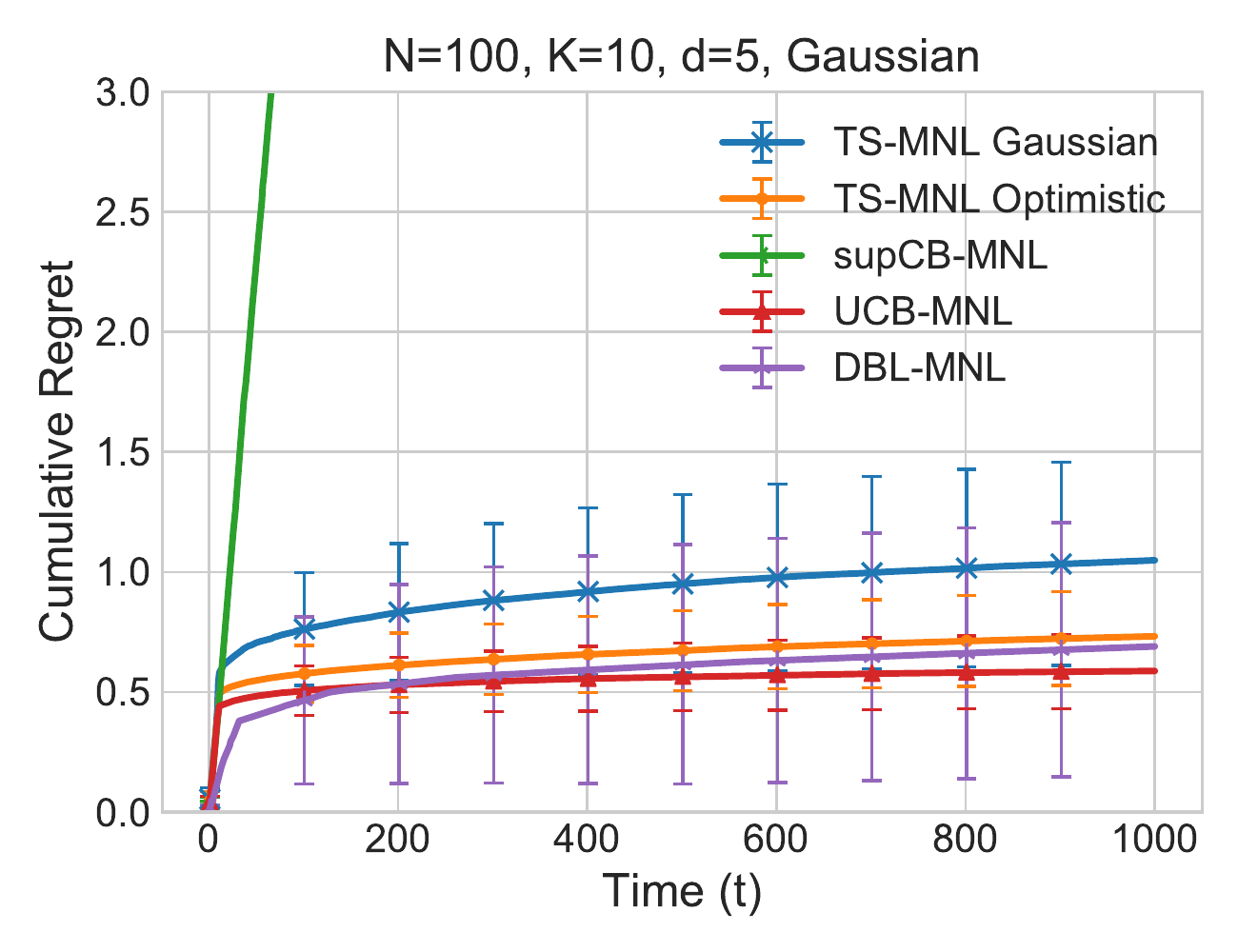}
\end{subfigure}
\begin{subfigure}[b]{0.32\textwidth}
    \includegraphics[width=\textwidth]{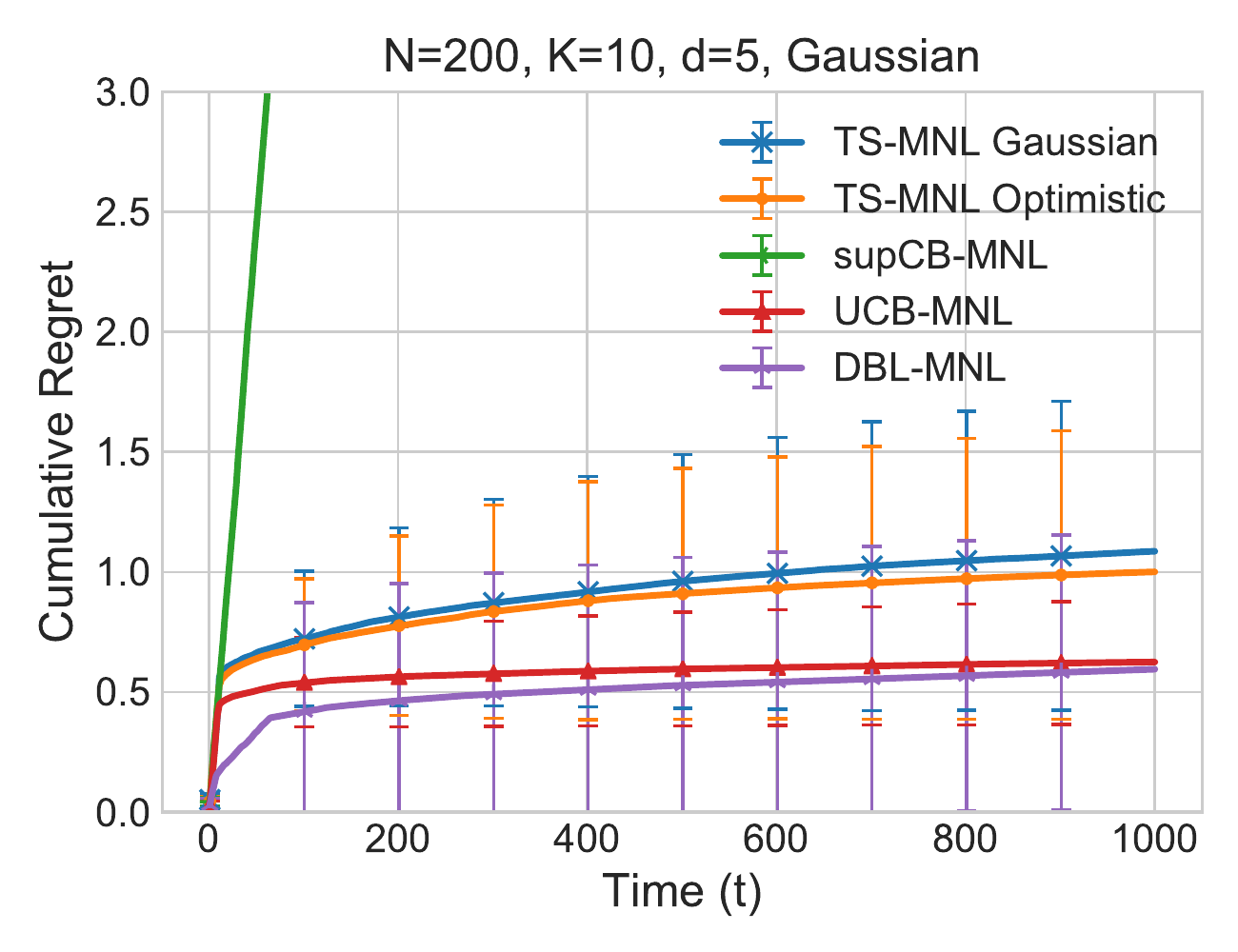}
\end{subfigure}\\
\caption{The regret plots show that the proposed algorithms, \textsc{UCB-MNL} and \textsc{DBL-MNL}, perform at start-of-the-art levels across different problem instances.}
\label{fig:mnl_regret_additional}
\vspace{-0.3cm}
\end{figure*}

Figure~\ref{fig:mnl_regret} and Figure~\ref{fig:mnl_regret_additional} show the sample results.  
The performance of \textsc{UCB-MNL} and \textsc{DBL-MNL} are superior to or comparable to the existing method. \textsc{UCB-MNL} 
As expected, \textsc{supCB-MNL} that relies on the Auer-framework~\citet{auer2002using}
is not competitive. It wastes too many samples for random exploration.
We also conduct run-time experiments for the algorithms reported in Table~\ref{tab:runtime}. We observe that \textsc{DBL-MNL} is significantly more efficient computationally compared to the other methods due to its logarithmic number of parameter updates. Note that \textsc{supCB-MNL} has a pruning assortment step which can be computationally expensive. However, for uniform revenues (which is considered in the experiments shown here), this procedure can be performed in a much more manageable manner. Furthermore, in our experiments almost all of the action selections of \textsc{supCB-MNL} came from the exploration step (which explains the poor performances), and therefore the run-time was reported smaller than \textsc{DBL-MNL} and TS methods, but this may not be true in general once the pruning step is used more often.
Overall, the experiments show that both \textsc{UCB-MNL} and \textsc{DBL-MNL} can learn to find the optimal policy quickly while \textsc{DBL-MNL} is also very efficient computationally.

\end{document}